\documentclass{article}
\usepackage{arxiv}
\usepackage[T1]{fontenc}    % use 8-bit T1 fonts

\usepackage{url}            % simple URL typesetting
\usepackage{booktabs}       % professional-quality tables
\usepackage{amsfonts}       % blackboard math symbols
\usepackage{nicefrac}       % compact symbols for 1/2, etc.
\usepackage{microtype}      % microtypography
\usepackage{lipsum}		% Can be removed after putting your text content
\usepackage{graphicx}
\usepackage{doi}
\usepackage{multirow}
\usepackage{authblk}
\usepackage[symbol]{footmisc}
\usepackage{tabularx}
\usepackage{xcolor}
\usepackage{marvosym}
\usepackage{ifsym}
\usepackage{pifont}
\usepackage{url}
\usepackage{multicol}
\usepackage{amsmath}
\usepackage{hyperref}       % hyperlinks
\begin{document}

\title{A unified multi-task framework enables interpretable chest radiograph analysis}

% Running title（页眉用）
\renewcommand{\headeright}{Learning a Multi-task Transformer for Chest Radiograph Interpretation}

\renewcommand{\shorttitle}{}

% 第一页：Graphical Abstract
\thispagestyle{empty}

% 标题（居中）
\begin{center}
    {\LARGE \textbf{A unified multi-task framework enables interpretable chest radiograph analysis}}
\end{center}

\vspace{0.5cm}

% \section*{Graphical abstract}
\noindent \textbf{\large Graphical abstract}

% 图片（居中，无浮动体）
\begin{center}
    \includegraphics[width=0.99\textwidth]{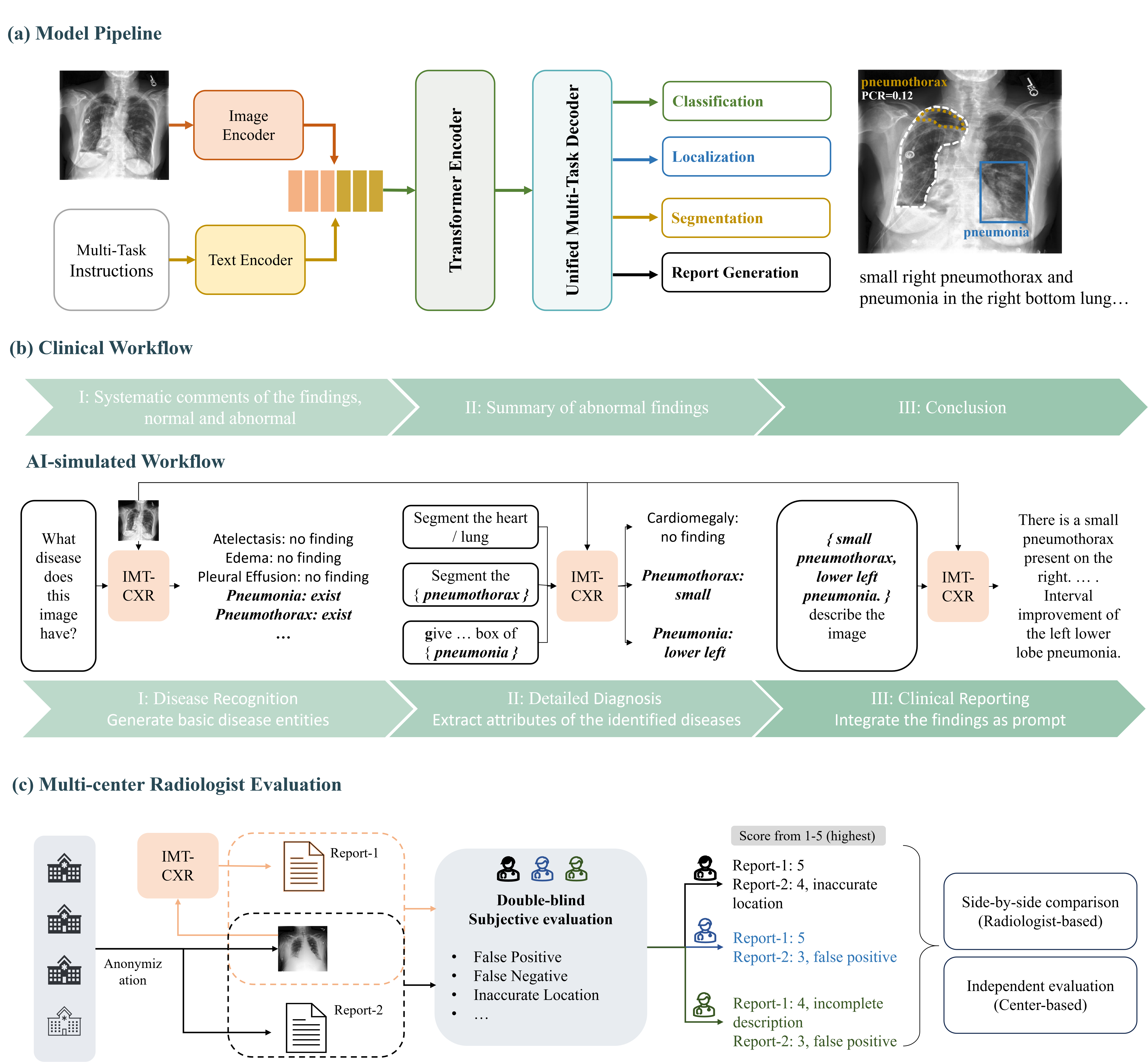}
\end{center}

\vspace{0.5cm}

% % Highlights（图片正下方）
% \noindent \textbf{\large Highlights}
% \begin{itemize}
%     \renewcommand{\labelitemi}{-}
%     \item We propose a unified multi-task transformer for chest X-ray analysis integrating classification, localization, segmentation, and report generation.
%     \item The framework enables interpretable diagnosis by linking disease recognition, attribute characterization, and evidence-based report generation.
%     \item The model achieves competitive performance across ten chest X-ray benchmarks under direct inference and fine-tuning settings.
%     \item Radiologist evaluation shows that 66\% of AI-generated reports are comparable to or better than clinical reports in diagnostic clarity.
% \end{itemize}

% \vspace{0.6cm}

\vspace{0.3cm}

% In brief
\noindent \textbf{In brief} \\
Xu et al. present a unified multi-task transformer for chest radiograph analysis that integrates classification, localization, segmentation, and report generation. The framework enables interpretable diagnosis through evidence-linked reasoning and demonstrates strong performance across benchmarks, with radiologist evaluation confirming clinical relevance and report quality.

\newpage

\author[$\ast$,\Letter,1,2,5]{Lijian~Xu}
\author[$\ast$,3]{Ziyu Ni} 
\author[3]{Xinglong~Liu}
\author[\Letter,3]{Xiaosong~Wang}
\author[2,4]{Hongsheng~Li} 
\author[2,3,6]{Shaoting~Zhang}

\affil[1]{Shenzhen University of Advanced Technology}
\affil[2]{Centre for Perceptual and Interactive Intelligence, the Chinese University of Hong Kong, Hong Kong}
% \affil[3]{Sensetime Research, Shanghai}
\affil[3]{Shanghai Artificial Intelligence Laboratory, Shanghai}
\affil[4]{Department of Electronic Engineering, the Chinese University of Hong Kong, Hong Kong}

\affil[5]{Lead contact}
\affil[6]{Senior author}
\affil[$\ast$]{Equal contributions}
% \affil[$\ast$]{Correspondence: xulijian@suat-sz.edu.cn (L.X.), xiaosong.wang@live.cn (X.W.)}

% \renewcommand{\preprint}{}   % 清空 preprint 文字
\date{}

% Uncomment to override  the `A preprint' in the header
\renewcommand{\headeright}{Learning a Multi-task Transformer for Chest Radiograph Interpretation}

 \renewcommand{\shorttitle}{}

\maketitle

% \begin{abstract}

\noindent \textbf{\large Context and Significance}

% \section*{Context and Significance}

Chest X-ray interpretation is a fundamental task in clinical radiology, but most artificial intelligence systems provide limited transparency regarding how diagnostic decisions are made. This work develops a new multi-task framework that models the step-by-step reasoning process of radiologists by integrating disease classification, lesion localization, anatomical segmentation, and automated report generation within a unified model. The study shows that this structured approach achieves strong performance across multiple benchmarks and produces radiology reports that are frequently judged by experts to be comparable to clinical reports. These findings suggest that incorporating intermediate reasoning steps can improve both the accuracy and interpretability of medical imaging artificial intelligence systems, potentially supporting safer and more trustworthy clinical decision support tools.

% \section*{Summary}
\noindent \textbf{\large Summary}

\textbf{Background:}
While multimodal deep learning has advanced medical imaging analysis, existing black-box systems \textcolor{black}{may remain confined to isolated tasks, often overlooking} the trust-sensitive nature of clinical diagnosis as a multi-task process. 

\textbf{Methods:}
We propose IMT-CXR (Interpretable Multi-task Transformer for Chest X-ray Analysis), a framework that emulates radiologists' diagnostic workflow through three evidence-driven stages: 1) Disease recognition; 2) Attribute characterization (e.g., size, location, severity quantification); 3) Evidence-integrated report generation with traceable decision pathways. The framework employs a unified transformer architecture optimized via medical-domain instruction tuning, sequentially executing four clinical tasks: multi-label disease classification, lesion localization, anatomical segmentation, and radiology report generation. 

\textbf{Findings:}
Experimental validation demonstrates competitive performance on ten CXR benchmarks under direct inference and fine-tuning settings. In a blinded evaluation of 160 historical reports from four medical centers, three radiologists rated 66\% of AI-generated reports as comparable to or surpassing original clinical reports in diagnostic clarity, highlighting the framework's translational potential. 

\textbf{Conclusions:}
By establishing traceable diagnostic pathways from anatomical findings to conclusions, this work bridges the gap between AI technical metrics and clinical utility, advancing trustworthy AI systems in medical imaging.

\textbf{Funding:}
This research was partially supported by the Centre for Perceptual and Interactive Intelligence (CPII) Ltd under the Innovation and Technology Commission (ITC)'s InnoHK.

% keywords can be removed

% \newpage

\section{Introduction}

Chest X-ray (CXR) is a non-invasive and relatively low-cost diagnostic radiology examination for screening and diagnosis of various thoracic diseases affecting the lung and heart \cite{xu2024foundation,xu2024medvilam}. However, the interpretation of CXR is greatly challenged by its low sensitivity of subtle abnormalities, overlapping structures, and limited soft tissue details, and therefore, depends heavily on the capability and experience of radiologists \cite{feng2026efficient,young2026xrayclaw}. On the other hand, the growing demand for CXR examination has brought a burden on medical professionals, which also limits the clinical application of CXR, especially in community clinics or primary hospitals. In this context, automated diagnosis by AI could potentially contribute to reducing the workload of radiologists.

Large Language Models \cite{dosovitskiy2021image,liu2021swin} 
% have revolutionized natural language processing, and self-attention-based architectures, in particular Transformers~\cite{vaswani2017attention}, have become the model of choice in language and visual tasks~\cite{dosovitskiy2021image,liu2021swin}. By leveraging extensive textual data, these models 
have revolutionized natural language processing and developed the capability to generate responses that closely resemble those from humans. They excel at a wide range of tasks, including language translation, question answering, and text generation 
%\cite{devlin2018bert,touvron2023llama,scao2022bloom,chowdhery2023palm,du2022glam,zeng2022glm}. 
\cite{devlin2018bert,touvron2023llama,chowdhery2023palm,du2022glam,zeng2022glm}. 
% \cite{devlin2018bert,touvron2023llama,scao2022bloom,chowdhery2023palm,zhang2022opt,du2022glam,zeng2022glm}. 
Models like ChatGPT (OpenAI) \cite{ouyang2022training} and Med-PaLm (Google) \cite{singhal2023towards} 
% \cite{brown2020language,ouyang2022training} and Med-PaLm (Google) \cite{singhal2023large,singhal2023towards} 
have also demonstrated powerful reasoning capabilities of language models in complex scenarios like medical diagnosis to assist professionals in delivering care.
% Language models with prompt learning~\cite{brown2020language,ouyang2022training,chen2021evaluating} prove powerful zero-/few-shot learners and provide a new paradigm for human-computer interaction.
% Building upon this, multimodal large models utilize pre-trained image encoder and text encoder then align visual-language features with simple linear layers\cite{radford2021learning,alayrac2022flamingo,li2023blip2,chenpix2seq,wang2022ofa,wang2022image,chen2023pali,li2022grounded}. The stronger generalizable language features are leveraged to guide the extraction of visual features. 
% These models demonstrate impressive joint understanding capabilities of language and images, allowing users to give instructions in natural language to perform specific tasks. 
% Multimodal large models can recognize objects in images and accomplish tasks such as Visual Question Answering (VQA) and Image Captioning. 
Nonetheless, such tasks are limited to a more general medical scope and largely rely on the visual features on the image level \cite{yang2025one,young2026fewer,he2026autoselect}, without addressing the pixel-level vision tasks, e.g., lesion localization and segmentation \cite{yang2024segmentation}. Moreover, to further improve the model's ability on downstream tasks, supervised instruction tuning with specific downstream task-oriented data is often required on language only \cite{wei2022finetuned,chung2024scaling,gao2026zerosense} and vision-language tasks 
% \cite{dai2023instructblip,zhu2024minigpt,liu2024visual}, 
\cite{dai2023instructblip,young2026scalar,liu2024visual}, 
individually.

On the other hand, the development of the multi-modal models in the medical field has lagged. Most models are designed primarily for "pure" language tasks \cite{rasmy2021med,singhal2023large}. 
Several generalist models for the biomedical field have been recently proposed and achieved progress in VQA tasks \cite{moor2023foundation,zhang2024generalist,xu2024medvilam,li2024llava}.
However, the present multi-modality models are not well-suited for traditional image processing tasks like detection and segmentation. Existing methods face discrepancies in input, output, and training processes between visual tasks and language tasks, which hinders efficient collaboration. Furthermore, relying solely on textual outputs restricts the answer capacity and interpretability to some extent. For instance, in computer-aided diagnosis using medical images, while the model can identify the disease type and provide treatment recommendations, it is unable to pinpoint the exact location and region of the pathologies, limiting its clinical usefulness as a reference for explainable diagnosis prediction.

\begin{figure}[tbh!]
    \centering
\includegraphics[width=0.99\linewidth]{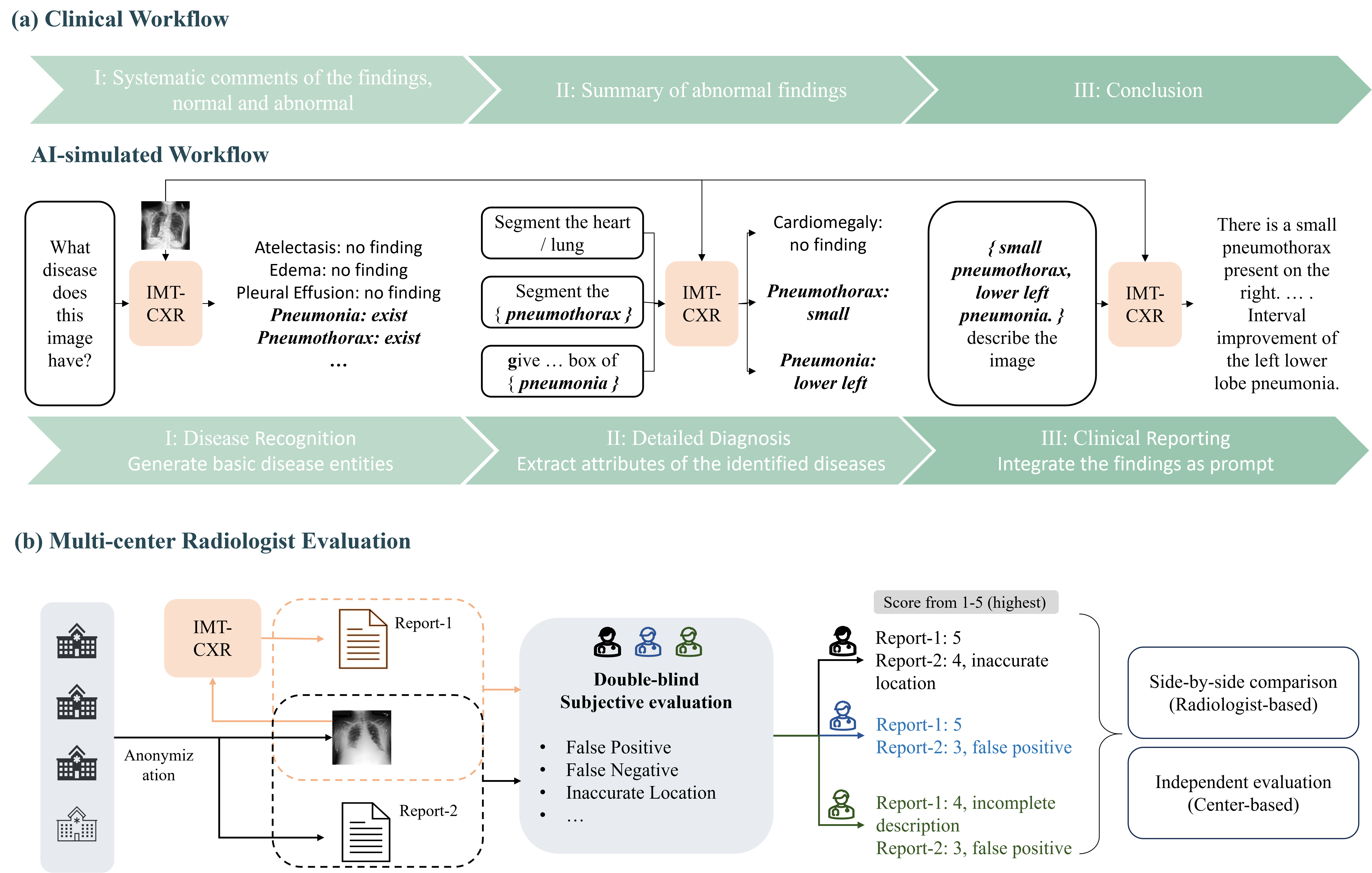}
%     \caption{The potential clinical application of the proposed method. (a) Conventional clinical workflow and AI-simulated workflow of chest X-ray diagnosis. 
%     Doctors think systematically to make sure nothing is missed, they also want to know the status of all relevant structures, even they are normal. 
%     Correspondingly, the AI-simulated workflow includes three stages: 1) Identify the diseases present; 2) Analyze attributes of the identified diseases (dashed box) and the status of relevant structures (solid box); 3) Integrate the findings to generate a report. 
% (b) Multi-center radiologist evaluation conducted across three private centers and one public center. Three radiologists, ranging from junior to senior levels, evaluated both clinical reports and the reports generated by IMT-CXR in a double-blind manner. Evaluation criteria included standards such as false positive and false negative. Each radiologist assigned a score from 1 to 5, with 1 being the lowest and 5 being the highest, along with a justification for their rating. }
\caption{
(a) Comparative clinical workflows: Conventional chest X-ray diagnosis relies on radiologists' systematic evaluation of all anatomical structures (normal/abnormal) to prevent oversight, whereas the proposed AI-simulated workflow emulates this cognitive rigor through three stages: 1) Disease identification; 2) Attribute analysis (dashed box: disease characteristics [size/location/severity]; solid box: the status of all relevant structures); 3) Evidence-integrated report generation.
(b) Multicenter radiologist evaluation: A double-blind assessment was conducted across three private hospitals and one public institution, involving three radiologists (junior to senior). Participants independently scored (1-5 scale) and justified ratings for both AI-generated (IMT-CXR) and original clinical reports based on diagnostic criteria (e.g., false positives/negatives).}
    \label{fig:workflow}
\end{figure}

To bridge the critical gaps between clinical translatability and technical integration, we propose IMT-CXR (Interpretable Multi-task Transformer for Chest X-ray Reading), an AI-simulated clinical decision framework that enables evidence-based reasoning through hierarchical feature discovery. As illustrated in Figure \ref{fig:workflow}, our system implements a three-phase clinical workflow:
1) Comprehensive disease screening via multi-class classification tasks;
2) Quantification of disease attributes (e.g., size, location, severity) and assessment of adjacent anatomical integrity;
3) Integration of findings through clinical decision pathways to generate diagnostic reports.
Technically, the proposed multi-task framework addresses critical challenges in chest X-ray analysis, including diagnosis of common thoracic diseases and localization of those image-visible disease patterns (see Figure \ref{fig:overview}). Additionally, it can perform image segmentation of pneumothorax, lungs, and heart regions. 
% Most importantly, IMT-CXR is capable of generating reports summarizing the findings by leveraging all the provided evidence mentioned here.
A key innovation lies in its ability to synthesize all extracted evidence into coherent diagnostic reports, thereby mimicking radiologists’ holistic reasoning process.

Our contributions are three-fold:

1) This study introduces an AI-simulated workflow to provide radiologists and other professional users with intermediate results, avoiding jumping to conclusions. 
% It closely aligns with the clinical reasoning and emulates the clinical workflow by providing comprehensive analysis through detailed outputs regarding disease attributes, such as size, location, severity, and contour. 
% This richer set of information enhances the interpretability of chest X-ray reports, offering stronger evidence for diagnosis and treatment decisions.
% Such richer outputs will make the diagnosis prediction more explainable, traceable and flexible, i.e., the radiologists may manually alter some intermediate results to correct and reproduce the final report, which other existing methods are incapable of.
It enhances the interpretability of chest X-ray reporting by generating more detailed information on disease attributes, including disease size, location, severity, and contour, providing stronger evidence for diagnosis and treatment. 
% Furthermore, we conducted a controlled trial and evaluation on the generated reports performed by three radiologists. In the blinded comparison of 160 retrospective cases from four centers, three radiologists perceived the quality of 66\% of the generated reports to be equivalent to or better than the original physician reports. 
Notably, the proposed method is further verified to perform equivalently to junior radiologists in a blinded comparison involving 160 historical reports from four centers, showing its potential clinical value in chest X-ray interpretation.

2) Our proposed model offers a versatile approach to analyzing chest X-ray images, allowing for comprehensive and accurate radiography image analysis across key tasks in clinical chest X-ray diagnosis.
We applied the proposed model to various downstream application benchmarks, and an overall comparable performance is shown compared to advanced specialist approaches.

3) For multi-task joint training, we developed a unique framework for building datasets that are tailored for customized instruction tuning. Unlike the conventional method of organizing pair-wise supervision (consisting of an image and its corresponding label), our framework involves cross-task training supervision for each sample, which enhances the learning of correlations among tasks. 
We have released the composed datasets with detailed image-instruction-label triplets, representing a novel dataset of this kind to our knowledge. Additionally, the private datasets included in the benchmark for report generation and pneumothorax segmentation tasks have also been released.

\section{Results}
\subsection{Interpretable AI-simulated Diagnostic Workflow}
As illustrated in Figure \ref{fig:workflow}(a), conventional radiology reports follow a structured three-phase documentation process: 1) Systematic comments on the finding, normal or abnormal; 2) Summary of abnormal findings;3) Conclusion. Doctors think systematically to make sure nothing is missed, they also want to know the status of relevant structures, even if they are normal.
% Through discussions with radiologists, we were inspired by the idea of mimicking this process of doctors through the proposed multi-task approach.
Inspired by clinician consultations, we propose an AI-simulated workflow that emulates this clinical reasoning through detailed outputs regarding disease attributes, such as size, location, severity, and contour. 
Such richer outputs will make the diagnosis prediction more explainable, traceable and flexible, i.e., the doctors may manually alter some intermediate results to correct and reproduce the final report, which other existing methods are incapable of.

\textbf{Multi-center Radiologist Evaluation} We conducted a controlled trial for the evaluation of the generated report in comparison to the original reports (see Figure \ref{fig:workflow}(b)). 
In a blinded comparison involving 160 historical reports from four different centers, 
% three radiologists consistently rated the quality of the generated reports as comparable to or better than original radiological reports, with a success rate of over 66\%. 
three radiologists rated 66\% of AI-generated reports as comparable to or surpassing original clinical reports in diagnostic clarity. 
Moreover, our proposed model exhibited an omission rate of 1.87\% (95\% CI 1.06-2.83\%) and an error rate of 2.24\% (95\% CI 1.39-3.20\%) per report, which are close to those of the radiologist-provided reference reports (i.e., 1.25\% (95\% CI 0.67-2.00\%) and 2.00\% (95\% CI 1.19-2.89\%)). \textcolor{black}{Errors occurred predominantly in Atelectasis (52.3\%), whereas omissions were mainly associated with Consolidation/Pneumonia (58.1\%).}

\begin{figure}[tbh!]
    \centering
    \includegraphics[width=0.99\linewidth]{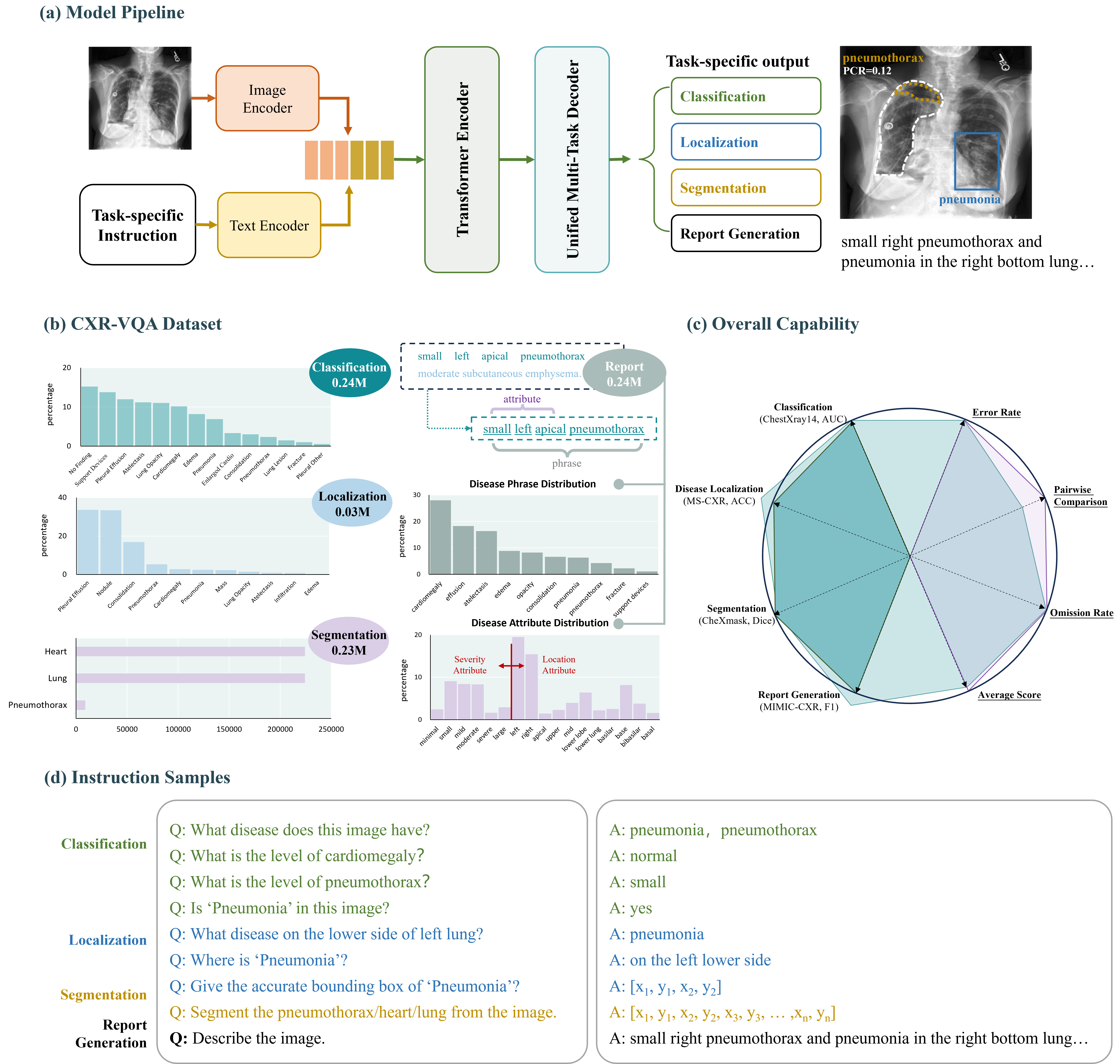}
\caption{Overview of IMT-CXR framework.
(a) Multimodal inputs (image and task-specific instructions) and multi-task outputs (classification, localization, segmentation, report generation).
(b) CXR-VQA training dataset: Disease attributes (severity, location) extracted from radiology reports.
(c) Performance comparison across datasets and tasks: Metrics outside/inside the circle denote superiority/inferiority to benchmarks.
(d) Representative VQA samples from the instruction set.}
    \label{fig:overview}
\end{figure}

\textbf{Multi-task Model for Chest X-ray analysis } We developed a multi-task model to enable comprehensive chest X-ray analysis with evidence-based interpretability. As shown in Figure \ref{fig:overview}, IMT-CXR integrates four clinical tasks, i.e., disease classification, lesion localization, anatomical segmentation, and report generation, while providing strong performance across individual evaluation metrics.
The architecture (Figure \ref{fig:overview}(a)) comprises three core components: an image encoder for processing X-ray images, a text encoder for interpreting clinical instructions, and an encoder-decoder language model for generating task-specific outputs. During inference, multimodal inputs (images and instructions) are independently encoded and fused into joint representations, which the language model decodes into final predictions (see Section 4.2 for implementation details).

We composed a multi-task training dataset 
\cite{johnson2019mimic,nguyen2022vindr,lian2021structure,gaggion2024chexmask,siim} 
% \cite{johnson2019mimic,padchest,fan2024deep,nguyen2022vindr,lian2021structure,gaggion2024chexmask,siim} 
with approximately 0.65 million radiographs (Figure \ref{fig:overview}(b)) and specially designed image-instruction-label triplets comprising 8.8 million instruction 
and ground-truth pairs,
involving both image- and pixel-level tasks (Figure \ref{fig:overview}(d)). 
Unlike conventional methods that organize supervision in pairs (an image and its corresponding label), our framework employs cross-task training supervision for each sample, thereby enhancing the learning of correlations among tasks.

\textbf{Comprehensive Benchmark} 
We established a comprehensive benchmark comprising 145 thousands of testing cases from both public 
\cite{wang2017chestx,irvin2019chexpert,shih2019augmenting,jsrt2000lung,boecking2022making} and private datasets. 
For multi-label classification, performance was assessed on ChestX-ray14 \cite{wang2017chestx} (macro-average AUC and F1 scores across 14 pathologies), CheXpert \cite{irvin2019chexpert} (5 pathologies), and RSNA Pneumonia \cite{shih2019augmenting}. lesion localization capabilities were evaluated using MS-CXR \cite{boecking2022making}, ChestX-ray14, and RSNA Pneumonia datasets, with accuracy (ACC) and mean Intersection over Union (mIoU) serving as primary metrics. Anatomical segmentation performance was quantified via the Dice coefficient across three specialized datasets: JSRT (lung), CheXMark (lung and heart), and MS-PS (pneumothorax). Report generation quality was validated on MIMIC-CXR through dual evaluation paradigms, i.e., clinical efficacy metrics (F1 score, precision, recall) and natural language processing benchmarks (BLEU-4, METEOR, ROUGE-L). 
We have included both advanced specialist models with direct inference and those require further fine-tuning for comprehensive evaluation
\cite{chenSimpleFrameworkContrastive2020,zhang2022contrastive,Huang2021GLoRIAAM,boecking2022making,Zhou_2022,geng2022multimodal,wang2022multi,wu2023medklip,xiao2023delving,zhou2023advancing,VGTR2022vg,Zhu_2022,Deng_2021_ICCV,chen2023medical,muller2025chex,zhang2023knowledge,li2021referring,anderson2018bottom,rennie2017self,chen2021cross,liu2023contrastive,tanida2023interactive,jeong2024multimodal}.
This multi-modal validation strategy rigorously aligns technical performance with clinical workflow requirements.

\begin{figure}[t]
	\centering
\includegraphics[width=0.98\linewidth]{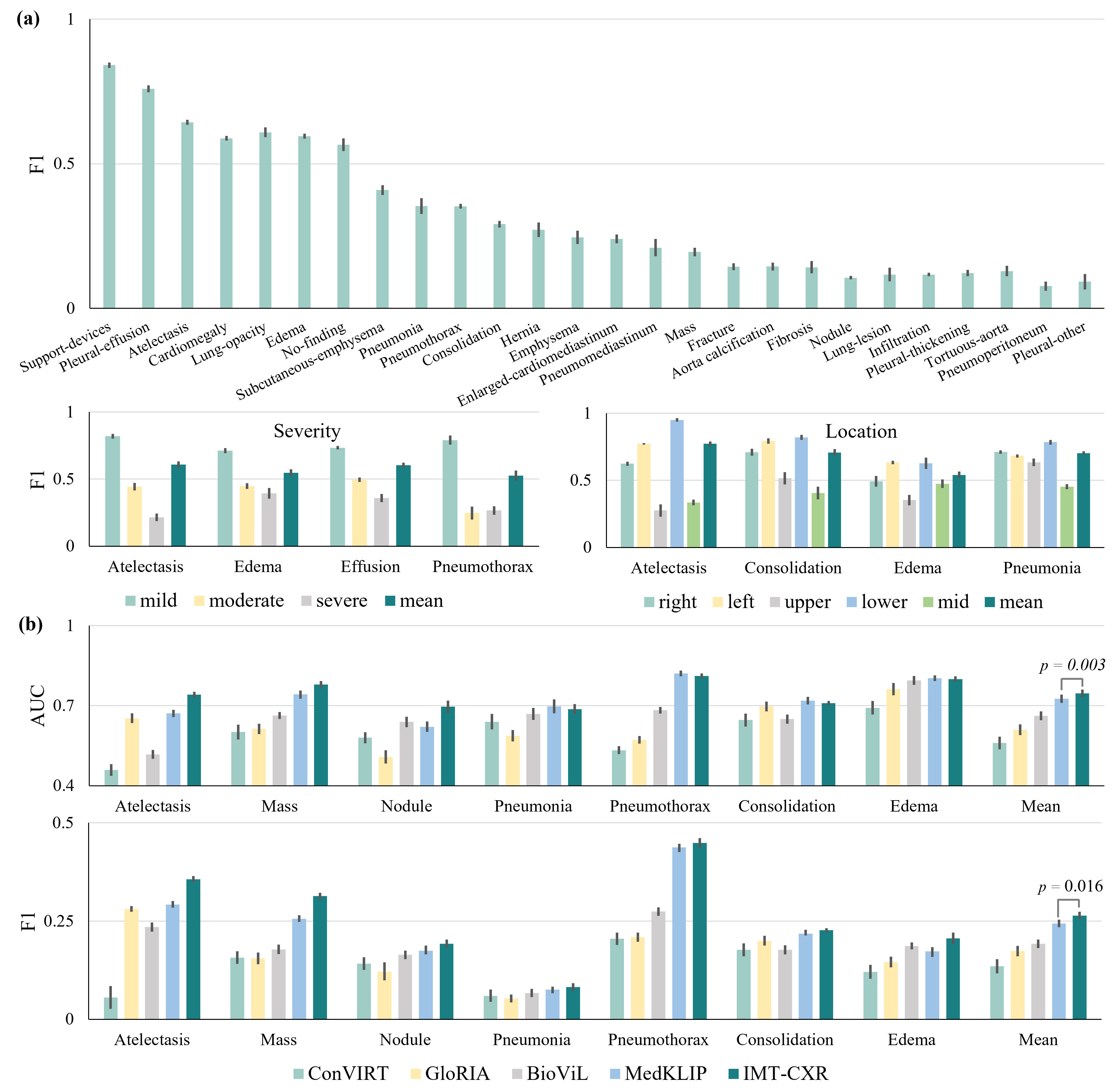}
	\caption{(a) In domain evaluation of 26 disease entities on the MIMIC-CXR dataset (upper panel) and attribute classification task in disease severity level and location (lower panel). 
    % The mean represents the micro average. 
    (b) Out-of-domain generalization evaluation between IMT-CXR and other classification models (i.e., ConVIRT, GloRIA, BioViL, and MedKLIP) on the ChestX-ray14 dataset. AUC score and F1 are utilized to assess the classification task. The mean represents macro average. The error bars of our method show 95\% confidence intervals, and the bars’ center represents the mean value of the AUC or F1.
    All evaluations are conducted under direct inference setting. 
    % \textcolor{black}{It should be noted that "zero-shot" in this work refers to the model's direct inference capability using its pre-trained representations, without any fine-tuning on the target dataset, rather than generalization to unseen instructions or previously unseen disease categories.}
}
% \caption{
% (a) In-domain evaluation demonstrates the model's diagnostic capability on the MIMIC-CXR dataset, achieving robust classification performance across 26 thoracic diseases (micro-average metrics) and accurate attribute characterization of disease severity levels (e.g., mild/moderate/severe) with anatomical localization precision (e.g., left lower lobe). (b) Out-of-domain generalization analysis on ChestX-ray14 reveals superior performance of IMT-CXR compared to advanced specialist models (ConVIRT, GloRIA, BioViL, MedKLIP), with mean macro-average AUC scores of 0.89 (±0.03) and F1 scores of 0.82 (±0.05), where error bars denote 95\% confidence intervals. All evaluations were conducted under direct inference setting.}
	\label{fig:cls}
\end{figure}

% \subsection{Disease Classification}
% \subsection{Image Screening Stage}
\subsection{IMT-CXR enables accurate classification of disease entity and attribute}
% We explored two types of classification tasks: disease entity classification and attribute classification. The entity classification task focuses on classifying disease categories, while the attribute classification task determines the disease attributes, e.g., location and severity. The classification task could systematically screen disease entities at the beginning of AI-simulated workflow.

% Figure \ref{fig:cls}(a) shows the detailed distribution of F1 results for 26 different disease entities that are aligned with the long-tail distribution on the MIMIC-CXR dataset.
% Additionally, our model attains high F1 scores across the diseases in the attribute classification task. For instance, the severity classification ACC for Effusion and Pneumothorax reaches \textcolor{black}{60.3\% (95\% CI 59.1-61.5\%)} and \textcolor{black}{52.6\% (95\% CI 50.8-54.4\%)}, respectively.
\textcolor{black}{
Clinical decision-making in chest X-ray analysis requires dual-level assessment: first identifying disease existence (entity classification), then characterizing its clinical manifestations (attribute classification). Our AI-simulated workflow addresses this hierarchy through two coordinated tasks:
1) Disease entity classification: Detects 26 major cardiopulmonary pathologies.
2) Attribute classification: Determines critical clinical features, including anatomical location and severity grades.
Figure \ref{fig:cls}(a) demonstrates our model's capability to handle real-world data distributions, achieving robust F1 scores across the long-tail spectrum of MIMIC-CXR diseases. 
Our model attains high F1 scores across the diseases in the attribute classification task. For instance, the severity classification ACC reaches 61.8\% (95\% CI 60.5-63.1\%) for Effusion and 57.8\% (55.8-59.8\%) for Pneumothorax, respectively.}

When tested under domain-shifted conditions using ChestX-ray14 data, our method maintains diagnostic precision for subtle abnormalities (see Figure \ref{fig:cls}(b)).
For instance, our method achieves an AUC and F1 of 74.2\% (73.5-74.9\%) and 37.4\% (36.2-38.6\%) for Atelectasis and 69.6\% (68.6-70.6\%) and 20.0\% (18.6-21.6\%) for Nodule, respectively (see
Supplementary Table~\ref{tab:cls}).
% These results are comparable to the advanced specialist methods, demonstrating the effectiveness of our method in accurately identifying and localizing relatively small abnormalities in Chest X-ray images.  
% Additionally, the performance improvement of the proposed method on the disease classification is statistically significant compared to the advanced specialist models ( with \textit{p} < 0.05).
These results statistically outperform specialist models (\textit{p}<0.05), particularly in localizing subcentimeter lesions.
To verify clinical adaptability, we validate the framework on CheXpert and RSNA Pneumonia datasets through fine-tuning setting 
% (see 
% Supplementary Table \ref{tab:cls_fulldata})
and also obtain 
clinical-relevant metrics for classification tasks
(see 
Supplementary Tables \ref{tab:cls_fulldata}-\ref{tab:combined_metrics}).
% Supplementary Table \ref{tab:combined_metrics}.
% To assess the generalizability of our model, we extend the evaluation of disease entity classification to additional datasets (i.e., CheXpert and RSNA Pneumonia) under both direct inference (Supplementary Table~\ref{tab:cls}) and fine-tuning paradigms (Supplementary Table~\ref{tab:cls_fulldata}). The experimental results demonstrate that our model consistently achieves competitive performance across all three datasets in both evaluation scenarios.

% However, there is a noticeable performance gap when compared to the top pre-trained models, such as MRM\cite{zhou2023advancing}, during fine-tuning on the full dataset.

\begin{figure}[t]
	\centering
    \includegraphics[width=0.995\linewidth]{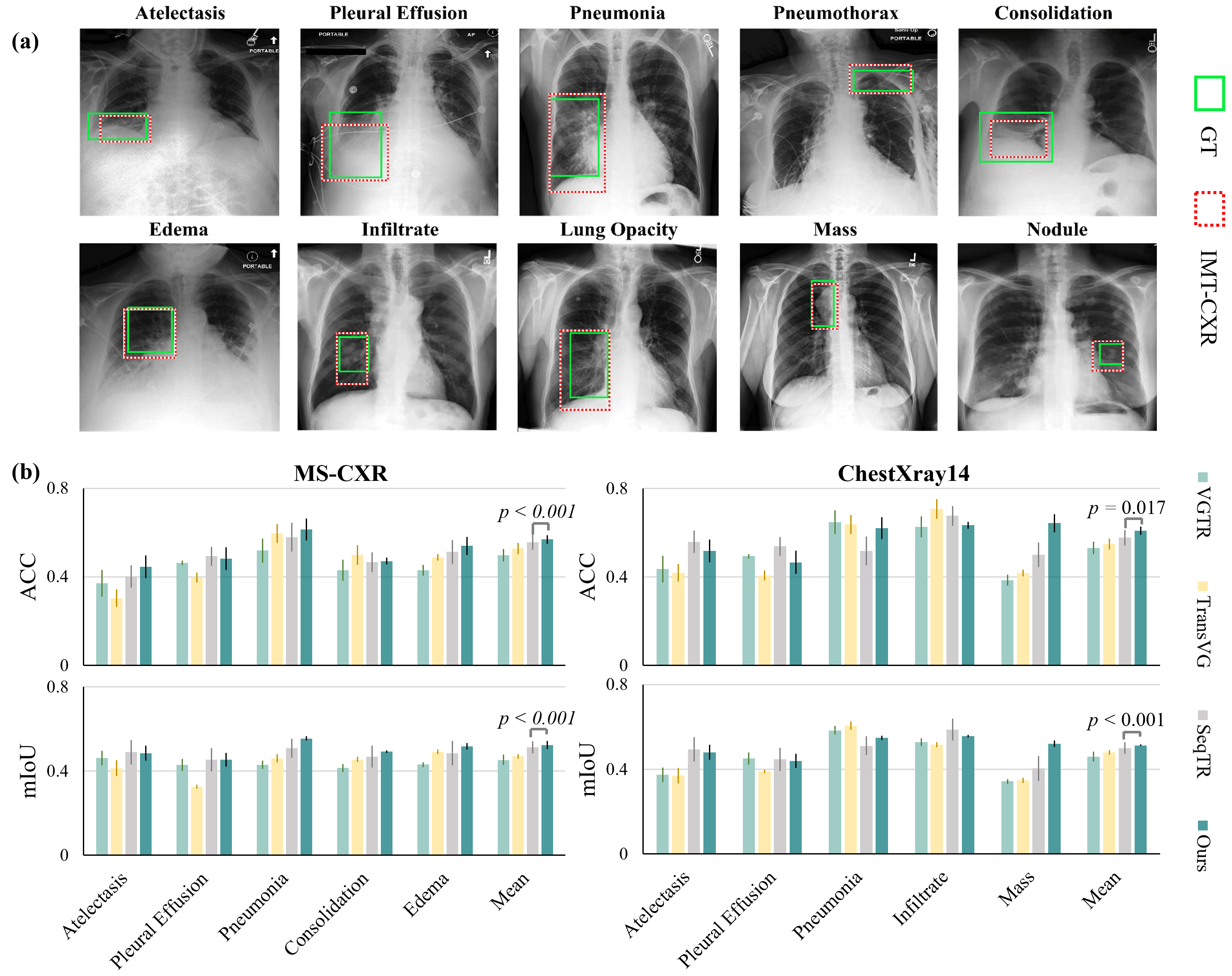}
	\caption{Assessment of the lesion localization task with ACC and mIoU metrics on the MS-CXR and ChestX-ray14 datasets with 20-shot fine-tuning setting. (a) Examples with bounding box prediction and (b) comparisons between IMT-CXR and other methods. Baseline methods incorporate location-aware queries for fair comparison \cite{chen2023medical,muller2025chex}. Error bars indicate 95\% confidence intervals and bar centers represent mean ACC/mIoU values.} 
	\label{fig:vg}
\end{figure}

\begin{figure}[t]
	\centering
    \includegraphics[width=0.995\linewidth]{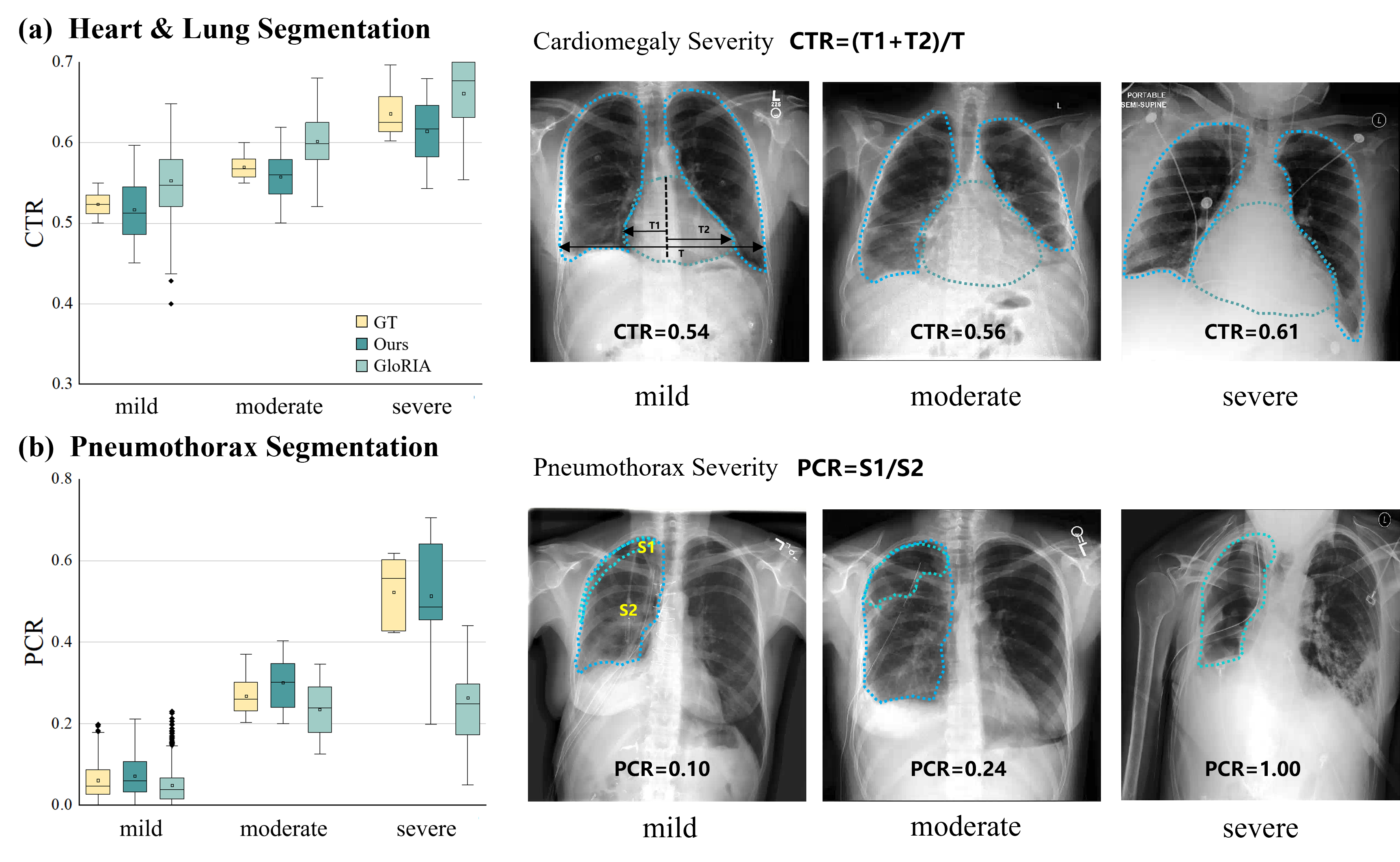}
	\caption{
 IMT-CXR segmentation performance for disease severity assessment \cite{fan2024deep} to quantify cardiomegaly severity. (b) Pneumothorax segmentation: Pneumothorax ratio (PCR) is derived from segmentation to quantify pneumothorax severity.
Left panels: Box plots compare CTR/PCR predictions (IMT-CXR vs.GloRIA) against ground-truth distributions across severity grades.}
	\label{fig:segmentation}
\end{figure}
 % Quantitative details are available in Supplementary Table~\ref{tab:cls}.
 
% \subsection{lesion localization}
% \subsection{Diagnostic Reasoning Stage}
\subsection{IMT-CXR supports detailed diagnosis by vision-intensive tasks}
\subsubsection{The localization task can accurately identify abnormalities}

As illustrated in Figure \ref{fig:workflow}(a), our AI-simulated workflow integrates:
1) Disease classification (e.g., Q: What disease does this image have? A: Pneumonia);
2) Attribute localization (e.g., Q: Where is Pneumonia? A: on the left lower side);
3) Visual grounding via natural language queries to pinpoint disease regions (e.g., Q: Give the accurate bounding box of Pneumonia? A: [x1,y1,x2,y2]) (Figure \ref{fig:overview}(d)).
We herein evaluate lesion localization performance on ChestX-ray14, RSNA Pneumonia, and MS-CXR datasets under direct inference and 20-shot fine-tuning settings. 
It should be noted that the model’s direct inference capability refers specifically to its use of pre-trained representations without any fine-tuning on the target dataset, rather than its ability to generalize to unseen instructions or previously unseen disease categories.

Among them, MS-CXR (derived from MIMIC-CXR) is the only visual grounding benchmark providing clinically annotated disease attributes (location, severity) extracted from radiology reports.
ChestX-ray14 and RSNA Pneumonia originally contain only disease category labels. To enable fair comparison, we extract anatomical location descriptors (e.g., left lower side) from their ground-truth bounding box coordinates, thereby aligning their annotation granularity with MS-CXR.

Under the direct inference setting, IMT-CXR achieves 
56.3\% (95\% CI 55.7-56.9\%), 42.5\% (41.7-43.3\%), and 46.7\% (44.9-48.5\%) accuracy on ChestX-ray14, RSNA Pneumonia, and MS-CXR, respectively.
% , outperforming existing visual grounding models.
When fine-tuned on the downstream dataset with 20 samples for each label, 
% accuracy improves to 60.9\% (59.7-62.1\%) on ChestX-ray14 and 54.2\% (53.1-55.3\%) on RSNA Pneumonia dataset.
accuracy improves to 60.9\% (59.7-62.1\%), 54.2\% (53.1-55.3\%) and 56.9\% (55.7-57.3\%), respectively.
Figure \ref{fig:vg} presents the model's capability of lesion localization across multiple diseases and demonstrates IMT-CXR’s superior lesion localization versus VGTR, TransVG, and SeqTR under 20-shot settings. 
% Notably:
% All baselines perform better on MS-CXR due to its attribute-enriched queries (e.g., positional cues), aligning with prior findings on phrase query limitations \cite{chen2023medical}.
Generally, our model demonstrated competitive performance on the test datasets compared to baseline models \cite{chen2023medical}.
% 55.4\% (95\% CI 53.7-57.3\%)} on the three datasets. 
% Additionally, our cross-task training supervision for each sample enhances performance in downstream tasks. 
% By incorporating additional disease attribute information through organized VQA tasks, our multi-task model achieves relative advancements, particularly in the direct inference setting. On the other hand, the differences in full fine-tuning results among methods become small. 
Further details of 20-shot and full-data finetune
results can be found in Supplementary Tables \ref{tab:VG}-\ref{tab:VG_fulldata}.

\subsubsection{The segmentation task can quantitatively evaluate the severity of cardiomegaly and pneumothorax} 
% In clinical practice, accurate assessment of thoracic structures is critical for diagnosing cardiopulmonary pathologies. Cardiothoracic ratio (CTR) and pneumothorax ratio (PCR) serve as key quantitative metrics. 
% CTR evaluates cardiomegaly severity through the ratio of heart to chest transverse diameters, while PCR quantifies pneumothorax extent via the affected lung area proportion. These measurements directly inform clinical decisions on disease staging (mild/moderate/severe) and treatment planning.
\textcolor{black}{
In clinical practice, precise evaluation of thoracic anatomy is fundamental for diagnosing cardiopulmonary pathologies. The cardiothoracic ratio (CTR) and pneumothorax ratio (PCR) serve as pivotal quantitative biomarkers: CTR quantifies cardiomegaly severity through the transverse cardiac diameter relative to thoracic cage width, while PCR measures pneumothorax extent via the proportion of collapsed lung area. These objective metrics critically inform clinical decision-making processes, including disease severity stratification (mild/moderate/severe classification) and therapeutic intervention planning, such as determining the need for chest tube placement or conservative management.}

% To assess the severity of cardiomegaly and pneumothorax in potentially more detailed analyses, we perform post-processing on the segmented lung and heart masks to calculate the cardiothoracic ratio (CTR) and pneumothorax ratio (PCR). 
% CTR is calculated as the ratio between the maximum transverse diameter of the heart and the chest, commonly used to evaluate the cardiomegaly severity as mild, moderate, or severe. The area method is used to calculate the PCR, which is represented by the ratio of the pneumothorax area to the affected lung area. 

To enable these analyses, we adopted a polygon-based contour representation for the segmentation task to achieve a uniform input-output format with other tasks, i.e., predicting a list of polygon vertexes instead of region masks. 
% The Dice coefficient is utilized for evaluating the segmentation of lung and heart contours.  
% For the segmentation of the heart and lung contours, both our unified model and the pixel-based segmentation method achieved Dice scores around 90\%. However, for the segmentation of pneumothorax, our method achieved better results with a significant improvement of 10\% in Dice score, both in direct inference and finetuning settings.
The segmentation performance is evaluated using Dice coefficients, with our unified model achieving comparable accuracy to pixel-based method \cite{Huang2021GLoRIAAM} (around 90\% Dice) for routine heart and lung contours of JSRT and CheXmask datasets. 
% Notably, our approach demonstrates superior pneumothorax segmentation with a 10\% Dice improvement across both direct inference and fine-tuning paradigms.
Post-processing of segmented masks enables precise anatomical quantification:
Figure \ref{fig:segmentation} illustrates the segmentation results of cardiomegaly and pneumothorax with different severity on the test splits of the CheXmask and SIIM dataset. We observe that the overall performance of lung and heart contours is satisfactory across different CTR (see Figure \ref{fig:segmentation}(a) box plot). GT achieves mean values of 0.52/0.57/0.63 on mild/moderate/severe CTR, while our model achieves 0.51/0.56/0.62 and GloRIA achieves 0.55/0.60/0.68. 
Furthermore, pneumothorax exhibits significant variations in location and size, and our model exhibits higher accuracy in predicting severe pneumothorax compared to GloRIA (see Figure \ref{fig:segmentation}(b) box plot). GT achieves mean values of 0.05/0.26/0.56 on mild/moderate/severe PCR, while our model achieves 0.06/0.30/0.49 and GloRIA achieves 0.04/0.24/0.25. 
% For more accurate segmentation of mild pneumothorax, we incorporate a segmentation head with a U-Net decoder and achieve a Dice value of 59.6\% (95\% CI 57.5-61.7\%) and 66.2\% (64.9-67.5\%) under direct inference and fine-tune settings. 

\begin{figure}[t]
	\centering
	%\fbox{\rule[-.5cm]{4cm}{4cm} \rule[-.5cm]{4cm}{0cm}}
    \includegraphics[width=0.99\linewidth]{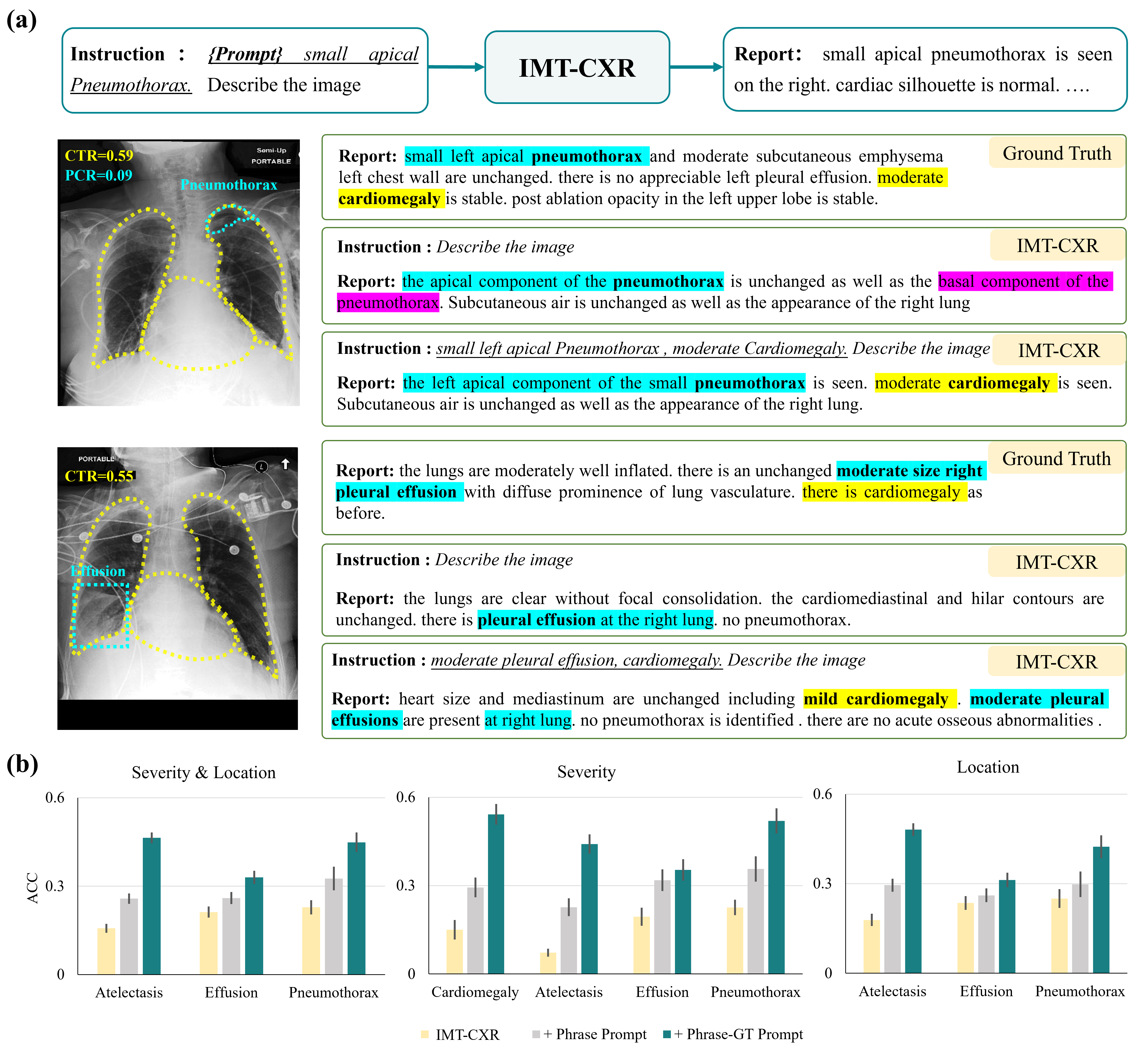}
	\caption{
    % (a) Comparison of generated reports with and without customized instructions, illustrated through two examples: Pneumothorax and Pleural Effusion. The customized instructions encompass the disease category, location, and severity level, all inferred from related tasks. In the highlighted regions of the reports, light blue and yellow represent correct predictions, while magenta represents incorrect predictions.
    % (b) Comparison of generated reports in three ways: without prompt, with prompt from multi-task results, and with prompt from ground truth (GT). We use ACC as the evaluation metric. The error bars show 95\% confidence intervals, and the bars’ center represents the mean value of the ACC.
(a) Impact of task-specific prompts on report generation.
Case studies: pneumothorax and pleural effusion reports generated with vs. without clinical attribute prompts (category, location, severity).
Highlighted regions: light blue/yellow = correct predictions (e.g., 'small left apical pneumothorax'); magenta = errors/missing attributes (e.g., severity omitted in baseline reports).
(b) Quantitative comparison of prompting strategies across three settings:
No prompt,
Multi-task-derived prompts,
Ground-truth (GT) prompts.
We use ACC as the evaluation metric.
Statistical representation: Bar centers = mean ACC; error bars = 95\% confidence intervals.
    }
	\label{fig:prompt}
\end{figure}

\subsection{IMT-CXR enhances the interpretability of report generation}
% As the summary of the radiology reading process, reports contain major findings and possible disease diagnoses from the radiologists. As the unique feature of our proposed framework, we hypothesize adding pre-generated evidence could significantly improve the quality of AI-generated reports. Figure \ref{fig:prompt} demonstrates this quality improvement with the disease attributes prompt. 
% During inference, the customized prompts (e.g., entity, severity, approximate location) are initially derived from the classification task and will be updated when a more accurate bounding box or severity of Pneumothorax/Cardiomegaly is available from the lesion localization or segmentation task.  
As a summary of the radiology reading process, 
reports systematically synthesize critical findings and diagnostic hypotheses, a process our framework enhances through evidence-driven dynamic prompting. As illustrated in Figure \ref{fig:prompt}, the proposed method integrates pre-generated clinical attributes (e.g., disease entity, severity, approximate location) as initial prompts derived from classification tasks. These prompts are iteratively refined using lesion localization and segmentation outputs, such as precise pneumothorax bounding boxes or cardiomegaly CTR metrics, to ensure anatomical and diagnostic consistency.

% The generated reports of typical cases of Pneumothorax and Pleural Effusion are provided in Figure~\ref{fig:prompt}(a). 
% The proposed model is capable of identifying the diseases with an accurate lesion localization box and generated report.For pneumothorax, 
For typical Pneumothorax case (see Figure~\ref{fig:prompt}(a)), the model generates reports with detailed descriptors (e.g., "small left apical pneumothorax"), validated against PCR metrics and pneumothorax mask, whereas non-prompted models omit critical attributes (e.g., "small" of pneumothorax, "moderate cardiomegaly").
% Detailed descriptions of pneumothorax entities and attributes such as "small left apical pneumothorax" are highlighted by blue text in the ground truth report and are well predicted in the generated report. The location and severity of the pneumothorax are further verified by the bounding boxes and calculated metric (i.e., PCR). 
% On the other hand, the descriptions of disease attributes (e.g., "small" of pneumothorax, "moderate cardiomegaly") are omitted in the generated report without proper prompt. 
% For the Effusion case, we also find the importance of proper input prompts in the task of report generation.
% The descriptions of pleural effusion and cardiomegaly are more detailed and accurate in the generated report with prompt. The bounding box of pleural effusion and mild cardiomegaly is indicated by the lesion localization and segmentation task (i.e., CTR=0.55). With more specific prompts, the generated report shows accurate descriptions as "mild cardiomegaly . moderate pleural effusions are present at right lung". In contrast, the general model without the designed prompt fails to recognize cardiomegaly.
Similarly, in Pleural Effusion case, prompted reports accurately describe "mild cardiomegaly . moderate pleural effusions are present at right lung", while the baseline model without the designed prompt fails to recognize cardiomegaly.

Figure \ref{fig:prompt}(b) further compares the accuracy of three different conditions (i.e., baseline, with phrase prompt, and with phrase-GT prompt).
With the help of disease attributes prompt, the accuracy of severity and location description has been improved by 10.1\%, 4.8\%, 9.8\% for Atelectasis, Effusion, and Pneumothorax, respectively. The accuracy of Cardiomegaly in severity description is improved by 14.4\%.
% (Cardiomegaly does not have a location attribute).
When the ground truth disease phrase is utilized as the prompt, we find the quality of the report improved further as an upbound for our model.
Ablation experiment of multi-task and comparison with other methods can be found in Supplementary Table \ref{tab:ablation}.

To further validate the causal role of intermediate predictions, we compared report outputs using correct intermediate prompts versus deliberately wrong severity/location prompts (see Supplementary Figure \ref{fig:wrongprompt}).
Our results show that incorrect prompts consistently reduce the factual accuracy of the generated attribute descriptions, confirming that the generator relies on intermediate attributes as effective evidence and that erroneous attributes can adversely affect report correctness.
Furthermore, we visualized cross-attention maps between key text tokens (e.g., disease names and severity descriptors) and their corresponding image feature maps (see Supplementary Figure \ref{fig:grad}). The gradient-based class activation mapping (Grad-CAM) visualizations indicate that the regions with the highest response largely overlap with the clinically relevant areas described in the report, particularly the right lower lung, where the reported abnormalities are located. This suggests that the model predictions are primarily driven by disease-related visual evidence rather than instruction patterns alone. However, we also observe several limitations: non-pathological regions, such as parts of the abdomen and lower neck, exhibit secondary activations, indicating that the attention is not perfectly localized and may still be influenced by surrounding structures.

Finally, we compared IMT-CXR against traditional methods (Supplementary Table~\ref{tab:report}) and recent multimodal large language models (LLMs) optimized for radiology report generation (Supplementary Table~\ref{tab:report2}). The results demonstrate that, despite using a significantly smaller language model and prioritizing interpretability, IMT-CXR remains competitive with these task-specific, high-parameter models in report generation performance.

\textbf{Multi-center Radiologist Evaluation}
% \subsection{Radiologist evaluation}
To assess the clinical interpretation, three radiologists, ranging from junior or senior levels are invited to examine the quality of radiology reports in a double-blind manner. 
The side-by-side comparison study focuses on errors associated with the presence, location, and severity of clinical findings. Non-clinical errors, such as referring to views or previous studies that do not exist, are excluded from our evaluation.
% In parallel evaluation, 

\begin{figure}[t]
	\centering
    \includegraphics[width=1.0\linewidth]{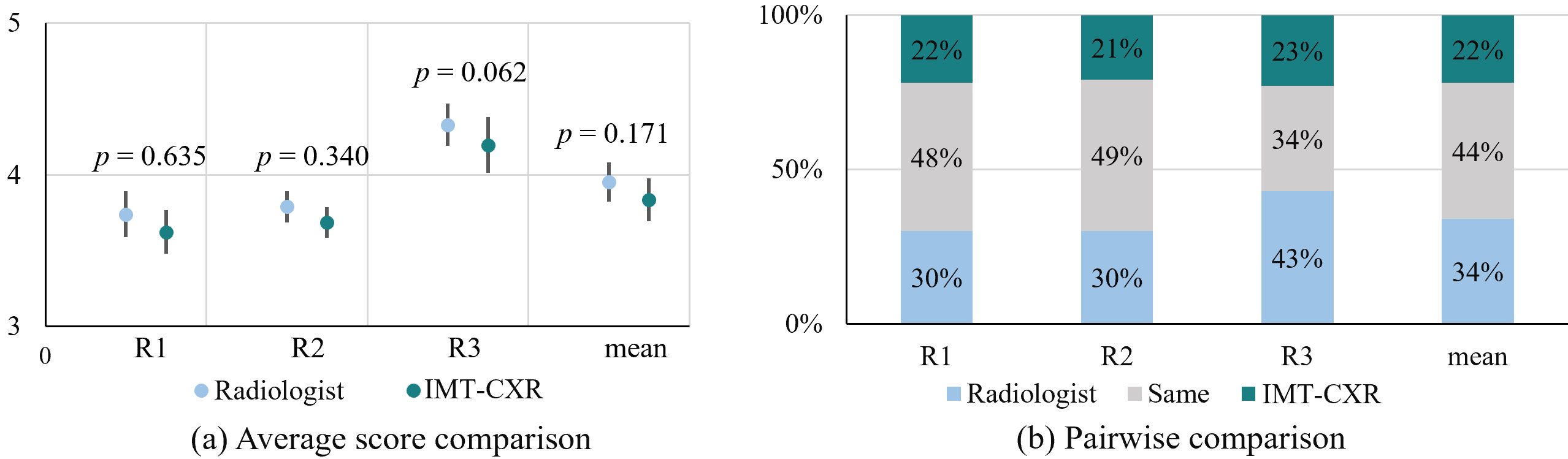}
	\caption{\textbf{Side-by-side comparison}. Three radiologists reviewed and scored the clinically derived reports from four centers and reports generated by IMT-CXR. The error bars show 95\% confidence intervals, and the bars’ center represents the mean value of the average scores.}
	\label{fig:sbs-evaluation}
\end{figure}

% In the side-by-side comparison,
% Figure \ref{fig:evaluation}(a) presents the average score comparison result by three radiologists. Three radiologists provided average results for the original reports and generated reports in four centers: 4.01/3.61, 4.03/3.84, 4.08/3.63, and 3.76/3.68. The average result across the four centers is 3.97/3.69, indicating that the quality of the generated reports is comparable to the original doctor reports. Figure \ref{fig:evaluation}(b) presents the pairwise comparison result, radiologists believe that 54\%/64\%/54\%/62\% of the generated reports are of equal or better quality than the original doctor reports at four centers, respectively. Despite the observed fluctuations in results across different centers, the collective analysis suggests that our generated reports exhibit a comparable level of quality to the original reports generated by medical professionals.
Figure \ref{fig:sbs-evaluation}(a) presents the average score comparison result from three radiologists. The average ratings for the radiologist-provided reference reports and generated reports are 3.74 \textcolor{black}{(95\% CI 3.62-3.86) / 3.62 (3.48-3.76)}, 3.79 \textcolor{black}{(3.69-3.89) / 3.69 (3.59-3.79)}, and 4.33 \textcolor{black}{(4.19-4.47) / 4.20 (4.02-4.38)}, respectively. The average result across three radiologists is 3.95 \textcolor{black}{(3.82-4.08) / 3.84 (3.74-3.94)}. \textcolor{black}{We utilize the Chi-Square test to verify the significance of performance differences between the model-generated reports and the original ones across different data centers. 
% The p value of radiologists 1, 2, and 3 are 0.645. 0.340, 0.031, respectively. 
With the computed \textit{p} value of three tests (compared to the original reports by three doctors) are 0.635, 0.340, and 0.062, respectively, it indicates that the generated reports are comparable to the reference reports in overall quality.} Figure \ref{fig:sbs-evaluation}(b) presents the pairwise comparison result. Three radiologists believe that the quality of generated reports is equal to or even better than the original reports by \textcolor{black}{70\%, 70\%, and 57\%}, respectively. Despite the observed fluctuations in results across radiologists, the collective analysis suggests that our generated reports exhibit a comparable level of quality to the original reports generated by medical professionals.

\begin{figure}[t]
	\centering
    \includegraphics[width=1.0\linewidth]{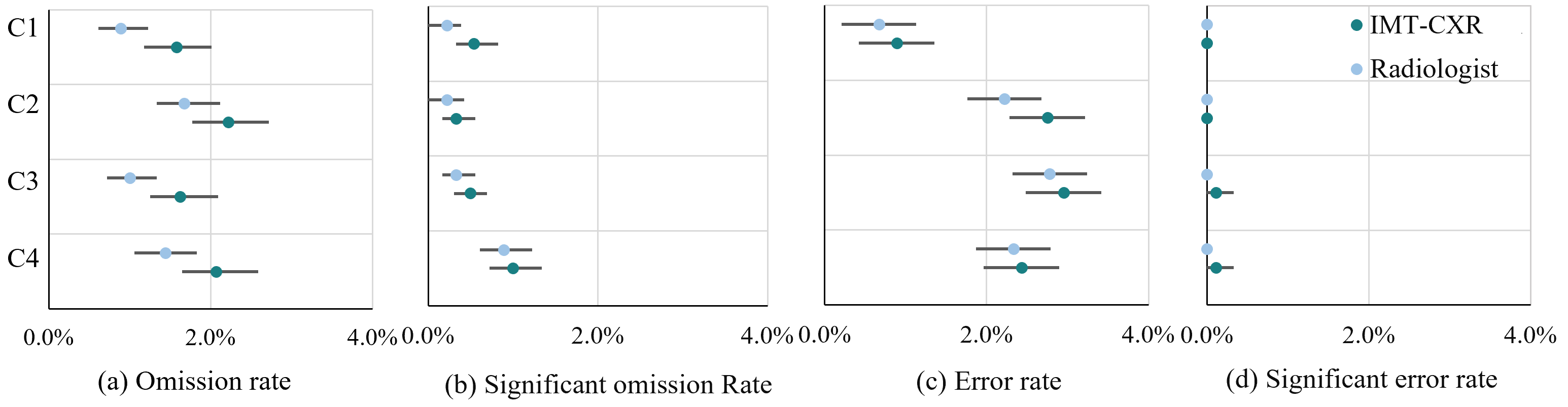}
	\caption{\textbf{Independent evaluation} of omissions and clinical errors for reports generated by IMT-CXR. Significant errors are related to the presence, location, or severity of clinical findings, which are identified by radiologists. The error bars show 95\% confidence intervals, and the bars’ center represents the mean value.}
	\label{fig:ie-evaluation}
\end{figure}

In the detailed evaluation, we present the omission and error rates, where omission rate indicates missed disease diagnoses, and error rate reflects inaccuracies in severity or location descriptions or false diagnoses.
We report the results on the report level. Figure \ref{fig:ie-evaluation}(a) shows the total omission rate for the reference and generated reports in the four centers are as follows: 0.89\% \textcolor{black}{(0.33-1.56\%) / 1.58\% (0.78-2.44\%)}, 1.67\% \textcolor{black}{(1.00-2.56\%) / 2.22\% (1.33-3.22\%)}, 1.00\% \textcolor{black}{(0.44-1.67\%) / 1.62\% (0.89-2.56\%)}, 1.44\% \textcolor{black}{(0.67-2.22\%) / 2.07\% (1.22-3.11\%)}. 
% On average, the omission rate for the generated reports 2.53\% is close to that of the radiologist-provided reference reports, i.e., 1.25\%, with a significant omission rate of 0.42\%/1.00\%. 
Figure \ref{fig:ie-evaluation}(c) shows that
% the error rate, i.e., inaccuracies in severity or location descriptions or false positives. 
the total error rate for the reference and generated reports in the four centers are 0.67\% \textcolor{black}{(0.22-1.22\%) / 0.89\% (0.33-1.56\%)}, 2.22\% \textcolor{black}{(1.33-3.22\%) / 2.75\% (1.89-3.89\%)}, 2.78\% \textcolor{black}{ (1.78-3.89\%) / 2.95\% (1.89-4.11\%)}, 2.33\% \textcolor{black}{ (1.44-3.22\%) / 2.36\% (1.45-3.24\%)}. \textcolor{black}{Based on the results, the omission rate and error rate observed in IMT-CXR reports are comparable to those original ones from three centers.}
% Three radiologists ranks the quality of the reports in a side-by-side evaluation, comparing the radiologist-provided reference report from MIMIC-CXR with reports generated by different. As shown in Figure \ref{fig:evaluation}(a), the generated report is preferred over the reference report in half of cases. On the other hand, three radiologists identifies therates of omissions and clinical errors in generated reports. Clinical errors are those related to the presence, location or severity of a clinical finding. 

 \section{Discussion}

Our study presents a unified framework that integrates four critical chest X-ray analysis tasks, i.e., disease classification, lesion localization, anatomical segmentation, and radiology report generation, through a multi-task instruction-tuning paradigm. This integration enables synergistic learning across tasks, enhancing clinical interpretability through evidence-based hierarchical reasoning. The model demonstrates robust direct inference capabilities, achieving competitive diagnostic accuracy while maintaining computational efficiency.

The framework's clinical utility stems from its ability to provide traceable diagnostic pathways. 
It is crucial that automated medical report generators produce trustworthy and accurate reports for effective utilization in practice. 
The interpretability of the reports can be validated mutually with the results obtained from other tasks within the model. For instance,
disease entities, severity levels, and anatomical locations identified in classification tasks are cross-validated with localization bounding boxes and segmentation-derived quantitative metrics. For pneumothorax and cardiomegaly, the segmentation function quantifies disease severity by analyzing anatomical contours (e.g., pneumothorax masks or cardiac boundaries). This multi-task verification mechanism ensures report consistency, as evidenced by 66\% of AI-generated reports being rated equal or superior to radiologist-authored counterparts in blinded evaluations. Customized clinical prompts further enhance diagnostic precision, improving severity and location description accuracy by approximately 10\% for atelectasis and pneumothorax compared to baseline methods.

Image-instruction-label triplet dataset is designed for promoting multi-task learning. 
While there exist various single-task datasets, there have been limited attempts to unify them and create benchmarks for the development of a single and more comprehensive model. As one of our major contributions, we designed and released a comprehensive dataset of chest X-ray data. This dataset includes full-label annotations, enabling researchers to explore and leverage the benefits of multi-task learning in this domain. By sharing this dataset, we aim to encourage further advancements in multi-task learning for chest X-ray analysis. This can potentially create new opportunities in clinical applications \cite{johnson2019mimic,irvin2019chexpert,wang2017chestx}.
Our model demonstrates robust performance across a variety of tasks, with notable improvements in the explainability of the generated outputs. Recent advancements \cite{bannur2024maira,chavesllava,lee2025cxr,chen2024chexagent,bai2025deepseek,zhou2024generalist}, particularly the integration of large language models with additional input data for chest radiograph interpretation, have shown potential in improving generalization capabilities. Future work will further explore stronger medical-domain language priors, improved robustness under distribution shifts, and more complex instruction settings. We are also addressing minor task prediction inconsistencies by augmenting training data diversity, implementing regularization techniques, and refining model architecture to ensure cross-task stability. These efforts aim to enhance model robustness for broader clinical applications.

Beyond diagnostic accuracy, the practical deployment of an AI system is critical for clinical translation.
Our unified multimodal model contains approximately $0.9$B parameters, with FP16 weights occupying about $1.8$\,GB (see Supplementary Table \ref{tab:dep}). 
During inference on an NVIDIA V100 GPU (32\,GB), the peak GPU memory usage typically ranges from $3.5$ to $5$\,GB. A single forward pass requires only one image encoding, followed by the serial execution of classification, localization (requiring $\sim$4 tokens), segmentation ($\sim$30 tokens), and report generation ($\leq 50$ tokens). The total generated text length is approximately $90$ tokens, resulting in an end-to-end latency of about $3$--$5$ seconds per case.
We compared this with a task specific baseline system composed of separate models, for example one ResNet-152 per task together with an independent report generation model. Although the individual networks in this baseline are lighter, with total weights of about $0.5$--$0.8$\,GB, the system requires sequential calls to multiple models. This leads to a comparable end-to-end latency of $1.5$--$4$ seconds and introduces additional overhead in model scheduling and maintenance within a practical deployment pipeline.

IMT-CXR demonstrates transformative potential for streamlining chest X-ray diagnostics through its unified multi-task capabilities, offering abnormality detection, anatomical quantification, and evidence-linked reporting. The framework could optimize workflows in high-volume radiology departments by automating preliminary assessments while maintaining radiologist oversight, and serve as a force multiplier in resource-limited regions through integrated screening-triage-reporting functionalities. 
While the current unified architecture demonstrates strong performance across tasks, its multimodal fusion mechanism can be further refined. Future work will explore advanced fusion strategies tailored for dense vision-language alignment \cite{young2026xrayclaw,chen2026tc}, to enhance the model's capacity in processing fine-grained visual and textual information. Beyond multi-task integration, an important future direction is the development of more explicit visual reasoning and grounding mechanisms. The current framework primarily relies on sequential information transfer, and more structured reasoning paradigms remain to be explored. To enable deeper clinical reasoning, future iterations of IMT-CXR will investigate the incorporation of explicit reasoning modules, including neuro-symbolic approaches that construct structured graphs (e.g., 〈disease, attribute, location〉) from intermediate predictions and enforce logical consistency during report generation. The integration of a dedicated relational encoder to model dependencies between clinical findings prior to synthesis is another promising direction to enhance diagnostic coherence.

Future iterations should also expand clinical adaptability by incorporating rare pathology recognition and complex scenario handling (e.g., postoperative changes), while strengthening human-AI collaboration through radiologist-guided prompt engineering and PACS workflow integration. By evolving into a collaborative diagnostic assistant that augments rather than replaces clinical expertise, this approach lays the foundation for human–AI symbiotic systems in medical imaging.
Safe integration into clinical workflows remains essential for real-world deployment. This requires explicitly acknowledging and mitigating the potential risk of error propagation from upstream visual analysis tasks to the final generated report. Based on the multi-center blinded evaluation, we observed that evidence-induced discrepancies generally fall into two categories. Severe diagnostic errors were relatively infrequent and were primarily associated with incorrect disease classification (e.g., missed pneumothorax or false-positive mass detection), which directly correspond to the omission and false-positive cases reported in the radiologist evaluation. In contrast, the majority of discrepancies were minor descriptive deviations related to localization or segmentation outputs, such as slightly imprecise anatomical location descriptions or small differences in quantitative measurements (e.g., cardiothoracic ratio). Radiologists indicated that these differences rarely altered the primary diagnostic conclusion and were typically corrected during routine review. These observations define the practical safety boundary of the proposed system: IMT-CXR is intended to function as a decision-support assistant rather than an autonomous reporting system, with radiologists validating intermediate evidence before final report confirmation. The explicit multi-task evidence structure facilitates this workflow by making intermediate predictions (e.g., bounding boxes, segmentation masks, or quantitative metrics) transparent and traceable prior to report synthesis. To further mitigate potential risks in real-world deployment, future iterations will incorporate additional technical safeguards, including confidence thresholding for upstream predictions, automatic uncertainty annotations for low-confidence findings in generated reports, and explicit prompts for manual review when high-risk conditions are detected with marginal certainty.

\paragraph{Limitations of Study}

First, the retrospective multi-center evaluation was designed as an initial feasibility assessment rather than a definitive clinical validation. The sample size (160 cases), while sufficient for common conditions, is insufficient to adequately cover rare pathologies. In addition, the evaluation was not conducted with pre-specified endpoint definitions, which may introduce ambiguity, particularly in distinguishing clinically relevant errors from non-clinical hallucinations.
Second, the study cohort was derived from a combination of public datasets and retrospective clinical data. Participant-level demographic variables, including sex, gender, age, race or ancestry, ethnicity, and socioeconomic status, were not consistently available and therefore not incorporated into the analysis, which may limit the representativeness of the findings.
Third, the heterogeneous annotation structure constrains the optimization of the unified model. Most training data are aggregated from single-task datasets, with limited samples containing comprehensive annotations across all tasks. This restricts the effectiveness of joint supervision and may introduce task imbalance or negative transfer.
Fourth, the current framework relies on a general-domain pretrained model without explicit medical-domain pretraining or domain adaptation strategies, which may limit robustness under distribution shifts, such as variations in institutional reporting styles or rare pathologies.
Fifth, the current evidence-integration paradigm primarily relies on sequential information transfer by using intermediate task outputs as prompts. While this improves traceability, it does not fully constitute explicit visual reasoning or enforce strict cross-modal grounding, leaving room for potential bias from language priors.
Finally, although the unified framework improves interpretability and system integration, minor inconsistencies across tasks and potential error propagation from upstream predictions to report generation may still occur, highlighting the need for additional safeguards and more rigorous clinical validation in future work.

\section{STAR$\star$METHODS}
\subsection{Key resources table}

\begin{table}[htbp]
\centering
\caption{Key Resources Table}
\begin{tabular}{@{}lll@{}}
\toprule
\textbf{REAGENT or RESOURCE} & \textbf{SOURCE} & \textbf{IDENTIFIER} \\

\midrule
\multicolumn{3}{l}{\textbf{Deposited data}} \\
\midrule
% IMT-CXR dataset (new) & This paper & \url{https://huggingface.co/datasets/MedHK23/IMT-CXR} \\
MIMIC-CXR & PhysioNet & \url{https://physionet.org/content/mimic-cxr/2.0.0/} \\
VinDR-CXR & Kaggle & \url{https://vindr.ai/datasets/cxr} \\
ChestX-Det & Deepwise-AILab & \url{https://github.com/Deepwise-AILab/ChestX-Det-Dataset} \\
% ChestX-ray14 & NIH & \url{https://nihcc.app.box.com/v/ChestXray-NIHCC} \\
% CheXpert & Stanford ML Group & \url{https://stanfordmlgroup.github.io/competitions/chexpert/} \\
% TBX11K & Kaggle & \url{https://www.kaggle.com/datasets/vbookshelf/tbx11k-simplified} \\
% object-CXR & hlk-1135 & \url{https://github.com/hlk-1135/object-CXR} \\
JSRT Database & JSRT & \url{http://db.jsrt.or.jp/eng.php} \\
% Shenzhen chest X-ray Set & NIH & \url{https://www.ncbi.nlm.nih.gov/pmc/articles/PMC4256233/} \\
% Montgomery County chest X-ray Set & NIH & \url{https://www.ncbi.nlm.nih.gov/pmc/articles/PMC4256233/} \\
CheXmask & NIH  & https://github.com/ngaggion/CheXmask-Database \\
SIIM-ACR & SIIM & https://www.kaggle.com/c/siim-acr-pneumothorax-segmentation \\
% In-house Dataset & This paper & 2,531 clinical radiographs with BBox annotations across nine thoracic pathologies \\
IMT-CXR dataset & This study & \url{https://huggingface.co/datasets/MedHK23/IMT-CXR} \\
\midrule
\multicolumn{3}{l}{\textbf{Software and algorithms}} \\

Training and evaluation & This study & \url{https://github.com/MedHK23/IMT-CXR} \\

% DICOM Anonymizer tool & RSNA & \url{http://mirc.rsna.org/download} \\
% Python & Python Software Foundation & Version 3.8.18 \\
PyTorch & Meta & Version 1.11.0 (CUDA 11.8) \\
% Pillow & Python Imaging Library & Version 10.0.1 \\
% timm & Hugging Face & Version 0.9.2 \\
Torchvision & Meta & Version 0.12.0 \\
Opencv-python & OpenCV & Version 4.6.0 \\
% pycocotools & COCO consortium & Version 2.0.7 \\
% pycocoevalcap & COCO consortium & Version 1.2 \\
% torchmetrics & PyTorch Lightning & Version 1.0.1 \\
% NumPy & NumPy & Version 1.21.6 \\
% Pandas & Pandas & Version 2.0.3 \\

\bottomrule
\end{tabular}
\label{tab:key_resources}
\end{table}

\subsection{Resource availability}
\subsubsection*{Lead contact}

Further information and requests should be directed to the lead contact, Lijian Xu (xulijian@suat-sz.edu.cn).

\subsubsection*{Materials availability}
This study did not generate new unique reagents.

\subsubsection*{Data and code availability}
Code for training and evaluation is available at \url{https://github.com/MedHK23/IMT-CXR}.
The new dataset released in this study can be found at \url{https://huggingface.co/datasets/MedHK23/IMT-CXR}.
The MultiMedBench is all open source, and the respective download link is described in Git Hub.

\subsection{Experimental models and participant details}
% \subsubsection{Human participants and samples}
This study is based on a combination of publicly available datasets and retrospective multi-center clinical data. Participant-level demographic variables, including sex, gender, age, race or ancestry, ethnicity, and socioeconomic status, were not consistently available across the included datasets and were therefore not incorporated into the analysis.

All data were originally collected as part of routine clinical care, and demographic information, when available, was derived from existing medical records rather than prospectively collected for this study. For the proprietary datasets used in this work, the data were fully anonymized prior to analysis, and no additional demographic information was accessible.

As this study relies on historical datasets aggregated from multiple sources, the available variables are limited to those defined at the time of original data collection, and no further participant-level attributes were collected or inferred.

\subsection{Method details}
\subsubsection{Building Dataset for Customized Instruction Tuning} 
We constructed a multi-task dataset for joint training of disease classification, localization, segmentation, and report generation. In general, we unify the input and output labels of all sub-tasks into a uniform format for consistent modeling and joint training, i.e., a set of instruction-label samples are shown in \textcolor{black}{Figure~\ref{fig:overview}(d)} and more in the supplementary materials. We further built a subset including the attributes and phrases for chest X-ray images like "small base effusion, normal cardiac silhouette," which can be used as instruction for the report generation task. Additionally, the dataset underwent quality assurance by radiologists to ensure its accuracy and reliability. 

\paragraph{Instruction Design}
To build and utilize multiple instruction sets (for each of four sub-tasks) during the joint training approach, we design a set of seed instructions with placeholders (later replaced with corresponding targets) to create diverse related task descriptions for coarse-grained task-level customization. Following various instructions, our model can elegantly switch among different vision-centric tasks and accomplish them in a unified manner. More details about the organization of instructions for task-level customization, including disease classification, localization, segmentation, and report generation, are introduced in the supplementary material. 
We trained the transformer model using six publicly available chest X-ray datasets: MIMIC-CXR \cite{johnson2019mimic}, VinDr-CXR \cite{nguyen2022vindr}, ChestX-Det \cite{lian2021structure}, CheXmask \cite{gaggion2024chexmask}, SIIM-ACR \cite{siim}, and an in-house clinical dataset. Through rigorous data harmonization, we aligned lesion taxonomies across datasets and integrated three complementary annotation types: radiology reports, bounding boxes (BBox), and pixel-level masks. To prevent data contamination, we excluded any cases overlapping with downstream test/validation sets. 
Population characteristics of our training and validation dataset could be found in Supplementary Table~\ref{tab:demography}.
All datasets followed official patient-wise splits where available; others were partitioned randomly (7:1:2 train/validation/test ratio) as detailed in Supplementary Table~\ref{tab:dataset_overview}.
\leavevmode  % 添加这一行

1) \textbf{MIMIC-CXR} \cite{johnson2019mimic}: 377,110+ radiographs from 227,835 clinical studies, this dataset provides multi-label pathology classifications paired with radiology reports. We utilized it for disease classification and report generation tasks.

2) \textbf{VinDr-CXR} \cite{nguyen2022vindr}: 15,000 scans annotated with 28 thoracic pathologies. We selected eight diseases with verified BBox annotations for lesion localization training.

3) \textbf{ChestX-Det} \cite{lian2021structure}: Derived from NIH ChestX-ray14 \cite{wang2017chestx}, this subset includes 3,578 images with BBox annotations for 13 pathologies. Seven diseases meeting quality thresholds were used for localization training.

4) \textbf{CheXmask} \cite{gaggion2024chexmask}: Aggregates 676,803 lung/heart segmentation masks from six public databases. We trained on 224,316 cases and evaluated on 10,000 ChestX-ray14-derived masks.

5) \textbf{SIIM-ACR} \cite{siim}:  12,090 images (3,000 pneumothorax-positive) with segmentation masks from the SIIM Pneumothorax Challenge.

6) \textbf{In-house Dataset}: 2,531 clinical radiographs with BBox annotations across nine thoracic pathologies for lesion localization validation.

\subsubsection{Model Architecture \textcolor{black}{and Training Settings}}

 % The attributes of the diseases  are extracted from the ground truth reports and used as customized instruction for report generation. Cardiotoracic Ratio (CTR) and Pneumothorax Compress Ratio (PCR) is obtained by post-processing the results of cardiac, lung and pneumothorax segmentation.
In this work, we have proposed a multimodal 
% multi-task 
model that leverages the 
% widely-used 
sequence-to-sequence learning paradigm for joint multi-task training. The specific tasks encompass disease classification, localization, segmentation, and report generation. For each task, we design specific task instructions to facilitate the model's differentiation between tasks. 
Inspired by the multi-modal models \cite{wang2022ofa,chenpix2seq,lu2022unified},
% \cite{wang2022ofa,chenpix2seq,chen2022unified,lu2022unified},
IMT-CXR leverages an encoder-decoder architecture for perceiving pixel inputs and generating the target sequence and performs unified modeling and joint training on downstream visual and language tasks as shown in Supplementary Figure \ref{fig:method}. Bounding boxes and class labels are converted into sequences of discrete tokens. This enables IMT-CXR to robustly perform diverse language and vision tasks based on instructions, providing diverse and complex output results. 
% Benefiting from pre-trained language models and mutual guidance between tasks, IMT-CXR can engage in continuous question-answering and provide visual explanations of answers during conversations, which is particularly crucial in safety-critical domains like medical diagnosis. 

% \subsection{Model Architecture}
\textbf{Image and Language Encoder} With an input chest X-ray image $ x_{i}  \in  \mathbb{R}^{H\times W} $, visual features are extracted by image encoder and further projected to the feature space:
\begin{equation}
    v_{i} = P_{img}(E_{img}(x_{i})) \in \mathbb{R}^{(h_{f} \times w_{f}) \times d}
\end{equation}
where $h_{f}$ and $w_{f}$ are the output size of visual features, and $d$ represents the feature dimension. $E_{img}$ can be any common visual backbones and we use ResNet152 in this study,\textcolor{black}{ which is initialized from standard ImageNet-pretrained weights.} Specifically, we take output features from the 4th residual block. Visual features are then projected to a pre-defined feature dimension by using $P_{img}$, which is composed of two linear layers.
% \textbf{Language and encoder } 
With any processed input instruction sequence $t_{i}$, text features are extraced by language encoder:
\begin{equation}
    l_{i} = E_{txt}(t_{i}) \in \mathbb{R}^{n_{t} \times d}
\end{equation}
where $n_{t}$ is the number of input tokens and $d$ represents the feature dimension. 
% Bert \cite{devlin2018bert} is used as a language encoder.
\textcolor{black}{We employ BERT \cite{devlin2018bert} as the text encoder, initialized from its publicly available pretrained checkpoint.} 

\textbf{Multi-modality Module} This module follows an encoder-decoder architecture format. Given the input visual features $v_{i}$ and text features $l_{i}$, we first generate fused multi-modal representations by combining the image and text embeddings. These fused features serve as the keys and values in the cross-attention blocks in the decoder. By conditioning on the partial sequence $y_{i, <j}$ predicted so far, the decoder recursively makes predictions for the token at position $j$, effectively generating aligned descriptions across modalities. 
% The fused multi-modal embeddings capture relationships between the input modalities, guiding the decoder toward generating coherent descriptions grounded in both the image and text content.
\begin{equation}
    y_{i, j} = D_{mm}(E_{mm}(concat(v_{i}, l_{i})), y_{i, <j}) \in \mathbb{R}^{1 \times d}
\end{equation}
In our experiments, we leverage BART \cite{lewis2019bart} for multi-modal encoding and decoding. BART utilizes BERT \cite{devlin2018bert} as the encoder. The decoder is based on GPT \cite{radford2018improving} and generates output sequences in an autoregressive manner. 

\textbf{Model joint-training and Inference} We optimize the sequence-to-sequence model using cross-entropy loss as follows. 
\begin{equation}
    L = - \sum_{i=1}^{n}\sum_{j=1}^{|y|} \log P_{\theta}(y_{i, j} | y_{i, <j}, x_i, t_i)
\end{equation}
where $n$ is the batch size, $\theta$ represents the model parameters, $x_{i}$ represents the input image, $t_{i}$ stands for the input instruction, and $y_{i, j}$ denotes the output token at position $j$ for the \( i \)-th sample at each batch. To enhance the quality of generation during inference, we employ various decoding techniques, such as beam search.

\textcolor{black}{\textbf{Data Preprocessing} All images were uniformly resized to 512x512 pixels and padded as necessary to maintain the aspect ratio, followed by adjustments for contrast and brightness. We removed redundant white spaces from the instructions and answers and limited the length to 128 tokens. For report generation, we extracted the findings and impression sections of the report 
% and remove redundant white space, as outlined in \cite{chen-etal-2020-generating}. 
and filter out irrelevant information, such as "compared with the previous report", focusing on diagnosis-related information that can be obtained from the images. Following Pix2seq \cite{chenpix2seq}, the coordinates used for location and segmentation are divided uniformly into integers that lie between 1 and 1000.}

% Training Protocol and Implementation Details
\textbf{Training Details} During training, each batch systematically incorporates data from all four tasks, ensuring a balanced and comprehensive approach. Each sample within a batch is composed of task-specific instructions paired with corresponding images, which are essential for the model’s ability to differentiate between tasks and generate the appropriate outputs. The loss function is calculated concurrently for all tasks within each batch, allowing for efficient optimization across the different objectives.
The distribution of training data within each batch is set at 0.15 for classification, 0.2 for lesion localization, 0.5 for report generation, and 0.15 for segmentation. Our sensitivity analysis shows that increasing the weight of a specific task generally improves its corresponding performance (see Supplementary Table \ref{tab:abl}). Notably, the report generation task is the most sensitive to its weight, with higher weight leading to clear gains, while only marginally affecting other tasks. Since report generation is our primary objective, we adopt (0.15, 0.20, 0.15, 0.50) to prioritize its performance while maintaining reasonable results on basic tasks.
During training, 
we employ a learning rate of $10^{-5}$, a warm-up learning rate of $10^{-7}$, and a dropout rate of 0.1, with a batch size of 256 for 30 epochs. During inference, a beam width of 6 is used to improve output accuracy.
Detailed training hyperparameters could be found in Supplementary Table ~\ref{tab:hyperparameters}.

\textbf{Computing Platform}
Standardized anonymization of patient information is performed before leaving the hospitals using the DICOM Anonymizer tool (http://mirc.rsna.org/download). Python (version 3.8.18) is employed for all experiments and data analyses.
% leveraging a range of open-source libraries to ensure reproducibility.
For model pre-training, we utilize eight NVIDIA V100 GPUs, each with 32 GB of memory, using the DistributedDataParallel (DDP) feature in PyTorch (version 1.11.0, CUDA 11.8) to efficiently distribute training across multiple GPUs.
For image input handling, the Pillow library (v10.0.1) is responsible for loading the images. During the training phase, additional image manipulation and loading are carried out with timm (version 0.9.2), torchvision (version 0.12.0), and opencv-python (version 4.6.0).
To evaluate the model's performance, we calculate language-based metrics like ROUGE-L and CIDEr using pycocotools (version 2.0.7) and pycocoevalcap (version 1.2). Other performance indicators are measured using the torchmetrics library (version 1.0.1). For data preprocessing and analysis, we rely on Numpy (version 1.21.6) and Pandas (version 2.0.3), which provide efficient support for handling and analyzing large datasets.

\textcolor{black}{
\textbf{Deployment Specifications} For clinical deployment, the inference of IMT-CXR is designed to be efficient and accessible. The model can run on a single GPU with at least 16 GB of memory, i.e., NVIDIA V100. Built upon standard open-source frameworks, it can be containerized and integrated via REST APIs, minimizing software integration complexity. On such hardware, the average end-to-end inference time for processing a single chest radiograph through all four tasks is approximately 3 seconds. These specifications demonstrate the framework's practical feasibility for integration into clinical workflows.}

\subsubsection{Downstream Finetuning and Evaluation}

We evaluated four clinical tasks, i.e., multi-label classification, lesion localization, anatomical segmentation, and radiology report generation across ten publicly and privately sourced datasets, totaling over 145 thousand samples. For multi-label classification, performance was assessed on ChestX-ray14 \cite{wang2017chestx}, CheXpert \cite{irvin2019chexpert}, and the RSNA Pneumonia Challenge \cite{shih2019augmenting}, while lesion localization capabilities were evaluated using MS-CXR \cite{boecking2022making}, ChestX-ray14, and RSNA Pneumonia datasets. Report generation quality was validated on the MIMIC-CXR dataset \cite{johnson2019mimic}.
Supplementary materials provide detailed test set distributions: Supplementary Table~\ref{table:cls_dataset_distribution} specifies class frequencies for ChestX-ray14 and CheXpert classification tasks; Supplementary Tables \ref{table:loc_dataset_distribution}-\ref{table:loc_dataset_split} document class distribution in ChestX-ray14 and MS-CXR localization evaluations.

Standardized metrics were employed to ensure cross-task comparability: F1 score and AUC for classification tasks; accuracy (ACC) and mean Intersection-over-Union (mIoU) for lesion localization; Dice coefficient for anatomical segmentation; and both clinical efficacy metrics (F1, precision, recall) and natural language processing benchmarks (BLEU-1 \cite{papineni2002bleu}, ROUGE-L \cite{lin2004rouge}) for report generation. All NLP metrics utilized unigram-based calculations by default.
The proposed method was rigorously benchmarked against 20+ specialist approaches 
\cite{chenSimpleFrameworkContrastive2020,zhang2022contrastive,Huang2021GLoRIAAM,boecking2022making,Zhou_2022,geng2022multimodal,wang2022multi,wu2023medklip,xiao2023delving,zhou2023advancing,VGTR2022vg,Zhu_2022,Deng_2021_ICCV,chen2023medical,muller2025chex,zhang2023knowledge,li2021referring,anderson2018bottom,rennie2017self,chen2021cross,liu2023contrastive,tanida2023interactive,jeong2024multimodal}, with comprehensive results detailed in the Supplementary Materials.
The evaluation datasets include

1) \textbf{ChestX-ray14} \cite{wang2017chestx}: 112,120 images with 14 disease labels, including 984 manually annotated bounding boxes for eight pathologies. Classification followed official patient-wise splits, while localization used random 7:1:2 stratification.

2) \textbf{CheXpert} \cite{irvin2019chexpert}: 224,316 images (65,240 patients) evaluated on five key pathologies. Expert-curated test sets \cite{wang2022multi} were adopted with 5,000 validation samples randomly selected from the training set.

3) \textbf{MS-CXR} \cite{boecking2022making}: 1,153 samples (MIMIC-CXR subset) with radiologist-curated BBox and attribute descriptions. Random patient-wise 7:1:2 split applied.

4) \textbf{RSNA Pneumonia} \cite{shih2019augmenting}: 26,683 images (Pneumonia/Normal) with official splits: 25,184/1,500/3,000 for train/val/test.

5) \textbf{JSRT} \cite{jsrt2000lung}: 247 images (154 nodule cases). Healthy subsets were randomly split (7:1:2) for lung segmentation.

6) \textbf{MS-PS}: Proprietary pneumothorax segmentation dataset (N=233 from MS-CXR) with identical random split protocol.

\subsection{Quantification and statistical analysis}

To assess the clinical utility of AI-generated reports, we conducted a radiologist-led evaluation involving three radiologists, ranging from junior to senior levels. The study cohort comprised 160 cases: 120 retrospectively collected from three tertiary medical centers (listed in details below) and 40 retrospectively sampled from the MIMIC-CXR test set. To ensure methodological consistency, we excluded cases containing multiview comparisons or longitudinal imaging analyses in report narratives.
Our study involved two distinct yet complementary human evaluations: (a) a parallel evaluation, where radiologists compared and ranked alternative reports based on their quality, and (b) an independent evaluation conducted to assess the quality of each individual report.

\textbf{Ethical Statement}
The private data used in this retrospective study was approved by the Ethics Committee of three institutes, i.e., Fengcheng People's Hospital, Huanggang Hospital of Traditional Chinese Medicine, and Longkou People's Hospital.
The committees waived the consent since the retrospective research would not change the patients' examination process. All data were adequately anonymized, and the risk of disclosing patient privacy via imaging data was minimal.

\textbf{Parallel Evaluation} \textcolor{black}{All 160 original and generated reports were randomly sampled from four centers. For each center, 10\% of the selected cases were negative, while the remaining 90\% were randomly drawn from positive cases. The distribution of our 160 clinical samples is as follows: Pleural Effusion (36.94\%), Pneumonia/Consolidation (27.26\%), Pneumothorax (11.17\%), Normal (10.84\%), Atelectasis (10.51\%), and Edema (3.28\%).
The radiologists were unaware of the source of the reports and reviewed them in a randomized order.
Report quality was subjectively scored on a 1-5 scale (1=worst, 5=best). The detailed scoring guidelines are provided in Table \ref{tab:scoring_guidelines}.}

\begin{table}[htbp]
\centering
\caption{Detailed Guidelines for Scoring Chest X-ray Report Quality}
\label{tab:scoring_guidelines}
\vspace{0.5em} % 如果还需要这个垂直间距的话
\footnotesize
\begin{tabularx}{\textwidth}{p{1.8cm} X X X}
\toprule
\textbf{Score} & \textbf{Diagnostic Accuracy} & \textbf{Descriptive Completeness \& Precision} & \textbf{Example} \\
\midrule

\textbf{5(Excellent)} &
Diagnosis is \textbf{fully correct} with no false positive or negative findings. &
Description is \textbf{comprehensive and precise}, covering all relevant findings (normal/abnormal) with accurate attributes (location, size, severity). &
Correctly identifies ``small left apical pneumothorax'' and states ``normal cardiac silhouette, no pleural effusion.'' \\

\textbf{4 (Good)} &
\textbf{Core diagnosis is correct} with no major errors. &
Description is \textbf{largely complete}; core findings accurate with minor vagueness or secondary omissions. &
Identifies ``right pleural effusion'' but misgrades severity or omits mild atelectasis. \\

\textbf{3 (Fair)} &
Diagnosis is \textbf{generally correct} with minor deviations not affecting the conclusion. &
Description \textbf{contains some imprecision} (ambiguous laterality, severity deviation, minor omission). &
Describes ``left lower lobe pneumonia'' as ``consolidation in left lung base'' or omits ``mild scoliosis.'' \\

\textbf{2 (Poor)} &
Contains \textbf{clear diagnostic errors} (false positive/negative) not involving critical conditions. &
Description has \textbf{major inaccuracies or omissions}, potentially misleading. &
Misreads post-op changes as ``active pneumonia'' or misses obvious ``cardiomegaly.'' \\

\textbf{1(Unacceptable)} &
Contains \textbf{serious diagnostic errors} involving critical findings or complete misjudgment. &
Description is \textbf{severely erroneous or irrelevant}, misleading clinical decisions. &
Reports normal study despite extensive pulmonary edema or misses pneumothorax. \\

\bottomrule
\end{tabularx}
\end{table}

\textbf{Independent Evaluation} Radiologists were provided with one chest X-ray image paired with the disease findings and tasked with assessing the generated reports and original reports. During the evaluation, the radiologists were unaware of the source of the reports. They aimed to determine whether there are discrepancies or errors, any missing elements, or inaccurate descriptions (e.g., location and severity) in the reports and evaluate their clinical significance referring to the methodology \cite{yu2023evaluating, xu2023elixr}.
Six types of diseases are evaluated, i.e., Pneumothorax, Pleural Effusion, Edema, Consolidation or Pneumonia (grouped together), Atelectasis, and \textcolor{black}{N}ormal.
% following\cite{xu2023elixr}. 
Radiologists were required to assess whether every type of error exists for each specific disease when evaluating reports.
The considered errors are agreed by the radiologists and listed as follows:

1) False positives. Incorrect disease detection;
% \item False positives: Incorrect disease detection. For instance, pneumonia is mentioned in the report when it is not present, or pneumonia is identified as another disease;

2) False negatives. Missed disease detection;
% \item False negatives: missed disease detection. For instance, pneumonia is present but not mentioned or described in the report;
    
3) Inaccurate location. For instance, left lung pneumonia is described as right lung pneumonia;
    
4) Inaccurate severity. For instance, a minor pleural effusion is described as a major pleural effusion;
    
5) Non-existent references. For instance, "compared with previous" and "based on front-lateral image";

\textbf{Statistical Analysis} 
We conducted five training iterations of the model using different random seeds and recorded its performance on all tasks each time \cite{zhou2023foundation}. We calculated the mean and standard deviation of the model's performance and obtained a 95\% confidence interval through 
% \( \text{E} \pm 1.96\sigma/\sqrt{5} \)
\( mean \pm 1.96 \times standard \: deviation /\sqrt{5} \).
For tasks with multiple categories, such as multi-class classification or localization, we first calculated the performance for each category and then averaged them to obtain an overall performance measure, and used two-sided t-tests to calculate the significant differences between our model and the most competitive advanced model. In the clinical experiment section, the side-by-side comparison analysis was conducted using a Chi-Square test to verify that there was no significant difference between the generated reports by the model and the reports by doctors. The independent evaluation analysis was performed by generating 1000 bootstrap samples and reporting the 2.5th and 97.5th percentiles as the 95\% confidence interval.

% \subsection{Additional resources}
% Not applicable.

\section*{Acknowledgements}
This research was partially supported by the Centre for Perceptual and Interactive Intelligence (CPII) Ltd under the Innovation and Technology Commission (ITC)'s InnoHK (L.X., H.L. and S.Z.) and the Guangdong Basic and Applied Basic Research Foundation (No. 2023B1515130008, XW). H.L. and S.Z. are PI and co-PI of the CPII. Thanks to Dr. Simon Yu, Hao Sun, Xiaoyu Yang, Landi He and Zehong Chen for the valuable discussions in this study.

\section*{Author contributions}
All authors have contributed fully to the concept and design of the study.  LX is the project leader. LX and ZN collected the clinical data, performed the experiments, and analyzed the experiment results. XL performed the comparative experiments with other methods. XW, SZ, and HL supervised the projects and gave final approval of the manuscript. All authors have carefully read and approved the final manuscript.

\section*{Declaration of interests}
The authors declare no competing interests.

\bibliographystyle{unsrt}
\bibliography{references}  %%% Uncomment this line and comment out the ``thebibliography'' section below to use the external .bib file (using bibtex) 

@article{chowdhery2023palm,
  title={Palm: Scaling language modeling with pathways},
  author={Chowdhery, Aakanksha and Narang, Sharan and Devlin, Jacob and Bosma, Maarten and Mishra, Gaurav and Roberts, Adam and Barham, Paul and Chung, Hyung Won and Sutton, Charles and Gehrmann, Sebastian and others},
  journal={Journal of Machine Learning Research},
  volume={24},
  number={240},
  pages={1--113},
  year={2023}
}

@article{singhal2023towards,
  title={Towards expert-level medical question answering with large language models},
  author={Singhal, Karan and Tu, Tao and Gottweis, Juraj and Sayres, Rory and Wulczyn, Ellery and Hou, Le and Clark, Kevin and Pfohl, Stephen and Cole-Lewis, Heather and Neal, Darlene and others},
  journal={arXiv preprint arXiv:2305.09617},
  year={2023}
}

@article{touvron2023llama,
  title={Llama: Open and efficient foundation language models},
  author={Touvron, Hugo and Lavril, Thibaut and Izacard, Gautier and Martinet, Xavier and Lachaux, Marie-Anne and Lacroix, Timoth{\'e}e and Rozi{\`e}re, Baptiste and Goyal, Naman and Hambro, Eric and Azhar, Faisal and others},
  journal={arXiv preprint arXiv:2302.13971},
  year={2023}
}

@inproceedings{chenpix2seq,
  title={Pix2seq: A Language Modeling Framework for Object Detection},
  author={Chen, Ting and Saxena, Saurabh and Li, Lala and Fleet, David J and Hinton, Geoffrey},
  booktitle={International Conference on Learning Representations},
  year={2021}
}

@article{fan2024deep,
  title={A deep-learning-based framework for identifying and localizing multiple abnormalities and assessing cardiomegaly in chest X-ray},
  author={Fan, Weijie and Yang, Yi and Qi, Jing and Zhang, Qichuan and Liao, Cuiwei and Wen, Li and Wang, Shuang and Wang, Guangxian and Xia, Yu and Wu, Qihua and others},
  journal={Nature Communications},
  volume={15},
  number={1},
  pages={1347},
  year={2024},
  publisher={Nature Publishing Group UK London}
}

@inproceedings{lu2022unified,
  title={Unified-io: A unified model for vision, language, and multi-modal tasks},
  author={Lu, Jiasen and Clark, Christopher and Zellers, Rowan and Mottaghi, Roozbeh and Kembhavi, Aniruddha},
  booktitle={The Eleventh International Conference on Learning Representations},
  year={2022}
}

@inproceedings{muller2025chex,
  title={ChEX: Interactive localization and region description in chest X-rays},
  author={Muller, Philip and Kaissis, Georgios and Rueckert, Daniel},
  booktitle={European Conference on Computer Vision},
  pages={92--111},
  year={2025},
  organization={Springer}
}

@inproceedings{chen2023medical,
  title={Medical phrase grounding with region-phrase context contrastive alignment},
  author={Chen, Zhihao and Zhou, Yang and Tran, Anh and Zhao, Junting and Wan, Liang and Ooi, Gideon Su Kai and Cheng, Lionel Tim-Ee and Thng, Choon Hua and Xu, Xinxing and Liu, Yong and others},
  booktitle={International Conference on Medical Image Computing and Computer-Assisted Intervention},
  pages={371--381},
  year={2023},
  organization={Springer}
}

@inproceedings{chenSimpleFrameworkContrastive2020,
  title = {A Simple Framework for Contrastive Learning of Visual Representations},
  booktitle = {ICML},
  author = {Chen, Ting and Kornblith, Simon and Norouzi, Mohammad and Hinton, Geoffrey},
  year = {2020},
  pages = {1597--1607},
  publisher = {PMLR},
  eventtitle = {International Conference on Machine Learning}
}

@misc{wang2022ofa,
      title={OFA: Unifying Architectures, Tasks, and Modalities Through a Simple Sequence-to-Sequence Learning Framework}, 
      author={Peng Wang and An Yang and Rui Men and Junyang Lin and Shuai Bai and Zhikang Li and Jianxin Ma and Chang Zhou and Jingren Zhou and Hongxia Yang},
      year={2022},
      eprint={2202.03052},
      archivePrefix={arXiv},
      primaryClass={cs.CV}
}

@article{johnson2019mimic,
  title={MIMIC-CXR, a de-identified publicly available database of chest radiographs with free-text reports},
  author={Johnson, Alistair EW and Pollard, Tom J and Berkowitz, Seth J and Greenbaum, Nathaniel R and Lungren, Matthew P and Deng, Chih-ying and Mark, Roger G and Horng, Steven},
  journal={Scientific data},
  volume={6},
  number={1},
  pages={317},
  year={2019},
  publisher={Nature Publishing Group UK London}
}

@article{nguyen2022vindr,
  title={VinDr-CXR: An open dataset of chest X-rays with radiologist's annotations},
  author={Nguyen, Ha Q and Lam, Khanh and Le, Linh T and Pham, Hieu H and Tran, Dat Q and Nguyen, Dung B and Le, Dung D and Pham, Chi M and Tong, Hang TT and Dinh, Diep H and others},
  journal={Scientific Data},
  volume={9},
  number={1},
  pages={429},
  year={2022},
  publisher={Nature Publishing Group UK London}
}

@article{lian2021structure,
  title={A structure-aware relation network for thoracic diseases detection and segmentation},
  author={Lian, Jie and Liu, Jingyu and Zhang, Shu and Gao, Kai and Liu, Xiaoqing and Zhang, Dingwen and Yu, Yizhou},
  journal={IEEE Transactions on Medical Imaging},
  volume={40},
  number={8},
  pages={2042--2052},
  year={2021},
  publisher={IEEE}
}

@inproceedings{irvin2019chexpert,
  title={Chexpert: A large chest radiograph dataset with uncertainty labels and expert comparison},
  author={Irvin, Jeremy and Rajpurkar, Pranav and Ko, Michael and Yu, Yifan and Ciurea-Ilcus, Silviana and Chute, Chris and Marklund, Henrik and Haghgoo, Behzad and Ball, Robyn and Shpanskaya, Katie and others},
  booktitle={Proceedings of the AAAI conference on artificial intelligence},
  volume={33},
  pages={590--597},
  year={2019}
}

@inproceedings{boecking2022making,
  title={Making the most of text semantics to improve biomedical vision--language processing},
  author={Boecking, Benedikt and Usuyama, Naoto and Bannur, Shruthi and Castro, Daniel C and Schwaighofer, Anton and Hyland, Stephanie and Wetscherek, Maria and Naumann, Tristan and Nori, Aditya and Alvarez-Valle, Javier and others},
  booktitle={European conference on computer vision},
  pages={1--21},
  year={2022},
  organization={Springer}
}

@inproceedings{wang2017chestx,
  title={Chestx-ray8: Hospital-scale chest x-ray database and benchmarks on weakly-supervised classification and localization of common thorax diseases},
  author={Wang, Xiaosong and Peng, Yifan and Lu, Le and Lu, Zhiyong and Bagheri, Mohammadhadi and Summers, Ronald M},
  booktitle={Proceedings of the IEEE conference on computer vision and pattern recognition},
  pages={2097--2106},
  year={2017}
}

@inproceedings{zhang2022contrastive,
  title={Contrastive learning of medical visual representations from paired images and text},
  author={Zhang, Yuhao and Jiang, Hang and Miura, Yasuhide and Manning, Christopher D and Langlotz, Curtis P},
  booktitle={Machine Learning for Healthcare Conference},
  pages={2--25},
  year={2022},
  organization={PMLR}
}

@inproceedings{chen2021cross,
  title={Cross-modal Memory Networks for Radiology Report Generation},
  author={Chen, Zhihong and Shen, Yaling and Song, Yan and Wan, Xiang},
  booktitle={ACL-IJCNLP Volume 1: Long Papers},
  pages={5904--5914},
  year={2021}
}

@article{zhang2023knowledge,
  title={Knowledge-enhanced visual-language pre-training on chest radiology images},
  author={Zhang, Xiaoman and Wu, Chaoyi and Zhang, Ya and Xie, Weidi and Wang, Yanfeng},
  journal={Nature Communications},
  volume={14},
  number={1},
  pages={4542},
  year={2023},
  publisher={Nature Publishing Group UK London}
}

@inproceedings{zhou2023advancing,
title={Advancing Radiograph Representation Learning with Masked Record Modeling},
author={Hong-Yu Zhou and Chenyu Lian and Liansheng Wang and Yizhou Yu},
booktitle={The Eleventh International Conference on Learning Representations},
year={2023}
}

@article{Huang2021GLoRIAAM,
  title={GLoRIA: A Multimodal Global-Local Representation Learning Framework for Label-efficient Medical Image Recognition},
  author={Shih-Cheng Huang and Liyue Shen and Matthew P. Lungren and Serena Yeung},
  journal={2021 IEEE/CVF International Conference on Computer Vision (ICCV)},
  year={2021},
  pages={3922-3931}
}

@article{Zhou_2022,
  title={Generalized radiograph representation learning via cross-supervision between images and free-text radiology reports},
  author={Zhou, Hong-Yu and Chen, Xiaoyu and Zhang, Yinghao and Luo, Ruibang and Wang, Liansheng and Yu, Yizhou},
  journal={Nature Machine Intelligence},
  volume={4},
  number={1},
  pages={32--40},
  year={2022},
  publisher={Nature Publishing Group UK London}
}

@misc{geng2022multimodal,
      title={Multimodal Masked Autoencoders Learn Transferable Representations}, 
      author={Xinyang Geng and Hao Liu and Lisa Lee and Dale Schuurmans and Sergey Levine and Pieter Abbeel},
      year={2022},
      eprint={2205.14204},
      archivePrefix={arXiv},
      primaryClass={cs.CV}
}

@misc{liu2023contrastive,
      title={Contrastive Attention for Automatic Chest X-ray Report Generation}, 
      author={Fenglin Liu and Changchang Yin and Xian Wu and Shen Ge and Yuexian Zou and Ping Zhang and Yuexian Zou and Xu Sun},
      year={2023},
      eprint={2106.06965},
      archivePrefix={arXiv},
      primaryClass={cs.CV}
}

@article{li2021referring,
  title={Referring transformer: A one-step approach to multi-task visual grounding},
  author={Li, Muchen and Sigal, Leonid},
  journal={Advances in neural information processing systems},
  volume={34},
  pages={19652--19664},
  year={2021}
}

@inproceedings{Zhu_2022,
  title={Seqtr: A simple yet universal network for visual grounding},
  author={Zhu, Chaoyang and Zhou, Yiyi and Shen, Yunhang and Luo, Gen and Pan, Xingjia and Lin, Mingbao and Chen, Chao and Cao, Liujuan and Sun, Xiaoshuai and Ji, Rongrong},
  booktitle={European Conference on Computer Vision},
  pages={598--615},
  year={2022},
  organization={Springer}
}

@InProceedings{Deng_2021_ICCV,
    author={Deng, Jiajun and Yang, Zhengyuan and Chen, Tianlang and Zhou, Wengang and Li, Houqiang},
    title={TransVG: End-to-End Visual Grounding With Transformers},
    booktitle={Proceedings of the IEEE/CVF International Conference on Computer Vision (ICCV)},
    month={October},
    year={2021},
    pages={1769-1779}
}

@article{shih2019augmenting,
  title={Augmenting the national institutes of health chest radiograph dataset with expert annotations of possible pneumonia},
  author={Shih, George and Wu, Carol C and Halabi, Safwan S and Kohli, Marc D and Prevedello, Luciano M and Cook, Tessa S and Sharma, Arjun and Amorosa, Judith K and Arteaga, Veronica and Galperin-Aizenberg, Maya and others},
  journal={Radiology: Artificial Intelligence},
  volume={1},
  number={1},
  pages={e180041},
  year={2019},
  publisher={Radiological Society of North America}
}

@inproceedings{wu2023medklip,
  title={Medklip: Medical knowledge enhanced language-image pre-training for x-ray diagnosis},
  author={Wu, Chaoyi and Zhang, Xiaoman and Zhang, Ya and Wang, Yanfeng and Xie, Weidi},
  booktitle={ICCV},
  pages={21372--21383},
  year={2023}
}

@inproceedings{anderson2018bottom,
  title={Bottom-up and top-down attention for image captioning and visual question answering},
  author={Anderson, Peter and He, Xiaodong and Buehler, Chris and Teney, Damien and Johnson, Mark and Gould, Stephen and Zhang, Lei},
  booktitle={CVPR},
  pages={6077--6086},
  year={2018}
}

@inproceedings{rennie2017self,
  title={Self-critical sequence training for image captioning},
  author={Rennie, Steven J and Marcheret, Etienne and Mroueh, Youssef and Ross, Jerret and Goel, Vaibhava},
  booktitle={Proceedings of the IEEE conference on computer vision and pattern recognition},
  pages={7008--7024},
  year={2017}
}

@article{yu2023evaluating,
  title={Evaluating progress in automatic chest x-ray radiology report generation},
  author={Yu, Feiyang and Endo, Mark and Krishnan, Rayan and Pan, Ian and Tsai, Andy and Reis, Eduardo Pontes and Fonseca, Eduardo Kaiser Ururahy Nunes and Lee, Henrique Min Ho and Abad, Zahra Shakeri Hossein and Ng, Andrew Y and others},
  journal={Patterns},
  volume={4},
  number={9},
  year={2023},
  publisher={Elsevier}
}

@article{devlin2018bert,
  title={Bert: Pre-training of deep bidirectional transformers for language understanding},
  author={Devlin, Jacob and Chang, Ming-Wei and Lee, Kenton and Toutanova, Kristina},
  journal={arXiv preprint arXiv:1810.04805},
  year={2018}
}

@inproceedings{radford2018improving,
  title={Improving Language Understanding by Generative Pre-Training},
  author={Alec Radford and Karthik Narasimhan},
  year={2018}
}

@article{lewis2019bart,
  title={Bart: Denoising sequence-to-sequence pre-training for natural language generation, translation, and comprehension},
  author={Lewis, Mike and Liu, Yinhan and Goyal, Naman and Ghazvininejad, Marjan and Mohamed, Abdelrahman and Levy, Omer and Stoyanov, Ves and Zettlemoyer, Luke},
  journal={arXiv preprint arXiv:1910.13461},
  year={2019}
}

@article{singhal2023large,
  title={Large language models encode clinical knowledge},
  author={Singhal, Karan and Azizi, Shekoofeh and Tu, Tao and Mahdavi, S Sara and Wei, Jason and Chung, Hyung Won and Scales, Nathan and Tanwani, Ajay and Cole-Lewis, Heather and Pfohl, Stephen and others},
  journal={Nature},
  volume={620},
  number={7972},
  pages={172--180},
  year={2023},
  publisher={Nature Publishing Group}
}

@article{zhang2024generalist,
  title={A generalist vision--language foundation model for diverse biomedical tasks},
  author={Zhang, Kai and Zhou, Rong and Adhikarla, Eashan and Yan, Zhiling and Liu, Yixin and Yu, Jun and Liu, Zhengliang and Chen, Xun and Davison, Brian D and Ren, Hui and others},
  journal={Nature Medicine},
  pages={1--13},
  year={2024},
  publisher={Nature Publishing Group US New York}
}

@article{li2024llava,
  title={Llava-med: Training a large language-and-vision assistant for biomedicine in one day},
  author={Li, Chunyuan and Wong, Cliff and Zhang, Sheng and Usuyama, Naoto and Liu, Haotian and Yang, Jianwei and Naumann, Tristan and Poon, Hoifung and Gao, Jianfeng},
  journal={Advances in Neural Information Processing Systems},
  volume={36},
  year={2024}
}

@article{xu2023elixr,
  title={ELIXR: Towards a general purpose X-ray artificial intelligence system through alignment of large language models and radiology vision encoders},
  author={Xu, Shawn and Yang, Lin and Kelly, Christopher and Sieniek, Marcin and Kohlberger, Timo and Ma, Martin and Weng, Wei-Hung and Kiraly, Attila and Kazemzadeh, Sahar and Melamed, Zakkai and others},
  journal={arXiv preprint arXiv:2308.01317},
  year={2023}
}

@article{liu2024visual,
  title={Visual instruction tuning},
  author={Liu, Haotian and Li, Chunyuan and Wu, Qingyang and Lee, Yong Jae},
  journal={Advances in neural information processing systems},
  volume={36},
  year={2024}
}

@misc{wei2022finetuned,
      title={Finetuned Language Models Are Zero-Shot Learners}, 
      author={Jason Wei and Maarten Bosma and Vincent Y. Zhao and Kelvin Guu and Adams Wei Yu and Brian Lester and Nan Du and Andrew M. Dai and Quoc V. Le},
      year={2022},
      eprint={2109.01652},
      archivePrefix={arXiv},
      primaryClass={cs.CL}
}

@misc{dai2023instructblip,
      title={InstructBLIP: Towards General-purpose Vision-Language Models with Instruction Tuning}, 
      author={Wenliang Dai and Junnan Li and Dongxu Li and Anthony Meng Huat Tiong and Junqi Zhao and Weisheng Wang and Boyang Li and Pascale Fung and Steven Hoi},
      year={2023},
      eprint={2305.06500},
      archivePrefix={arXiv},
      primaryClass={cs.CV}
}

@inproceedings{du2022glam,
  title={Glam: Efficient scaling of language models with mixture-of-experts},
  author={Du, Nan and Huang, Yanping and Dai, Andrew M and Tong, Simon and Lepikhin, Dmitry and Xu, Yuanzhong and Krikun, Maxim and Zhou, Yanqi and Yu, Adams Wei and Firat, Orhan and others},
  booktitle={International Conference on Machine Learning},
  pages={5547--5569},
  year={2022},
  organization={PMLR}
}

@article{zeng2022glm,
  title={Glm-130b: An open bilingual pre-trained model},
  author={Zeng, Aohan and Liu, Xiao and Du, Zhengxiao and Wang, Zihan and Lai, Hanyu and Ding, Ming and Yang, Zhuoyi and Xu, Yifan and Zheng, Wendi and Xia, Xiao and others},
  journal={arXiv preprint arXiv:2210.02414},
  year={2022}
}

@article{chung2024scaling,
  title={Scaling instruction-finetuned language models},
  author={Chung, Hyung Won and Hou, Le and Longpre, Shayne and Zoph, Barret and Tay, Yi and Fedus, William and Li, Yunxuan and Wang, Xuezhi and Dehghani, Mostafa and Brahma, Siddhartha and others},
  journal={Journal of Machine Learning Research},
  volume={25},
  number={70},
  pages={1--53},
  year={2024}
}

@misc{dosovitskiy2021image,
      title={An Image is Worth 16x16 Words: Transformers for Image Recognition at Scale}, 
      author={Alexey Dosovitskiy and Lucas Beyer and Alexander Kolesnikov and Dirk Weissenborn and Xiaohua Zhai and Thomas Unterthiner and Mostafa Dehghani and Matthias Minderer and Georg Heigold and Sylvain Gelly and Jakob Uszkoreit and Neil Houlsby},
      year={2021},
      eprint={2010.11929},
      archivePrefix={arXiv},
      primaryClass={cs.CV}
}

@inproceedings{liu2021swin,
  title={Swin transformer: Hierarchical vision transformer using shifted windows},
  author={Liu, Ze and Lin, Yutong and Cao, Yue and Hu, Han and Wei, Yixuan and Zhang, Zheng and Lin, Stephen and Guo, Baining},
  booktitle={Proceedings of the IEEE/CVF international conference on computer vision},
  pages={10012--10022},
  year={2021}
}

@inproceedings{papineni2002bleu,
  title={Bleu: a method for automatic evaluation of machine translation},
  author={Papineni, Kishore and Roukos, Salim and Ward, Todd and Zhu, Wei-Jing},
  booktitle={Proceedings of the 40th annual meeting of the Association for Computational Linguistics},
  pages={311--318},
  year={2002}
}

@inproceedings{lin2004rouge,
  title={Rouge: A package for automatic evaluation of summaries},
  author={Lin, Chin-Yew},
  booktitle={Text summarization branches out},
  pages={74--81},
  year={2004}
}

@article{ouyang2022training,
  title={Training language models to follow instructions with human feedback},
  author={Ouyang, Long and Wu, Jeffrey and Jiang, Xu and Almeida, Diogo and Wainwright, Carroll and Mishkin, Pamela and Zhang, Chong and Agarwal, Sandhini and Slama, Katarina and Ray, Alex and others},
  journal={Advances in Neural Information Processing Systems},
  volume={35},
  pages={27730--27744},
  year={2022}
}

@article{jsrt2000lung,
  title={Development of a digital image database for chest radiographs with and without a lung nodule: receiver operating characteristic analysis of radiologists' detection of pulmonary nodules},
  author={Shiraishi, Junji and Katsuragawa, Shigehiko and Ikezoe, Junpei and Matsumoto, Tsuneo and Kobayashi, Takeshi and Komatsu, Ken-ichi and Matsui, Mitate and Fujita, Hiroshi and Kodera, Yoshie and Doi, Kunio},
  journal={American Journal of Roentgenology},
  volume={174},
  number={1},
  pages={71--74},
  year={2000},
  publisher={Am Roentgen Ray Soc}
}

@article{gaggion2024chexmask,
  title={CheXmask: a large-scale dataset of anatomical segmentation masks for multi-center chest x-ray images},
  author={Gaggion, Nicolas and Mosquera, Candelaria and Mansilla, Lucas and Saidman, Julia Mariel and Aineseder, Martina and Milone, Diego H and Ferrante, Enzo},
  journal={Scientific Data},
  volume={11},
  number={1},
  pages={511},
  year={2024},
  publisher={Nature Publishing Group UK London}
}

@article{VGTR2022vg,
  title={Visual Grounding with Transformers},
  author={Du, Ye and Fu, Zehua and Liu, Qingjie and Wang, Yunhong},
  journal={2022 IEEE International Conference on Multimedia and Expo},
  year={2022},
  pages={1-6},
}

@article{rasmy2021med,
  title={Med-BERT: pretrained contextualized embeddings on large-scale structured electronic health records for disease prediction},
  author={Rasmy, Laila and Xiang, Yang and Xie, Ziqian and Tao, Cui and Zhi, Degui},
  journal={NPJ digital medicine},
  volume={4},
  number={1},
  pages={86},
  year={2021},
  publisher={Nature Publishing Group UK London}
}

@article{padchest,
  title={PadChest: A large chest x-ray image dataset with multi-label annotated reports},
  author={Aurelia Bustos and Antonio Pertusa and Jose-Maria Salinas and Maria de la Iglesia-Vaya},
  journal={Medical Image Analysis},
  volume={66},
  year={2020}
}

@article{siim,
  title={SIIM-ACR Pneumothorax Segmentation},
  author={Anna Zawacki and Carol Wu and George Shih and Julia Elliott and Mikhail Fomitchev and Mohannad Hussain and ParasLakhani and Phil Culliton and Shunxing Bao},
  journal={Kaggle},
  year={2019},
}

@article{moor2023foundation,
  title={Foundation models for generalist medical artificial intelligence},
  author={Moor, Michael and Banerjee, Oishi and Abad, Zahra Shakeri Hossein and Krumholz, Harlan M and Leskovec, Jure and Topol, Eric J and Rajpurkar, Pranav},
  journal={Nature},
  volume={616},
  number={7956},
  pages={259--265},
  year={2023},
  publisher={Nature Publishing Group UK London}
}

@inproceedings{tanida2023interactive,
  title={Interactive and Explainable Region-guided Radiology Report Generation},
  author={Tanida, Tim and Muller, Philip and Kaissis, Georgios and Rueckert, Daniel},
  booktitle={CVPR},
  pages={7433--7442},
  year={2023}
}

@inproceedings{jeong2024multimodal,
  title={Multimodal image-text matching improves retrieval-based chest x-ray report generation},
  author={Jeong, Jaehwan and Tian, Katherine and Li, Andrew and Hartung, Sina and Adithan, Subathra and Behzadi, Fardad and Calle, Juan and Osayande, David and Pohlen, Michael and Rajpurkar, Pranav},
  booktitle={Medical Imaging with Deep Learning},
  pages={978--990},
  year={2024},
  organization={PMLR}
}

@article{zhou2023foundation,
  title={A foundation model for generalizable disease detection from retinal images},
  author={Zhou, Yukun and Chia, Mark A and Wagner, Siegfried K and Ayhan, Murat S and Williamson, Dominic J and Struyven, Robbert R and Liu, Timing and Xu, Moucheng and Lozano, Mateo G and Woodward-Court, Peter and others},
  journal={Nature},
  volume={622},
  number={7981},
  pages={156--163},
  year={2023},
  publisher={Nature Publishing Group UK London}
}

@inproceedings{xiao2023delving,
  title={Delving into masked autoencoders for multi-label thorax disease classification},
  author={Xiao, Junfei and Bai, Yutong and Yuille, Alan and Zhou, Zongwei},
  booktitle={Proceedings of the IEEE/CVF Winter Conference on Applications of Computer Vision},
  pages={3588--3600},
  year={2023}
}

@article{wang2022multi,
  title={Multi-granularity cross-modal alignment for generalized medical visual representation learning},
  author={Wang, Fuying and Zhou, Yuyin and Wang, Shujun and Vardhanabhuti, Varut and Yu, Lequan},
  journal={Advances in Neural Information Processing Systems},
  volume={35},
  pages={33536--33549},
  year={2022}
}

@article{bannur2024maira,
  title={MAIRA-2: Grounded Radiology Report Generation},
  author={Bannur, Shruthi and Bouzid, Kenza and Castro, Daniel C and Schwaighofer, Anton and Bond-Taylor, Sam and Ilse, Maximilian and Perez-Garcia, Fernando and Salvatelli, Valentina and Sharma, Harshita and Meissen, Felix and others},
  journal={arXiv preprint arXiv:2406.04449},
  year={2024}
}

@article{zhou2024generalist,
  title={A Generalist Learner for Multifaceted Medical Image Interpretation},
  author={Zhou, Hong-Yu and Adithan, Subathra and Acosta, Julian Nicolas and Topol, Eric J and Rajpurkar, Pranav},
  journal={arXiv preprint arXiv:2405.07988},
  year={2024}
}

@article{young2026fewer,
  title={Fewer Tokens, Greater Scaling: Self-Adaptive Visual Bases for Efficient and Expansive Representation Learning},
  author={Young, Shawn and Zeng, Xingyu and Xu, Lijian},
  journal={arXiv preprint arXiv:2511.19515},
  year={2026}
}

@article{young2026xrayclaw,
  title={XrayClaw: Cooperative-Competitive Multi-Agent Alignment for Trustworthy Chest X-ray Diagnosis},
  author={Young, Shawn and Xu, Lijian},
  journal={arXiv preprint arXiv:2604.02695},
  year={2026}
}

@inproceedings{yang2025one,
  title={One Leaf Reveals the Season: Occlusion-Based Contrastive Learning with Semantic-Aware Views for Efficient Visual Representation},
  author={Yang, Xiaoyu and Xu, Lijian and Li, Hongsheng and Zhang, Shaoting},
  booktitle={International Conference on Machine Learning},
  pages={71425--71440},
  year={2025}
}

@article{young2026scalar,
  title={SCALAR: Spatial-Concept Alignment for Robust Vision in Harsh Open World},
  author={Yang, Xiaoyu and Xu, Lijian and Zeng, Xingyu and Wang, Xiaosong and Li, Hongsheng and Zhang, Shaoting},
  journal={Pattern Recognition},
  pages={113203},
  year={2026}
}

@article{yang2024segmentation,
  title={Segmentation and vascular vectorization for coronary artery by geometry-based cascaded neural network},
  author={Yang, Xiaoyu and Xu, Lijian and Yu, Simon and Xia, Qing and Li, Hongsheng and Zhang, Shaoting},
  journal={IEEE Transactions on Medical Imaging},
  volume={44},
  number={1},
  pages={259--269},
  year={2024},
  publisher={IEEE}
}

@article{he2026autoselect,
  title={The Model Knows Which Tokens Matter:Automatic Token Selection via Noise Gating},
  author={He, Landi and Yang, Xiaoyu and Xu, Lijian},
  journal={arXiv preprint arXiv:2603.07135},
  year={2026}
}

@article{chen2026tc,
  title={TC-SSA: Token Compression via Semantic Slot Aggregation for Gigapixel Pathology Reasoning},
  author={Chen, Zhuo and Young, Shawn and Xu, Lijian},
  journal={arXiv preprint arXiv:2603.01143},
  year={2026}
}

@article{xu2024medvilam,
  title={MedViLaM: A multimodal large language model with advanced generalizability and explainability for medical data understanding and generation},
  author={Xu, Lijian and Sun, Hao and Ni, Ziyu and Li, Hongsheng and Zhang, Shaoting},
  journal={arXiv preprint arXiv:2409.19684},
  year={2024}
}

@article{xu2024foundation,
  title={A foundation model for generalizable disease diagnosis in chest X-ray images},
  author={Xu, Lijian and Ni, Ziyu and Sun, Hao and Li, Hongsheng and Zhang, Shaoting},
  journal={arXiv preprint arXiv:2410.08861},
  year={2024}
}

@article{gao2026zerosense,
  title={ZeroSense: How Vision matters in Long Context Compression},
  author={Gao, Yonghan and Chen, Zehong and Xu, Lijian and Chen, Jingzhi and Guan, Jingwei and Zeng, Xingyu},
  journal={arXiv preprint arXiv:2603.11846},
  year={2026}
}

@article{feng2026efficient,
  title={Efficient Chest X-ray Representation Learning via Semantic-Partitioned Contrastive Learning},
  author={Feng, Wangyu and Young, Shawn and Xu, Lijian},
  journal={arXiv preprint arXiv:2603.07113},
  year={2026}
}

@inproceedings{chen2024chexagent,
  title={Chexagent: Towards a foundation model for chest x-ray interpretation},
  author={Chen, Zhihong and Varma, Maya and Delbrouck, Jean-Benoit and Paschali, Magdalini and Blankemeier, Louis and Van Veen, Dave and Valanarasu, Jeya Maria Jose and Youssef, Alaa and Cohen, Joseph Paul and Reis, Eduardo Pontes and others},
  booktitle={AAAI 2024 Spring Symposium on Clinical Foundation Models},
  year={2024}
}

@article{chavesllava,
  author = {Zambrano Chaves, Juan Manuel and Huang, Shih-Cheng and Xu, Yanbo and Xu, Hanwen and Usuyama, Naoto and Zhang, Sheng and Wang, Fei and Xie, Yujia and Khademi, Mahmoud and Yang, Ziyi and Awadalla, Hany and Gong, Julia and Hu, Houdong and Yang, Jianwei and Li, Chunyuan and Gao, Jianfeng and Gu, Yu and Wong, Cliff and Wei, Mu-Hsin and Naumann, Tristan and Chen, Muhao and Lungren, Matthew and Chaudhari, Akshay and Yeung, Serena and Langlotz, Curtis and Wang, Sheng and Poon, Hoifung},
  title = {LLaVA-Rad MIMIC-CXR Annotations},
  journal = {PhysioNet},
  year = {2025},
  note = {Version 1.0.0},
  doi = {10.13026/4ma4-k740}
}

@article{lee2025cxr,
  title={CXR-LLAVA: a multimodal large language model for interpreting chest X-ray images},
  author={Lee, Seowoo and Youn, Jiwon and Kim, Hyungjin and Kim, Mansu and Yoon, Soon Ho},
  journal={European Radiology},
  pages={1--13},
  year={2025},
  publisher={Springer}
}

@article{bai2025deepseek,
  title={A DeepSeek-Powered AI System for Automated Chest Radiograph Interpretation in Clinical Practice},
  author={Bai, Yaowei and Zhang, Ruiheng and Lei, Yu and Duan, Xuhua and Yao, Jingfeng and Ju, Shuguang and Wang, Chaoyang and Yao, Wei and Guo, Yiwan and Zhang, Guilin and others},
  journal={arXiv preprint arXiv:2512.20344},
  year={2025}
}
\clearpage

\appendix

\section{Supplementary Material}
\renewcommand\theHtable{Appendix.\thetable}
\renewcommand{\tablename}{Supplementary Table}
\setcounter{table}{0}

\renewcommand{\theHfigure}{Appendix.\arabic{figure}} % 针对图片
\setcounter{figure}{0}
\renewcommand{\figurename}{Supplementary Figure}

\subsubsection*{Instruction Design Details}

\begin{figure}[t]
	\centering
    \includegraphics[width=0.995\linewidth]{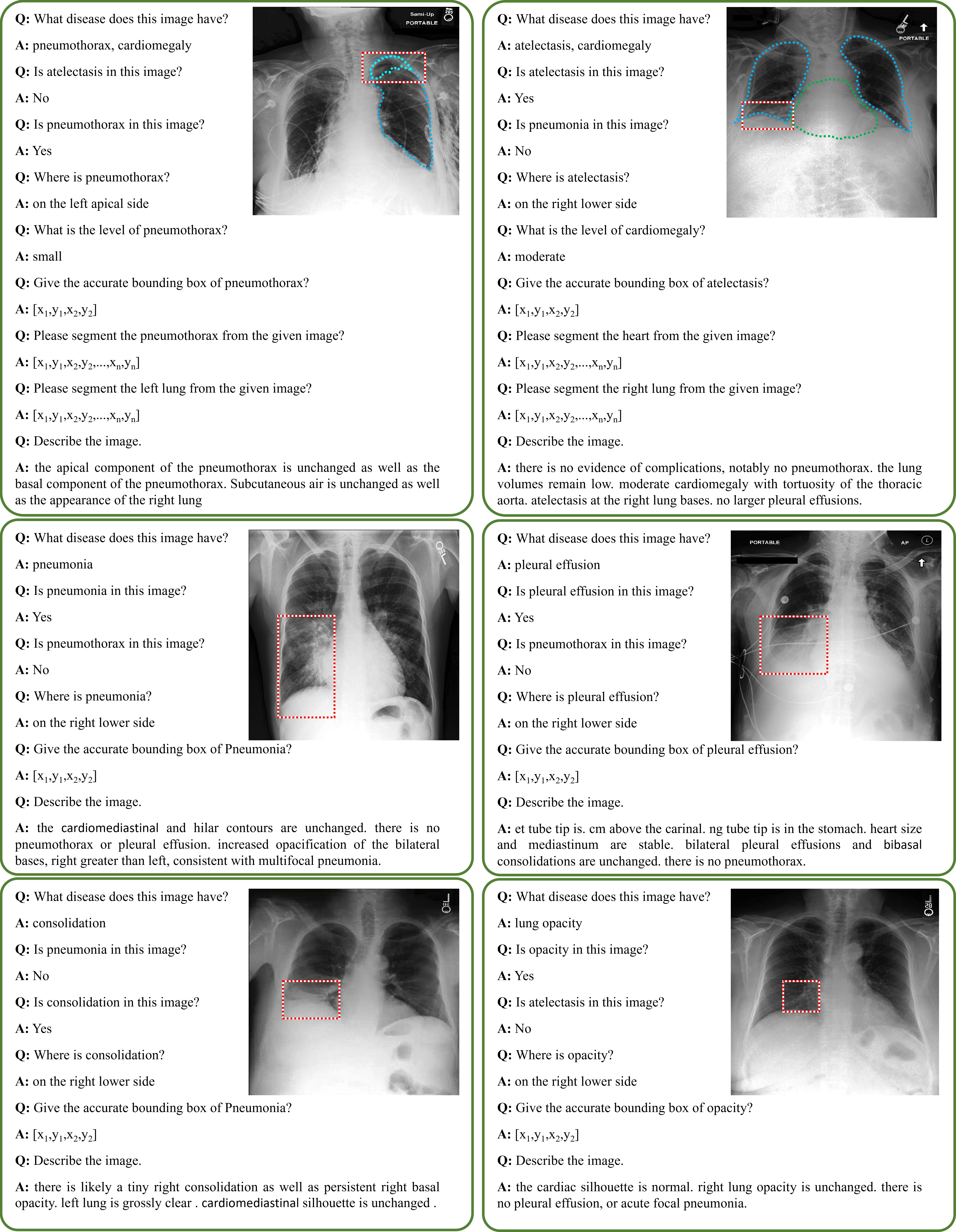}
	\caption{\textcolor{black}{Typical examples of instruction set for six disease labels: Pneumothorax, Atelectasis, Pneumonia, Pleural Effusion, Consolidation, and Opacity, related to Figure \ref{fig:overview}.} The left panel indicates the multiple instruction sets utilized during the training and testing phase. In the Chest X-ray image, the red dash line BBox denotes the region detected by IMT-CXR. }
	\label{fig:instruction}
\end{figure}

% In the clinical context of chest X-ray images, 
In clinical chest X-ray analysis, radiologists typically identify potential diseases, locate relevant regions, and subsequently generate a comprehensive report based on observation. This process involves tasks such as disease classification, localization, and report generation. Historically, either multiple single-task models or a single multi-task model were employed to accomplish these goals, but these approaches lacked intrinsic correlations between tasks.
By utilizing multiple instruction sets during the joint training approach, we not only enable the model to learn task-related features but also activate its potential capabilities to adapt to other tasks. 
% As described in Supplementary Figure \ref{fig:instruction}, 
% \ref{fig:overview}(d)
We develop a series of instructions containing placeholders (Supplementary Figure \ref{fig:instruction}), allowing us to generate a wide range of task descriptions for high-level task-specific modification. By adhering to these diverse instructions, our model is capable of smoothly transitioning between different vision-based tasks and executing them in a harmonized fashion. Here, we introduce the organization of instructions for task-level customization, including disease classification, localization, segmentation, and report generation as follows. 
% For example,

1) \textbf{Disease Classification Dataset} includes entity information across 193 categories and 0.24M images.
For the entity classification task, the instruction is "What disease does this image have?". The answer includes all possible diseases present in the data, such as "pneumonia" and "atelectasis.", "Is Pneumonia in this image?". The response can be either "yes" or "no". 
% In order to develop an attribute subset, 
We further extracted the textual phrases from the disease attributes (e.g., small left pneumothorax, normal cardiac silhouette) described in the original report of MIMIC-CXR and developed a subset that matches 135,751 images with phrases. 
The subset comprises position descriptions (e.g., left, right, base, mid) and severity descriptions (e.g., mild, moderate, severe) for ten common diseases, i.e., Cardiomegaly, Pneumonia, Effusion, Atelectasis, Edema, Consolidation, Pneumothorax, Opacity, Fracture, and Supported Devices.
% The inclusion of  are highly beneficial for improving the quality of reports. 
% During the inference stage, these phrases are obtained from the classification tasks and lesion localization tasks and subsequently utilized as customized instructions for report generation.
For the severity classification task, the instruction is "What is the level of cardiomegaly?". The response can be "moderate" or "severe". The instruction for the location classification task is like "Where is pneumothorax?". The response can be "on the left apical side".

2) \textbf{Lesion Localization Dataset} 
 incorporates 
 % CXR-AL14, 
 VinDR-CXR, ChestX-Det, and In-house datasets, consisting of 0.09M images and corresponding BBox for 12 diseases.
 The instruction given for the lesion localization task is "Give the accurate bounding box of \{\}.". Here, the placeholder \{\} represents the category of the specific disease, such as "pneumonia, in the lower left lung". The response is a distinct bounding box area defined by coordinates [x\textsubscript{1}, y\textsubscript{1}, x\textsubscript{2}, y\textsubscript{2}], representing the top-left and bottom-right points.
% \textbf{The pneumothorax \&  cardiac \& lung segmentation subset} includes contour points (polygon vertexes, recomputed from the region mask) for cases of pneumothorax. We get preliminary segmentation results of 233 pneumothorax cases by utilizing IMT-CXR (trained based on ????) from MS-CXR, which has pneumothorax BBox for reference. Subsequently, we submit the segmentation results to the annotation platform and seek rapid judgments from three qualified doctors (FOR WHAT PURPOSE????). Additionally, we supplement the SIIM positive data by converting it into the appropriate format and thus get a total of 2,717 pneumothorax samples in the dataset.

 3) \textbf{Segmentation Dataset} includes CheXmask and SIIM datasets for the segmentation task, comprising 0.23M images. 
 % \textbf{The cardiac \& lung segmentation subset} includes contour points of cardiac and lung segmentation. 
 % We use the CheXmask dataset for pretraining and employ IMT-CXR to inquire about the precise locations of the heart and left/right lungs in all the image data from MIMIC-CXR. 
 % Subsequently, 
 We calculate the Cardiothoracic Ratio (CTR) for each image and compare it with the corresponding relationship described in the reports
 % (e.g., CTR < 0.51: normal cardiac silhouette; 0.51 < CTR < 0.55: mild cardiomegaly; 0.55 < CTR < 0.6: moderate cardiomegaly; CTR > 0.6: severe cardiomegaly). 
 (normal (<0.51), mild (0.51-0.55), moderate (0.55-0.6), and severe cardiomegaly (>0.6)).
 This comparison allows us to filter the data accordingly. 
 The SIIM dataset is collected for pneumothorax segmentation. We further supplement the disease phrase subset and segmentation subsets as follows. The pneumothorax subset includes contour points (polygon vertexes, recomputed from the region mask) for 233 cases of pneumothorax.
 % Regarding the segmentation task, 
 The respective instruction is "Please segment the \{\} from the given image." For instance, "Please segment the heart from the image." The response is a polygon area defined by a set of 30 points (coordinates).

4) \textbf{Report Generation Dataset} 
includes the original MIMIC-CXR dataset of 0.24M front images, and paired radiology reports.
The instruction provided for the report generation task is "describe the image". This task specifically involves generating comprehensive reports based on chest X-ray images. 
Such brief instruction generates reports that lack accurate descriptions. We thus incorporate disease attributes in the instruction to improve the quality of the reports. During the training stage, we extract disease entities from ground truth reports and relevant severity and position attributes of the diseases within the corresponding sentences. These attributes are then combined with the original instruction for training. During the inference stage, we construct instructions for report generation using the results of the classification, segmentation, and lesion localization tasks. First, we obtain the disease category from the classification task. Then, we use lesion localization to determine the location and size of the lesion and compare it with the lung mask to determine the precise positional information.

\subsubsection*{Training and Evaluation of Baseline Models}

\textbf{Classification and Segmentation} 

\textcolor{black}{For the classification task, we have included both methods with direct inference and those without in the baseline model. 
These approaches include image contrast self-supervised methods (e.g., SimCLR\cite{chenSimpleFrameworkContrastive2020}), image mask self-supervised methods (e.g., Medical MAE\cite{xiao2023delving}, M3AE\cite{geng2022multimodal}), image/text mask-based methods (e.g., REFERS\cite{Zhou_2022}, MRM\cite{zhou2023advancing}), and CLIP-based approaches (e.g., ConVIRT\cite{zhang2022contrastive}, GLoRIA\cite{Huang2021GLoRIAAM}, BioVil\cite{boecking2022making}, MedKLIP\cite{wu2023medklip}, MGCA\cite{wang2022multi}).}

ConVIRT \cite{zhang2022contrastive}, GLoRIA \cite{Huang2021GLoRIAAM}, and BioVil \cite{boecking2022making} are utilized as baseline models for both disease classification and segmentation tasks.
% For classification, we strictly follow the official train-test split for ChestX-ray14 and MedKLIP\cite{wu2023medklip} test setting for RSNA, respectively. For segmentation, we randomly split the dataset into training/validation/test sets by 7:1:2. 
% In all three models, ResNet-50 and BERT are chosen as the visual and text encoders, respectively. To perform direct inference of classification, we adopt the methods proposed in GLoRIA and BioVil, which transform the image classification task into a text-image matching task. Specifically, the test image is fed into the image encoder to generate image features. The test disease labels are subsequently formulated as text prompts and fed into the text encoder to generate text features. We then calculate the similarity between the image and text features. The prediction scores are set with normalized similarities. For ACC and F1, we utilize the validation dataset to determine the best score threshold for each class. Furthermore, we adhere to the official training strategies and train each model for 50 epochs during the fine-tuning process. Supplementary Table~\ref{tab:cls} shows the classification tasks achieve satisfactory results across all diseases. 
For disease attribute classification, we considered combining synonymous location descriptions. When calculating the metrics, we treated "lower lobe/base/basal/basilar" as "lower" and "apical/upper" as "upper." We enhanced disease classification by considering vocabulary related to disease uncertainty in the reports.
For instance, we treated low-probability terms like 'not exclude' and 'cannot accurately assess' as negative indicators, while high-probability terms like 'likely' and 'probable' were considered positive indicators. 
Additionally, we extended the keyword search for describing disease severity. For example, terms like 'severe', 'moderate to severe', and 'moderate to large' all indicated 'severe' conditions.

\textbf{Lesion Localization} 

Medical visual grounding research remains constrained by limited annotated datasets and sparse granularity in existing references \cite{boecking2022making,chen2023medical,muller2025chex,bannur2024maira}. While BioViL \cite{boecking2022making} pioneered unsupervised medical VG, it demonstrates performance gaps relative to supervised counterparts. Subsequent work like MedRPG \cite{chen2023medical} enhanced cross-modal alignment through contrastive-attention mechanisms, though methodological discrepancies in MS-CXR dataset organization, particularly its single-box annotations for conditions like edema, limit direct comparability (Supplementary Table~\ref{table:loc_dataset_split}).
The field has since evolved with ChEX \cite{muller2025chex} acknowledging multitask synergies from our preliminary framework and MAIRA-2 \cite{bannur2024maira} leveraging large multimodal architectures. Given these chronological developments and methodological variations, our comparative analysis focuses on foundational baselines (TransVG, SeqTR, VGTR) rather than post-hoc implementations, ensuring equitable evaluation of IMT-CXR's core innovations against established paradigms.

The baseline models were pretrained with the RefCOCOg dataset and then finetuned with the dataset utilized in this study.
% Due to the absence of an official training/validation/test ratio for the lesion localization datasets, we randomly split it into training/validation/test sets by 7:1:2 for the localization task.
% Take note that the MS-CXR and ChestX-ray14 dataset doesn't come with an official division ratio for training/validation/test in the lesion localization task, and the RSNA Pneumonia dataset is devoid of ground truth bounding boxes for their test sets. As a response, we randomly partition the official training sets of these datasets into training/validation/test sets adhering to a 7:1:2 ratio for the upcoming fine-tuning experiments.
For TransVG\cite{Deng_2021_ICCV}, Resnet-50 is selected as the backbone. The BERT and ViT encoding length are 12 and 6 separately, while the maximum query length is set to 20, following the authors' recommendation. TransVG\cite{Deng_2021_ICCV} has been trained on five datasets and all models are validated, while the most competitive on RefCOCOg is reported in Supplementary Table~\ref{tab:VG}. 
For SeqTR\cite{Zhu_2022}, we follow the default settings of RefCOCOg, and DarkNet53 is selected as the detection backbone. The corresponding pre-calculated word embeddings are used to accommodate the pre-trained models. The authors have released three models on different datasets and training settings. We validate each model and the most competitive on RefCOCOg is reported in Supplementary Table~\ref{tab:VG}. For VGTR\cite{VGTR2022vg}, we followed the default settings as RefCOCOg and selected ResNet-50 and Bi-LSTM as the vision backbone and text encoder, respectively. For the evaluation of lesion localization, the IoU threshold 
% of TransVG, SeqTR, VGTR, and IMT-CXR
is set as 0.5 consistently. 
% The results of RefTR\cite{li2021referring} and MedRPG\cite{chen2023medical} can not be reproduced due to the absence of code.
% Supplementary Table~\ref{tab:VG} shows that IMT-CXR consistently achieved SOTA performance in the tests of direct inference, 100-shot and full-data finetuning.

\begin{figure}[t]
	\centering
    \includegraphics[width=0.89\linewidth]{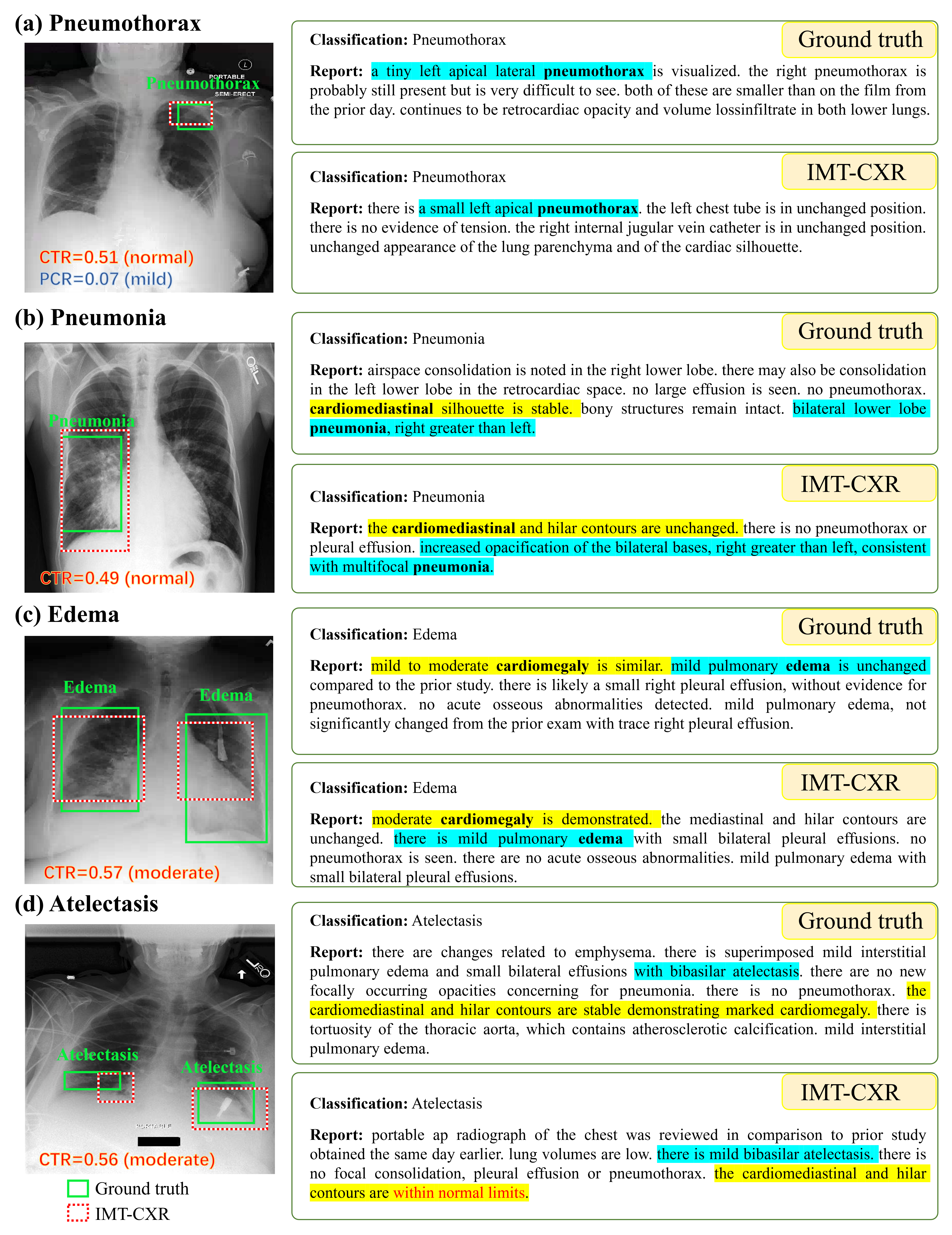}
 \caption{Typical success and failure cases of IMT-CXR and ground truth for three tasks: multi-disease classification, lesion localization, and report generation, related to Figure \ref{fig:overview}. (a) Pneumothorax; (b) Pneumonia; (c) Edema; (d) Atelectasis. In the left Chest X-ray image, the BBox with a green solid line denotes the ground truth, and the BBox with a red-white dashed line represents the region detected by IMT-CXR. In the right reports, the blue highlighted text represents the matched classified lesions compared to the ground truth report, and the yellow highlighted area represents the matched report describing other categories (e.g. cardiomegaly). CTR and PCR denote the Cardiothoracic Ratio and Pneumothorax Compress Ratio, respectively.
 For cases of (a) Pneumothorax, (b) Pneumonia, (c) Edema, our model is capable of predicting the classification, localization, and reporting of the lesions, and the results of each task are consistent.
In contrast, for case (d) Atelectasis, the model shows deviation in localizing the disease and produces an inconsistent, hallucinated statement by describing an "increased cardiac silhouette" in a comparative context, a logical contradiction given its single-image input framework due to residual comparative language in the training data.}
    \label{fig:examples}
\end{figure}

\textbf{Report Generation} The baseline methods  includes traditional captioning methods like Up-down \cite{anderson2018bottom} and Att2in \cite{rennie2017self}, multimodal fusion approaches such as R2GenCMN \cite{chen2021cross}, lesion-focused models including Contrastive Attention \cite{liu2023contrastive} and region-based methods like RGRG \cite{tanida2023interactive}, as well as retrieval-based techniques such as X-REM \cite{jeong2024multimodal}.
The quality of the generated report methods \cite{anderson2018bottom,rennie2017self,chen2021cross,liu2023contrastive,tanida2023interactive} is evaluated by report, entity, and attribute levels, with the overall performance assessed by metrics (i.e., BL-1, BL-4, METEOR, and Rouge-L), and the accuracy of the disease category evaluated by the CE metric (i.e., Precision, Recall, F1). The
RadGraph F1 metric was employed by X-REM \cite{jeong2024multimodal}, which converts two input radiology reports into knowledge
graphs and measures the overlap between the two graphs.
%We utilized Up-down\cite{anderson2018bottom}, Att2in\cite{rennie2017self}, and R2GenCMN\cite{chen2021cross} as the baseline models for report generation. 
The official MIMIC-CXR test dataset is used for evaluation. 
%Both Up-down and Att2in employ LSTM as the text encoder. Following their official implementation, Faster R-CNN and ResNet-101 are chosen as the image encoders for Up-down and Att2in, respectively. For R2GenCMN method, ResNet-101 serves as the image encoder, while a transformer-based module is utilized as the language model. 
% The number of memory vectors is set to 2048, with a dimension of 512. In terms of data preprocessing, we extract the finding and impression sections of the report and remove redundant white space, as outlined in \cite{chen-etal-2020-generating}. Moreover, we filter out irrelevant information, such as phrases like "compared with the previous report" and "discussed with doctors", focusing on diagnosis-related information that can be obtained from the images. 
%We retrained the three models following their training procedures with our preprocessed dataset and evaluate them on the official test dataset. As 
% For IMT-CXR, leveraging its multi-task capability, we find it beneficial to incorporate disease attributes as prompts during both the training and inference stages. During training, we include extracted phrases from radiologist reports as additional prompts. During inference, we utilize phrases predicted by our model as supplementary prompts. Supplementary Table~\ref{tab:ablation} shows that our report generation task achieved comparable results with the advanced specialist model on clinical efficacy metrics and comparable results on natural language processing metrics. 
Supplementary Figure \ref{fig:examples} provides more examples of multi-task results generated by our model. 
% It can be found that the proposed model is capable of identifying Pneumothorax(a), Pneumonia (b), Edema (c), and Atelectasis(d) with a lesion localization box, classification, and generated report. 
Take Supplementary Figure \ref{fig:examples}(b) for example, 
the generated report demonstrates the accurate pneumonia features and position described as "increased opacification of the bilateral bases, right greater than left", which are well consistent with the blue highlighted text in the golden standard report. The lesion localization and classification results also agree with the gold standard. 
Furthermore, the generated report shows a stable cardiomediastinal contour which could be verified by the cardiothoracic ratio of 0.49 calculated by the segmentation task. 
Through the validation of multi-tasks, the explainability of the generated reports could be greatly enhanced.

\textbf{Manual Annotation} Regarding the manual annotation process for the 160 historical reports used in the blinded comparison, three board-certified radiologists participated in establishing the ground truth. Two radiologists independently labeled each case, and the third radiologist subsequently selected the most accurate annotation. To ensure objectivity and prevent bias, all cases were randomly shuffled into a queue. Once a radiologist finished labeling a case, they were assigned the next case in the randomized queue. This procedure guaranteed that each case received annotations from three different radiologists. Crucially, throughout this process, each radiologist remained blinded to the assessments of the other two, ensuring that all labels were assigned independently. After the two initial annotations were complete, the third radiologist reviewed and selected the best label from the two, thereby synthesizing the final ground truth.

\begin{table}[t]
\caption{Comparison with other advanced specialist methods of \textbf{disease classification} task with \textbf{direct inference} setting on ChestX-ray14 dataset, related to Figure \ref{fig:cls}. 
The mean metrics (i.e., AUC and F1) refer to the macro average on the 14 diseases. }
    \centering
    \begin{tabularx}{\textwidth}{Xc*{15}{X}}
\toprule
 \rotatebox{90}{Metric} & Model & \rotatebox{90}{Mean}& \rotatebox{90}{Atelectasis} & \rotatebox{90}{Cardiomegaly} & \rotatebox{90}{Effusion} & \rotatebox{90}{Infiltration} & \rotatebox{90}{Mass} & \rotatebox{90}{Nodule}  &\rotatebox{90}{Pneumonia} &\rotatebox{90}{Pneumothorax} &\rotatebox{90}{Consolidation} &\rotatebox{90}{Edema} & \rotatebox{90}{Emphysema} & \rotatebox{90}{Fibrosis} & \rotatebox{90}{Pleural Thicken} & \rotatebox{90}{Hernia} \\
\midrule
AUC &  ConVIRT\cite{zhang2022contrastive}     & 56.0 & 45.1 & 44.3 & 63.2 & 65.1 & 61.6 & 57.2 & 63.6 & 54.1 & 63.7 & 70.2 & 41.9 & 47.4 & 56.0 & 51.2\\
                 &   GLoRIA \cite{Huang2021GLoRIAAM} & 60.9 & 65.9 & 70.2 & 74.9 & 65.2 & 60.9 & 52.1 & 59.4 & 56.6 & 69.3 & 74.4 & 50.1 & 46.2 & 60.7 & 46.0 \\
                &  BioViL\cite{boecking2022making}       & 65.9 & 52.1 & 68.4 & 74.4 & 61.2 & 65.4 & 63.4 & 67.2 & 68.3 & 64.1 & 78.1 & 64.4 & 62.0 & 64.8 & 69.0 \\
                & MedKLIP\cite{wu2023medklip}& 72.4 & 65.3 & \textbf{84.9} & \textbf{82.3} & \textbf{68.2} & 74.6 & 64.2 & \textbf{69.9} & 79.4 & 70.2 & 79.3 & \textbf{79.6} & 58.6 & 51.2 & 85.9 \\
                  &Ours & \textbf{74.3} & \textbf{74.2} & 75.9 & 79.7 & 60.8 & \textbf{77.9} & \textbf{69.6} & 68.7 & \textbf{82.6} & \textbf{70.9} & \textbf{80.0} & 72.8 & \textbf{63.8} & \textbf{74.9} & \textbf{88.4}\\
F1&  ConVIRT\cite{zhang2022contrastive}            & 13.4 & 0.6 & 0.4 & 36.4 & 43.4 & 15.5 & 14.0 & 5.8 & 20.3 & 17.5 & 12.3 & 8.1 & 3.2 & 9.9 & 0.5 \\
                 &   GLoRIA \cite{Huang2021GLoRIAAM} & 17.5 & 28.4 & 15.2 & 46.1 & 46.6 & 13.2 & 12.6 & 5.9 & 21.4 & 21.2 & 13.9 & 8.1 & 0.3 & 11.2 & 0.8 \\
                 &  BioViL\cite{boecking2022making}       & 19.2 & 23.5 & 20.9 & 43.8 & 41.4 & 17.8 & 16.4 & 6.7 & 27.4 & 17.7 & 18.7 & 12.3 & 5.6 & 11.9 & 4.5 \\
                & MedKLIP\cite{wu2023medklip}& 24.3 & 28.3 & \textbf{31.4} & \textbf{50.8} & \textbf{48.9} & 24.1 & 17.5 & 7.2 & 42.1 & 21.4 & 19.1 & \textbf{24.1} & 7.3 & 1.4 & 16.3 \\
                 &      Ours                              & \textbf{27.1} & \textbf{37.4} & 21.9 & 45.5 & 44.4 & \textbf{30.7} & \textbf{20.0} & \textbf{8.3} & \textbf{45.7} & \textbf{23.8} & \textbf{22.9} & 16.0 & \textbf{8.8} & \textbf{20.6} & \textbf{34.1}\\

\bottomrule
\hline
\end{tabularx}
    \label{tab:cls}
\end{table}

\clearpage

\begin{table}[t]
  \caption{Comparison with other specialist methods for the \textbf{disease classification} task on the ChestX-ray14, CheXpert, and RSNA Pneumonia datasets. The metric (i.e., AUC) refers to the macro average across the diseases, related to Figure \ref{fig:cls}.
  We included specialist models with direct inference (best results bolded) and those requiring further fine-tuning for comprehensive evaluation.
  The labeling ratio X\% means that X\% of the training set from a fully annotated downstream dataset are used for supervised fine-tuning. 
  The fine-tuning results of baseline models are primarily referenced here from MRM\cite{zhou2023advancing}, BioViL\cite{boecking2022making} and MedKLIP\cite{wu2023medklip}. Note that the referenced papers have minor differences in the split of the ChestX-ray14 dataset. Our approach, along with Medical MAE\cite{xiao2023delving} and REFERS\cite{Zhou_2022}, uses the official dataset split, while MRM\cite{zhou2023advancing} randomly divides the dataset into training, validation, and test sets in a 7:1:2 ratio.}
      \centering
      % \begin{tabularx}{\textwidth}{cc*{9}{X}} % 第一列为 X，其余 11 列也为 X
      \begin{tabular}{ccccccccccc}
  \toprule
  \multirow{2}{*}{Method}&
  \multirow{2}{*}{\begin{tabular}{c} Direct \\ Inference\end{tabular}}& 
  \multicolumn{3}{c}{ChestX-ray14} &
  \multicolumn{3}{c}{CheXpert} &
  \multicolumn{3}{c}{RSNA Pneumonia} \\
    & & 1\% & 10\% & 100\% &1\% & 10\% & 100\%&1\% & 10\% & 100\%  \\
  
  \midrule
        SimCLR\cite{chenSimpleFrameworkContrastive2020}  
        % &Image
        &\ding{55}   &-&- &-&- &- &- &70.1  &80.2 &84.9 \\
        Medical MAE\cite{xiao2023delving}    
        % &Image
        &\ding{55}   &-&- &82.3& &-&89.2 &-  &- &- \\
        M3AE\cite{geng2022multimodal}  
        % &Image
        &\ding{55}  &- &- &- &86.2 &87.3 &87.9 &89.0 &90.8 &92.3 \\
        REFERS\cite{Zhou_2022}  
        % &Image+Text
        &\ding{55}  &76.7 &80.9 &84.7 &87.2 &88.1 &88.2 &89.4 &91.6 &92.7 \\
        MRM\cite{zhou2023advancing}    
        % &Image+Text
        &\ding{55}   &79.4&84.0 &85.9 &88.5 &88.5 &88.7 & 91.3 &92.7 & 93.3\\
        \midrule
        MGCA\cite{wang2022multi} 
        % &Image+Text
        &$\checkmark$  &- &- &- &\textbf{88.8} &\textbf{89.1} &\textbf{89.7} &89.1 &89.9 &90.8 \\
        ConVIRT\cite{zhang2022contrastive}  
        % &Image+Text
        &$\checkmark$   &66.2 &76.6 &81.3 &85.9 &86.8 &87.3 &77.4 &80.1 &81.3 \\
        GloRIA\cite{Huang2021GLoRIAAM}  
        % &Image+Text
        &$\checkmark$   &67.1 &76.4 &81.8 &86.6 &87.8 &88.1 &86.1 &88.0 &88.6 \\
        BioViL\cite{boecking2022making}  
        % &Image+Text
        &$\checkmark$   &69.5 &75.3 &82.5 &- &- &- &88.1 &88.4 &89.1 \\
        MedKLIP\cite{wu2023medklip}  
        % &Image+Text
        &$\checkmark$   &77.2 &78.9 &83.2 &- &- &- &87.3 &88.0 &89.3 \\
        % MaCo\cite{huang2024enhancing}  &$\checkmark$   &79.3 &83.8 &85.9 &88.7 &88.7 &88.9 &91.5 &92.7 &93.6 \\
        \textbf{Ours}        
        % &Image+Text
        &$\checkmark$  &\textbf{79.3} &\textbf{82.0} &\textbf{84.3} &88.5 &89.0 &89.2 &\textbf{90.3} &\textbf{91.8} &\textbf{92.9} \\
      \bottomrule
  \hline
  % \end{tabularx}
  \end{tabular}
      \label{tab:cls_fulldata}
  \end{table}

\clearpage

\begin{table}[t]
\centering
\caption{Clinical-Relevant Metrics for Classification Tasks on the CheXpert and RSNA Datasets, related to Figure \ref{fig:cls}.}
\label{tab:combined_metrics}
\begin{tabular}{lccccccc}
\toprule
 & \multicolumn{6}{c}{CheXpert Dataset} & RSNA Dataset \\
\cmidrule(lr){2-7} \cmidrule(lr){8-8}
 & Mean & Atelectasis & Cardiomegaly & Consolidation & Edema & Pleural-Effusion & Pneumonia \\
\midrule
FN Rate  & 9.8 & 7.5 & 15.7 & 9.6 & 8.3 & 7.7  & 2.6 \\
Sensitivity  & 90.8 & 92.0 & 84.6 & 92.0 & 91.0 & 94.5  & 97.4 \\
FP Rate  & 32.3 & 35.5 & 20.6 & 18.0 & 40.5 & 47.0  & 39.8 \\
Specificity  & 68.1 & 64.0 & 78.5 & 83.4 & 59.8 & 55.0  & 60.8 \\
\bottomrule
\end{tabular}
\end{table}

\clearpage

\begin{table}[t]
\caption{Comparison with other advanced specialist methods of \textbf{lesion localization} task with \textbf{20-shot} setting on MS-CXR and ChestX-ray14 dataset, related to Figure \ref{fig:vg}.
The metrics (i.e., ACC and mIoU)  refer to the macro average on the eight diseases.}
    \centering
    \begin{tabularx}{\textwidth}{*{3}{c}*{12}{X}}
\hline
  Dataset&Metric & Model & Mean & \rotatebox{90}{Atelectasis} & \rotatebox{90}{Cardiomegaly} & \rotatebox{90}{Effusion}  &\rotatebox{90}{Pneumonia} &\rotatebox{90}{Pneumothorax} &\rotatebox{90}{Consolidation} &\rotatebox{90}{Edema} &\rotatebox{90}{Opacity} &\rotatebox{90}{Infiltrate} &\rotatebox{90}{Mass} &\rotatebox{90}{Nodule}\\
\midrule
       MS-CXR&ACC & VGTR\cite{VGTR2022vg}        &49.7 &37.1 &91.4 &46.3 &51.8 &45.5 &43.0 &43.0 &39.5 &- &- &-\\
                 && TransVG\cite{Deng_2021_ICCV} &52.8 &30.4 &91.3 &39.7 &59.6 &\textbf{61.1} &\textbf{49.9} &48.8 &41.9 &- &- &-\\
                 && SeqTR\cite{Zhu_2022}         &55.6 &40.2 &\textbf{95.5} &\textbf{49.3} &57.9 &59.7 &46.6 &51.2 &44.5 &- &- &-\\
                 && Ours                &\textbf{56.9} &\textbf{44.5} &95.4 &48.2 &\textbf{61.4} &46.5 &47.1 &\textbf{54.0} &\textbf{58.1} &- &- &-\\
       &mIoU & VGTR\cite{VGTR2022vg}&45.3 &46.2 &62.8 &43.0 &43.0 &39.3 &41.5 &43.1 &43.2 &- &- &-\\
                 &&  TransVG\cite{Deng_2021_ICCV}&47.0 &41.4 &67.1 &32.6 &46.1 &45.9 &45.5 &49.1 &48.3 &- &- &-\\
                 && SeqTR\cite{Zhu_2022} &51.3 &\textbf{49.0} &70.2 &45.3 &50.9 &\textbf{51.1} &46.8 &48.5 &48.6 &- &- &-\\
                 && Ours&\textbf{52.4} &48.5 &\textbf{71.3} &\textbf{45.4} &\textbf{55.5} &47.8 &\textbf{49.3} &\textbf{51.8} &\textbf{49.2} &- &- &-\\
ChestX-ray14 & ACC & VGTR\cite{VGTR2022vg} & 53.2 & 43.6 & 94.8 & 49.4 & 64.8 & 35.7 & - & - & - & 62.7 & 38.6 & 35.7\\
           &     & TransVG\cite{Deng_2021_ICCV} & 54.9 & 41.9 & 95.3 & 40.7 & \textbf{63.7} & 45.6 & - & - & - & \textbf{70.7} & 41.8 & 39.6\\
           &     & SeqTR\cite{Zhu_2022}       & 57.7 & \textbf{55.9} & 93.8 & \textbf{53.8} & 51.7 & \textbf{68.5} & - & - & - & 67.6 & 50.1 & 20.4\\
           &     & \textbf{Ours}            & \textbf{60.9} & 51.8 & \textbf{99.4} & 46.6 & 62.1 & 48.3 & - & - & - & 63.3 & \textbf{64.3} & \textbf{51.3}\\[5pt]
           & mIoU & VGTR\cite{VGTR2022vg}  & 46.0 & 37.4 & 72.6 & \textbf{45.1} & 58.4 & 28.3 & - & - & - & 52.8 & 34.3 & 25.5\\
           &      & TransVG\cite{Deng_2021_ICCV} & 48.0 & 37.0 & \textbf{87.4} & 39.0 & \textbf{60.5} & 35.9 & - & - & - & 51.6 & 34.8 & 23.6\\
           &      & SeqTR\cite{Zhu_2022}      & 50.0 & \textbf{49.4} & 85.1 & 44.7 & 51.1 & \textbf{63.5} & - & - & - & \textbf{58.7} & 40.4 & 24.1\\
           &      & \textbf{Ours}           & \textbf{51.3} & 48.0 & 71.5 & 44.0 & 54.8 & 46.7 & - & - & - & 55.6 & \textbf{52.0} & \textbf{38.0}\\
		\bottomrule
\end{tabularx}
    \label{tab:VG}
\end{table}

\clearpage

\begin{table}[t]
    \caption{Comparison with other advanced specialist methods of \textbf{lesion localization} task with \textbf{full data fine-tune} setting, related to Figure \ref{fig:vg}. The metrics (i.e., ACC and mIoU) refer to the macro average on the diseases for MS-CXR and ChestX-ray14.}
        \centering
        \begin{tabular}{ccccc}
    \hline
    \multirow{2}{*}{Model} &  
    \multicolumn{2}{c}{MS-CXR} &
    \multicolumn{2}{c}{ChestX-ray14} 
    % &\multicolumn{3}{c}{RSNA Pneumonia} 
    \\
      & ACC & mIoU  & ACC & mIoU  \\
    
    \midrule
        VGTR\cite{VGTR2022vg}&56.2&51.5&55.1&47.2 \\
        % SeqTR\cite{Zhu_2022} &\textbf{63.3}&\textbf{53.2}&57.8&49.3 \\
        % TransVG\cite{Deng_2021_ICCV} &59.4&50.3&58.9&51.2 \\
        TransVG\cite{Deng_2021_ICCV} &59.4&50.3&57.8&49.3 \\
        SeqTR\cite{Zhu_2022} &63.3&53.2&58.9&51.2 \\
        % ConVIRT\cite{zhang2022contrastive}\\
        % GloRIA\cite{Huang2021GLoRIAAM}&28.74&31.17&8.58&16.39\\
        % BioViL\cite{boecking2022making}&7.78&19.19&6.56&12.78\\
        % MedPRG\cite{chen2023medical}&-&-&-&-&-&-\\
        % \textbf{Ours} & 62.4 & 52.5 & \textbf{62.1} & \textbf{51.9}\\
        \textbf{Ours} & \textbf{64.2} & \textbf{54.3} & \textbf{62.1} & \textbf{51.9}\\
        \bottomrule
    \hline
    \end{tabular}
        \label{tab:VG_fulldata}
    \end{table}

\clearpage

\begin{table}[t]
    \caption{Ablation experiment of multi-task and prompt capability, related to Figure \ref{fig:prompt}. The quality of the generated report is evaluated by report, entity, and attribute levels, with the overall performance assessed by metrics (i.e., BL-4, METEOR, and Rouge-L), and the accuracy of the disease category evaluated by the CE metric (i.e., Precision, Recall, F1). The attribute metric focuses on the performance of disease severity and location described in the report.}
    \centering
    \begin{tabular}{cccccccccc}
    \toprule
     \multicolumn{2}{c}{} & \multicolumn{3}{c}{Report} & \multicolumn{3}{c}{Entity} & \multicolumn{2}{c}{Attribute} \\
     &&  BL-4 & METEOR & Rouge-L  & Precision & Recall & F1  & ACC\_S & ACC\_L \\
    \midrule
     Baseline & Ours  & 13.12 & 16.72  & 27.48  & 42.77 & 31.49 & 33.08 & 18.82 & 8.31\\
    \midrule
     Task & - LOC    & 12.90 & 16.63 & 27.07 & 41.53 & 30.21 & 31.71 & - & - \\
          & - CLS    & 12.81 & 16.50 & 27.11 & 41.35 & 30.45 & 31.54 & - & - \\
    \midrule
     Prompt & + Phrase     & 14.43 & 17.95 & 28.60 & 49.12 & 36.17 & 40.32 & 24.05 & 16.78 \\
            & + Phrase-GT  & 16.72 & 19.13 & 33.12 & 70.69 & 46.04 & 50.41 & 31.93 & 24.64 \\
    \bottomrule
    \end{tabular}
    \label{tab:ablation}
\end{table}

\clearpage

\begin{table}
	\caption{Diagnostic accuracy comparison with various \textbf{report generation} methods on MIMIC-CXR, related to Figure \ref{fig:prompt}. 
    The quality of the generated report is evaluated by report, entity, and attribute levels, with the overall performance assessed by metrics (i.e., BL-1, BL-4, METEOR, and Rouge-L), and the accuracy of the disease category evaluated by the CE metric (i.e., Precision, Recall, F1).
    The fine-tuning results of baseline models are primarily referenced here from \cite{anderson2018bottom,rennie2017self,chen2021cross,liu2023contrastive,tanida2023interactive}.
 % \# denotes the published methods that cannot be reproduced due to the lack of code. The corresponding metrics are cited from their paper.
 }
	\centering
    \begin{tabular}{cccccccccc}
		\toprule
		% \cmidrule(r){1-9}
		Dataset &Model &BL-1 &BL-4  &METEOR & Rouge-L & Precision & Recall & F1\\
		\midrule
  MIMIC-CXR &Up-down\cite{anderson2018bottom}   &31.3 &9.0&12.7&26.1&32.4&24.1&24.0\\
  &Att2in\cite{rennie2017self}&33.4 &9.8	&13.8	&27.9	&31.9	&24.3	&25.5\\
  &R2GenCMN\cite{chen2021cross}&35.1 &10.2&14.5&28.3&34.7&27.6&28.9\\
  &Constrastive Attention\cite{liu2023contrastive} &35.4&10.9&15.3&\textbf{28.4}	&36.3	&30.1	&30.6 \\
  
 &\textcolor{black}{RGRG}\cite{tanida2023interactive}&\textcolor{black}{37.3} &\textcolor{black}{12.6}&\textcolor{black}{\textbf{16.8}}&\textcolor{black}{26.4}&-&-&- \\
  &\textbf{Ours}&\textbf{38.6} &\textbf{13.1}&16.7&27.5&\textbf{42.8}&\textbf{31.5}&\textbf{33.1} \\
	\midrule
    
 % IU-Xray&R2GenCMN\cite{chen2021cross} &0.470 &0.165	&0.187	&0.371	&-	&-	&- &\\
 %  &Constrastive\cite{zhang2022contrastive} &0.492 &0.169	&0.193	&0.381	&-	&-	&- &\\
 %  &KiUT\cite{huang2023kiut}                  &\textbf{0.525}  &\textbf{0.199}	&\textbf{0.242}	&\textbf{0.409}	&-	&-	&- &\\
  % &KiUT\cite{huang2023kiut}                    &0.525 &0.199 &0.242 &0.409&-	&-	&- &\\
 %  &\textbf{Ours }                            &0.408  &0.133	&0.175	&0.317	&-	&-	&- &\\
  % \bottomrule
	\end{tabular}
	\label{tab:report}
\end{table}

\clearpage

\begin{table}
	\caption{Diagnostic accuracy comparison with various \textbf{report generation methods based on large language model} on MIMIC-CXR, related to Figure \ref{fig:prompt}. 
    The quality of the generated report is evaluated by report, entity, and attribute levels, with the overall performance assessed by metrics (i.e., BL-4 and Rouge-L), and the accuracy of the disease category evaluated by the CE metric (i.e., \textcolor{black}{Marco-avg CheXbert F1}, RadGraph F1).
    The fine-tuning results of baseline models are primarily referenced here from \cite{bannur2024maira,chavesllava,lee2025cxr,chen2024chexagent,bai2025deepseek}.
 }
	\centering
    \begin{tabular}{ccccccccc}
		\toprule
		 Model &language model &Vision  &CheXbert F1 & RadGraph F1 & Rouge-L & BL-4 \\
		\midrule
        MAIRA-2\cite{bannur2024maira}&Vicuna(7B) &RAD-DINO (0.3B)&41.6&34.6&38.4&23.1\\
  LLaVA-Rad\cite{chavesllava}&Vicuna(7B) &ViT-L/16(0.3B)	&39.5	&29.4	&30.6	&15.4	\\
  CXR-LLaVA\cite{lee2025cxr} &LLaMA-2(7B)&ViT-L/16(0.3B)&-&-	&-	&-\\
  CheXagent\cite{chen2024chexagent}   &Phi-2(2.7B) &SigLIP-Large(0.3B)&44.4&26.6&-&-\\
 Janus-Pro-CXR\cite{bai2025deepseek}&DeepSeek-LLM(1B) &SigLIP(0.086B)&34.7&26.4&28.6&11.0\\
  \textbf{Ours}&BART(0.9B) &ResNet-152(0.06B)&44.1&25.3&27.5&13.1\\
	\midrule
	\end{tabular}
	\label{tab:report2}
\end{table}

\clearpage

\begin{table}[t]
\centering
\small
\caption{\textcolor{black}{Task-level efficiency comparison under a single NVIDIA V100 (32GB) GPU with FP16 inference and batch size = 1, related to Figure \ref{fig:workflow}. The task-specific pipeline uses representative model families for each task.}}
\label{tab:efficiency_revision}

\begin{tabular}{>{\centering\arraybackslash}m{2.3cm}
                >{\centering\arraybackslash}m{3.2cm}
                >{\centering\arraybackslash}m{1.9cm}
                >{\centering\arraybackslash}m{1.8cm}
                >{\centering\arraybackslash}m{1.8cm}}
\toprule
\textbf{Pipeline} & \textbf{Task} & \textbf{\#Params} & \textbf{GPU Mem} & \textbf{Latency} \\
\midrule

\multirow{4}{*}{Task-specific}
& Classification (ResNet) 
& 0.05--0.07B 
& 0.6--0.9 GB 
& 0.1--0.3 s \\

& Localization (Transformer grounding) 
& 0.08--0.12B 
& 0.8--1.1 GB 
& 0.3--0.8 s \\

& Segmentation (U-Net) 
& 0.02--0.04B 
& 0.5--0.8 GB 
& 0.2--0.6 s \\

& Report generation (Seq2Seq Transformer) 
& 0.35--0.57B 
& 1.1--2.2 GB 
& 0.9--2.3 s \\

\midrule

& \textbf{Total} 
& \textbf{0.5--0.8B} 
& \textbf{3--5 GB} 
& \textbf{1.5--4 s} \\

\midrule

\textbf{Unified (Ours)}
& All tasks
& \textbf{0.9B}
& \textbf{3.5--5 GB}
& \textbf{3--5 s} \\

\bottomrule
\end{tabular}
\label{tab:dep}
\end{table}

\clearpage

%%%%%%%%%% Population characteristics %%%%%%%%%%
\begin{table}
	\caption{Population characteristics of our training and validation dataset, related to Figure \ref{fig:overview}.}
  % As there are three radiologists involved, the P threshold is set to 0.0167 (0.05/3).
	\centering
    \begin{tabular}{ccp{10cm}}
		\toprule
		\cmidrule(r){1-3}
		 Dataset &Split &Population characteristics\\
		\midrule
          \textbf{MIMIC-CXR}\cite{johnson2019mimic} & train, validation & The dataset was collected at the Beth Israel Deaconess Medical Center in Boston, MA between 2011 - 2016. \\
          % \textbf{Padchest}\cite{padchest} & train & This dataset was collected from San Juan Hospital (Spain) between 2009 and 2017. The dataset encompasses six distinct positional views. 50.3\% images correspond to women and 49.7\% to men. The median age of the population was 70, with a standard deviation of 20. \\
          % \textbf{CXR-AL14}\cite{fan2024deep} & train & - \\
          \textbf{VinDr-CXR}\cite{nguyen2022vindr} & train & The dataset were retrospectively collected from the Hospital 108 and the Hanoi Medical University Hospital between 2018 - 2020, two of the largest hospitals in Vietnam. The median age of the population was 43.77. \\
          \textbf{ChestX-Det}\cite{lian2021structure} & train & It is a subset of ChestX-ray14. The median age of population was 51 with standard deviation of 16. 57.2\% images correspond to men and 42.8\% to women.\\
          \textbf{In-house dataset} & train & The dataset consists of 2531 images which were collected from three hospitals in China. \textcolor{black}{All the images are captured in the anteroposterior or posteroanterior position.} \\
          \textbf{CheXmask}\cite{gaggion2024chexmask} & train & This dataset is collected from six public dataset. We only use MIMIC-CXR, Padchest, and VinDr-CXR for training. \\
          \textbf{SIIM}\cite{siim} & train & - \\
          \textbf{ChestX-ray14}\cite{wang2017chestx}  &validation
          % &The dataset is extracted from the clinical PACS database at National Institutes of Health Clinical Center and consists of ~60\% of all frontal chest x-rays in the hospital. The median age of population was 49 with standard deviation of 17. 56.5\% images correspond to men and 43.5\% to women.\\
          &The dataset is extracted from the clinical PACS database at the National Institutes of Health Clinical Center and comprises approximately 60\% of all frontal chest X-rays taken in the hospital. The median age of the population is 49, with a standard deviation of 17. Additionally, 56.5\% of the images correspond to men, while 43.5\% correspond to women.\\
          \textcolor{black}{\textbf{CheXpert}\cite{irvin2019chexpert}} &\textcolor{black}{validation} & \textcolor{black}{The dataset comprises chest radiographic studies gathered retrospectively from Stanford Hospital, spanning from October 2002 to July 2017 across both inpatient and outpatient settings, along with the related radiology reports.} \\
          \textbf{MS-CXR}\cite{boecking2022making} &validation & It is a subset of MIMIC-CXR. \\
          \textbf{RSNA Pneumonia}\cite{shih2019augmenting}  & validation & The dataset was collected from the 112,000-image public National Institutes of Health (NIH) CXR8 dataset. 56.9\% images correspond to man and 43.1\% to women. \\
          \textbf{JSRT}\cite{jsrt2000lung}  & validation &  This database were collected from 13 medical centers in Japan and one institution in the United States. The average age of patients with nodules was 60 years old. 51\% correspond to man and 49\% correspond to women.\\
          \textbf{MS-PS} & validation & It is a subset of MS-CXR. \\ 
          
        \midrule
	\end{tabular}
	\label{tab:demography}
\end{table}

\clearpage

\begin{table}[t]
\centering
% \caption{Dataset overview for multi-task joint training and fine-tuning. We strictly follow the official train/validation/test split. For the datasets without official split ratio, we randomly split them into train/validation/test sets by 7:1:2. \textcolor{black}{For CheXpert, we use the expert-labeled validation set as test data and randomly pick 5,000 radiographs from the training set for validation 
%  \cite{zhou2023advancing}.}}
 \caption{
 Dataset overview for multi-task joint training and fine-tuning, related to Figure \ref{fig:overview}. We strictly follow the official train/validation/test split. For datasets without an official split ratio, we randomly divide them into train/validation/test sets in a 7:1:2 ratio. Exceptionally, for CheXpert, we use the expert-labeled validation set as test data and randomly select 5,000 radiographs from the training set for validation \cite{zhou2023advancing}.}
\begin{tabular}{@{}l l l r r r c@{}}
\toprule
\textbf{} & \textbf{Task} & \textbf{Dataset} & \textbf{Train} & \textbf{Validation} & \textbf{Test} &\textbf{Official Split} \\
\midrule
\multirow{11}{*}{\textbf{Multi-Task Joint Train}} & \multirow{3}{*}{Classification} & MIMIC-CXR & 242,306 & 2,991 & 5,159 & Y \\
% & & Padchest & 112,608 & 16,086 & 32,174 & N \\
% & & CXR-AL14 & 141,209 & 1,500 & 3,000 & N \\
& & VinDr-CXR & 13,500 & 1,500 & 3,000 & Y \\
\cmidrule{2-6}
& \multirow{4}{*}{Localization} & VinDr-CXR & 13,500 & 1,500 & 3,000 & Y \\
& & ChestX-Det & 2,723 & 302 & 553 & Y \\
% & & CXR-AL14 & 141,209 & 1,500 & 3,000 & N \\
& & In-House & 1,772 & 253 & 506 & N\\
\cmidrule{2-6}
& \multirow{2}{*}{Segmentation} & CheXmask & 219,316 & 5,000 & 10,000 & N \\
& & SIIM & 8,463 & 1,209 & 2,418 & N \\
\cmidrule{2-6}
& Report Generation & MIMIC-CXR & 242,306 & 2,991 & 5,159 & Y \\
\midrule
\multirow{8}{*}{\textbf{Task-Specific Fine-tune}} & \multirow{3}{*}{Classification} & ChestX-ray14 & 77,871 & 8,652 & 25,595 & Y \\
& & CheXpert & \textcolor{black}{218,414} & \textcolor{black}{5,000} & \textcolor{black}{234} & \textcolor{black}{N} \\
& & RSNA Pneumonia & 25,184 & 1,500 & 3,000 & Y \\
\cmidrule{2-6}
& \multirow{3}{*}{Localization} & ChestX-ray14 & 690 & 98 & 196 & N \\
& & RSNA Pneumonia & 25,184 & 1,500 & 3,000 & Y \\
& & MS-CXR & 808 & 115 & 230 & N\\
% & & \textcolor{black}{MS-CXR} & 638 & 85 & 167 & N\\
\cmidrule{2-6}
& \multirow{3}{*}{Segmentation} & JSRT & 172 & 25 & 50 & N \\
& & CheXmask & 219,316 & 5,000 & 10,000 & N \\
& & MS-PS & 164 & 23 & 46 & N \\
\bottomrule
\end{tabular}
\label{tab:dataset_overview}
\end{table}

\clearpage

\begin{table*}[t]
\centering
\small
\caption{\textcolor{black}{Sensitivity analysis of task weight design, related to Figure \ref{fig:prompt}. We report one representative benchmark for each task: ChestX-ray14 for classification, MS-CXR for localization, JSRT for segmentation, and MIMIC-CXR for report generation. The default setting used in the manuscript is $(w_{\text{cls}}, w_{\text{loc}}, w_{\text{seg}}, w_{\text{rep}}) = (0.15, 0.20, 0.15, 0.50)$.}}
\label{tab:task_weight_ablation}
\begin{tabular}{ccccccccc}
\toprule
\multicolumn{4}{c}{\textbf{Task Weights}} & \multicolumn{4}{c}{\textbf{Representative Evaluation}} \\
\cmidrule(lr){1-4} \cmidrule(lr){5-8}
\textbf{$w_{\text{cls}}$} & \textbf{$w_{\text{loc}}$} & \textbf{$w_{\text{seg}}$} & \textbf{$w_{\text{rep}}$}
& \textbf{Classification} & \textbf{Localization} & \textbf{Segmentation} & \textbf{Report Gen.} \\
& & & 
& \textbf{ChestX-ray14} & \textbf{MS-CXR} & \textbf{JSRT} & \textbf{MIMIC-CXR} \\
& & & 
& \textbf{AUC / F1} & \textbf{ACC / mIoU} & \textbf{Dice} & \textbf{RadGraph F1 / ROUGE-L} \\
\midrule
0.50 & 0.15 & 0.15 & 0.20 & 75.4 / 28.8 & 55.9 / 51.3 & 88.2 & 31.4 / 27.1 \\
0.15 & 0.50 & 0.15 & 0.20 & 73.2 / 27.2 & 57.9 / 53.4 & 90.8 & 30.7 / 26.3 \\
0.15 & 0.15 & 0.50 & 0.20 & 73.6 / 28.5 & 57.1 / 52.2 & 91.7 & 31.2 / 26.7 \\
0.15 & 0.20 & 0.15 & 0.50 & 74.3 / 27.1 & 56.9 / 52.4 & 90.0 & 33.1 / 27.5 \\
\bottomrule
\end{tabular}
\label{tab:abl}
\end{table*}

\clearpage

\begin{table}[t]
\centering
\caption{Demonstration of training hyperparameters, related to Figure \ref{fig:overview}.}
\begin{tabular}{lc}
\hline
\textbf{Hyperparameters}        & \textbf{Values}     \\ \hline
Batch Size                      & 256                 \\ 
Learning Rate                   & 1e-5                \\ 
Warm-up Learning Rate           & 1e-7                \\ 
Learning Rate Decay             & Cosine              \\ 
Epochs                          & 30              \\ 
Optimizer                       & AdamW               \\ 
Adam $\beta$                    & (0.9, 0.999)        \\
Weight Decay                    & 0.01                \\ 
Dropout Rate                    & 0.1                 \\ 
Beam Width                      & 6                   \\ 
Image Size                      & 512x512             \\ \hline
\end{tabular}
\label{tab:hyperparameters}
\end{table}

\clearpage

\begin{table}[t]
\centering
\caption{\textcolor{black}{Class distribution in the ChestX-ray14 and CheXpert disease classification test dataset, related to Figure \ref{fig:overview}.}}
% \begin{tabular}{l*{14}{c}}
\begin{tabularx}{\textwidth}{lc*{14}{X}}
\toprule
\textbf{Dataset} & \textbf{Total} & \rotatebox{90}{Atelectasis} & \rotatebox{90}{Cardiomegaly} & \rotatebox{90}{Effusion} & \rotatebox{90}{Infiltration} & \rotatebox{90}{Mass} & \rotatebox{90}{Nodule} & \rotatebox{90}{Pneumonia} & \rotatebox{90}{Pneumothorax} & \rotatebox{90}{Consolidation} & \rotatebox{90}{Edema} & \rotatebox{90}{Emphysema} & \rotatebox{90}{Fibrosis} & \rotatebox{90}{Pleural Thickening} & \rotatebox{90}{Hernia} \\
\midrule
\textbf{ChextXray14} & 25,595 & 3,279 & 1,069 & 4,658 & 6,112 & 1,748 & 1,623 & 555 & 2,665 & 1,815 & 925 & 1,093 & 435 & 1,143 & 86 \\
\textbf{CheXpert} & \textcolor{black}{234} & \textcolor{black}{80} & \textcolor{black}{68} & \textcolor{black}{67} & - & - & - & - & - & \textcolor{black}{33} & \textcolor{black}{45} & - & - & - & - \\
\bottomrule
% \end{tabular}
\end{tabularx}
\label{table:cls_dataset_distribution}
\end{table}

\clearpage

\begin{table}[t]
\centering
\caption{\textcolor{black}{Class distribution in the ChestX-ray14 and MS-CXR lesion localization test dataset, related to Figure \ref{fig:overview}.}}
\begin{tabular}{lccccccccccccc}
\hline
\textbf{Dataset} &\textbf{Total} & \rotatebox{90}{Atelectasis} & \rotatebox{90}{Cardiomegaly} & \rotatebox{90}{Effusion} & \rotatebox{90}{Pneumonia} & \rotatebox{90}{Pneumothorax} & \rotatebox{90}{Consolidation} & \rotatebox{90}{Edema} & \rotatebox{90}{Opacity} & \rotatebox{90}{Infiltrate} & \rotatebox{90}{Mass} & \rotatebox{90}{Nodule} \\
\hline
ChestX-ray14 & 196 & 36 & 34 & 29 & 16 & 22 & - & - & - & 28 & 17 & 14 \\
% MS-CXR & 230 & 59 & 67 & 27 & 20 & 31 & 18 & 3 & 21 & - & - & - \\
 MS-CXR & 230 & 12 & 66 & 19 & 36 & 49 & 23 & 9 & 16 & - & - & - \\
% MS-CXR all& 1153 & 61 & 333 & 96 & 182 & 237 & 117 & 46 & 81 & - & - & - \\
% MS-CXR single box all& 890 &  &  &  &  &  &  &  &  & - & - & - \\
\hline
\end{tabular}
\label{table:loc_dataset_distribution}
\end{table}

\clearpage

\begin{table}[t]
\centering
\caption{\textcolor{black}{Comparisons of the test dataset split for the MS-CXR lesion localization dataset, related to Figure \ref{fig:overview}.
MedRPG \cite{chen2023medical} includes single-box cases from MS-CXR, comprising 167 phrases and 162 images. ChEX \cite{muller2025chex} includes only samples from the official MIMIC-CXR validation and test splits, totaling 196 phrases and 169 images. MAIRA-2 \cite{bannur2024maira} features a test split of 176 phrases and 155 images, alongside each of the splits from MedRPG (138 phrases, 124 images) and ChEX (30 samples, 24 images), respectively.
% Since the final MAIRA-2 \cite{bannur2024maira} was trained using a portion of the MS-CXR dataset, we report results on the intersections of our new held-out test split (176 phrases, 155 images) with each of the splits from MedRPG (138 phrases, 124 images) and ChEX (30 samples, 24 images), respectively.
}}
\begin{tabular}{llcccccccccc}
\hline
\textbf{Split}&Type &\textbf{Total} & \rotatebox{90}{Atelectasis} & \rotatebox{90}{Cardiomegaly} & \rotatebox{90}{Effusion} & \rotatebox{90}{Pneumonia} & \rotatebox{90}{Pneumothorax} & \rotatebox{90}{Consolidation} & \rotatebox{90}{Edema} & \rotatebox{90}{Opacity} \\
\hline

Ours&All& 1,153 & 61 & 333 & 96 & 182 & 237 & 117 & 46 & 81  \\
 % &Train&808\\
 % &Validation&115\\
 % &Test&230 & 59 & 67 & 27 & 20 & 31 & 18 & 3 & 21 \\
 &Test & 230 & 12 & 66 & 19 & 36 & 49 & 23 & 9 & 16 \\
MedRPG\cite{chen2023medical}&All& 890 & 24 & 333 & 50 &133  &230  & 59 & 5 &56  \\
% &Train& 638     & 16 & 232 & 35 & 96 &179  & 40 & 3 & 37 \\
% &Validation& 85 & 2 & 29 & 4 & 19 & 15 & 7 & 1 & 8 \\
&Test& 167      & 6&  72& 11 & 18 &  36& 12 & 1 & 11 \\
ChEX\cite{muller2025chex}
% &All& \\
&Test&196&-&-&-&-&-&-&-&-\\
MAIRA-2\cite{bannur2024maira}
% &All& \\
&Test&176&-&-&-&-&-&-&-&-\\
\hline
\end{tabular}
\label{table:loc_dataset_split}
\end{table}

\clearpage

\begin{figure}[t]
	\centering
    \includegraphics[width=0.99\linewidth]{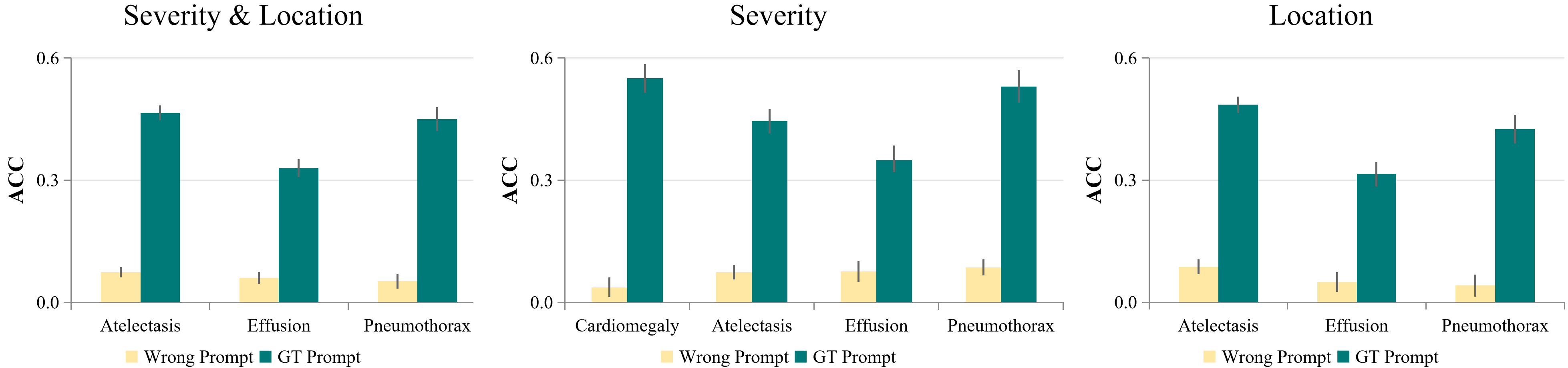}
 \caption{
Comparison of report generation with wrong and ground-truth attribute prompts, related to Figure \ref{fig:prompt}. We evaluate the accuracy of generated severity and location descriptions using deliberately wrong intermediate prompts and ground-truth prompts. Compared with correct intermediate prompts, deliberately wrong severity/location prompts consistently reduce the accuracy of generated attribute descriptions. This confirms that the report generator uses the intermediate attributes as effective evidence, and that incorrect attributes can adversely affect the factual accuracy of generated reports.}
    \label{fig:wrongprompt}
\end{figure}

\clearpage

\begin{figure}[t]
	\centering
    \includegraphics[width=0.99\linewidth]{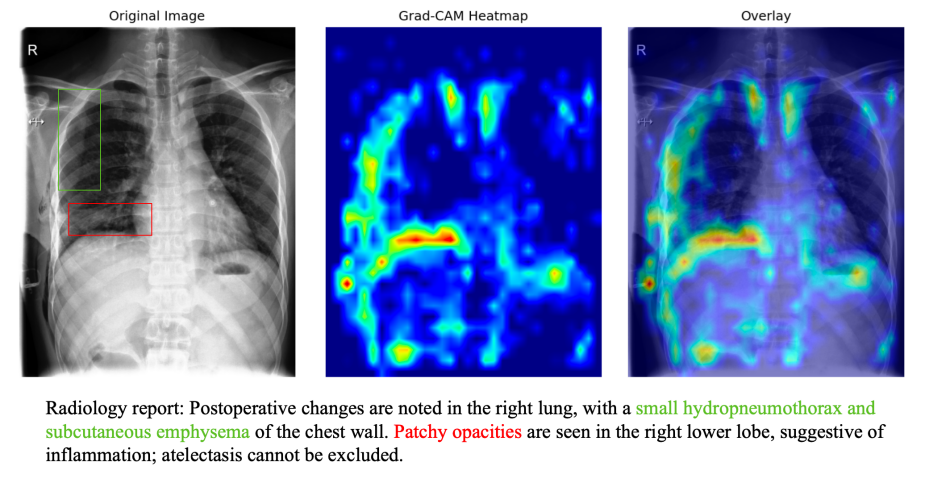}
 \caption{
\textcolor{black}{Cross attention and Grad CAM visualizations of model focus, related to Figure \ref{fig:prompt}. The gradient-based class activation mapping (Grad-CAM) visualizations indicate that the regions with the highest response largely overlap with the clinically relevant areas described in the report, particularly the right lower lung, where the reported abnormalities are located. This suggests that the model predictions are primarily driven by disease-related visual evidence rather than instruction patterns alone. However, we also observe several limitations: non-pathological regions, such as parts of the abdomen and lower neck, exhibit secondary activations, indicating that the attention is not perfectly localized and may still be influenced by surrounding structures.}}
    \label{fig:grad}
\end{figure}

\clearpage

\begin{figure}[t]
	\centering
    \includegraphics[width=0.99\linewidth]{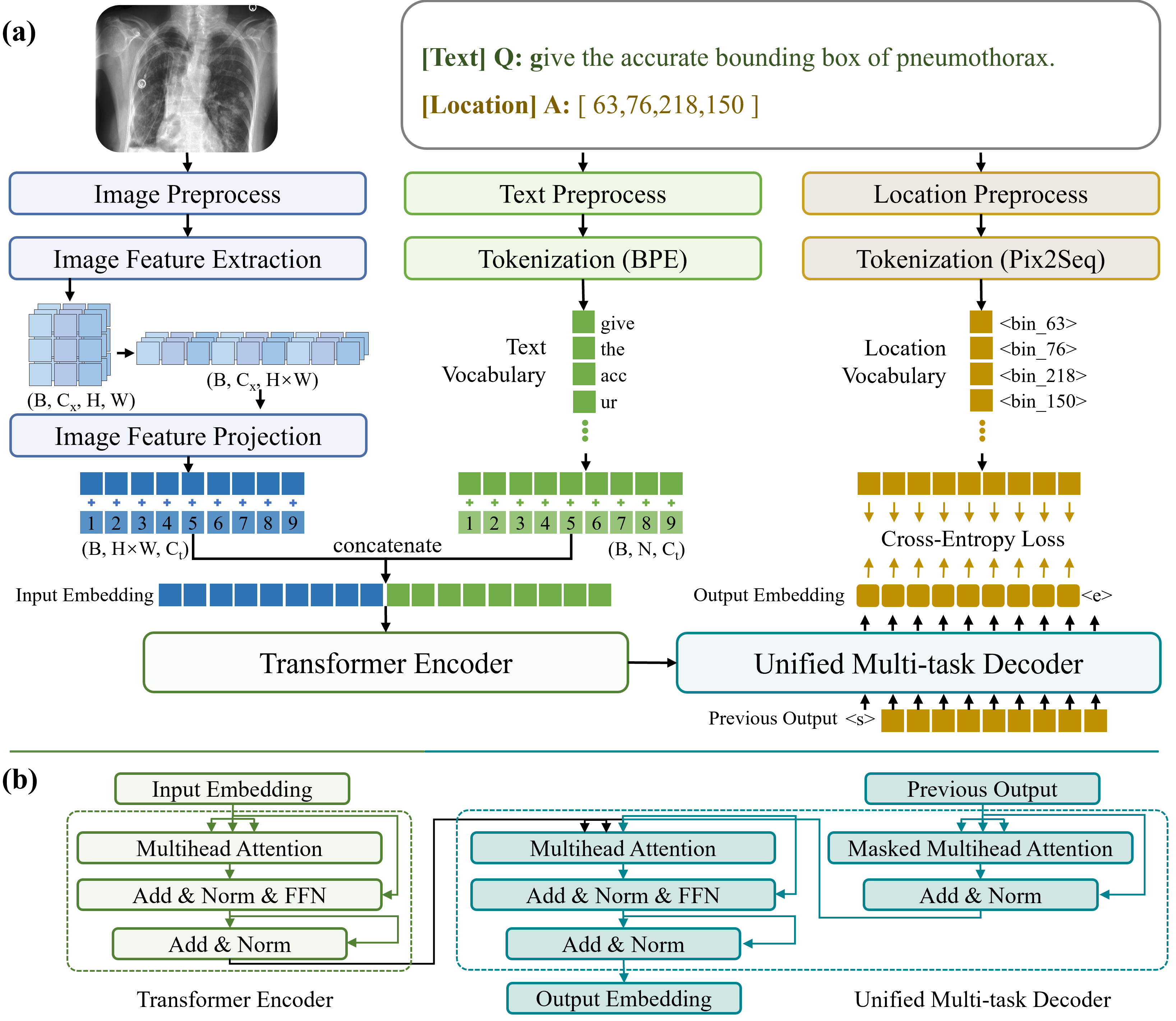}
     \caption{Model Details, related to Figure \ref{fig:overview}. (a) The proposed model’s training pipeline begins with multimodal input preprocessing: chest X-ray images are resized and encoded into feature maps aligned with text embedding dimensions, while clinical text and anatomical location data are tokenized via Byte Pair Encoding (BPE) and Pix2Seq vocabulary, respectively. These aligned embeddings are concatenated and processed by a transformer encoder, whose outputs serve as keys/values for the unified multi-task decoder via cross-attention. The entire framework is optimized end-to-end through autoregressive training. (b) Architectural details include the transformer encoder layers and task-agnostic decoder blocks, where B denotes batch size, H×W the spatial dimensions of image features, N the text token count, and $C_{x}$/$C_{t}$ the channel dimensions of image/text features.}
	\label{fig:method}
\end{figure}

\clearpage

\clearpage

\end{document}